\documentclass[]{jair}

\setcopyright{cc}
\acmDOI{10.1613/jair.1.20374}

\JAIRAE{Tongliang Liu}
\JAIRTrack{} 
\acmVolume{86}
\acmArticle{51}
\acmMonth{08}
\acmYear{2026}

\RequirePackage[
  datamodel=acmdatamodel,
  style=acmauthoryear,
  backend=biber,
  giveninits=true,
  uniquename=init
  ]{biblatex}

\usepackage{amsmath,amsfonts,bm}

\def\eqref#1{equation~\ref{#1}}

\def\1{\bm{1}}

\def\vmu{{\bm{\mu}}}
\def\vtheta{{\bm{\theta}}}

\def\vf{{\bm{f}}}

\def\vk{{\bm{k}}}

\def\vx{{\bm{x}}}
\def\vy{{\bm{y}}}

\def\mI{{\bm{I}}}

\def\mK{{\bm{K}}}

\def\mX{{\bm{X}}}

\DeclareMathAlphabet{\mathsfit}{\encodingdefault}{\sfdefault}{m}{sl}
\SetMathAlphabet{\mathsfit}{bold}{\encodingdefault}{\sfdefault}{bx}{n}

\def\sR{{\mathbb{R}}}

\newcommand{\E}{\mathbb{E}}

\DeclareMathOperator*{\argmax}{arg\,max}

\usepackage{subcaption,siunitx,booktabs} 

\begin{document}

\title[Information-Based Calibration of Uncertainty Quantification in Product-of-Experts Gaussian Process Models]{Information-Based Calibration of Uncertainty Quantification in Product-of-Experts Gaussian Process Models }

\author{Yean Hoon Ong}
\authornote{Corresponding Author.}
\orcid{0000-0002-8452-0999}
\email{yean-hoon.ong@ucl.ac.uk}
\affiliation{%
  \institution{University College London}
  \city{London}
  \country{UK}
}

\author{Paolo Barucca}
\email{p.barucca@ucl.ac.uk}
\orcid{0000-0003-4588-667X}
\affiliation{%
  \institution{University College London}
  \city{London}
  \country{UK}}

\author{Wei Pan}
\email{wei.pan@manchester.ac.uk}
\orcid{0000-0003-1121-9879}
\affiliation{%
  \institution{The University of Manchester}
  \city{Manchester}
  \country{UK}}

\author{Jun Wang}
\email{jun.l.wang@ucl.ac.uk}
\orcid{0000-0002-4021-4228}
\affiliation{%
  \institution{University College London}
  \city{London}
  \country{UK}}

\renewcommand{\shortauthors}{Ong, Barucca, Pan \& Wang}

\begin{abstract}
{\bf Background:} 
    In Gaussian process (GP) regression, the use of a single global GP (GP-glo) incurs cubic computational cost, which limits scalability to large datasets.
    Product-of-experts GP models (GP-pro), which combine a collection of local GP models that collaboratively capture global correlations, provide a prominent approach to alleviating this computational burden.
    However, GP-pro models often produce overestimated posterior variances due to training local experts on disjoint subsets of the data.
    
    {\bf Objectives:}
    This study aims to analyse the overestimation of posterior variances in GP-pro models and to address this issue using an information-based method.  
    
    {\bf Methods:}
    We propose GP-pro-c, a product-of-experts Gaussian process model that incorporates an information-based method to calibrate the overestimated posterior variances. The method uses the monotonicity and submodularity properties of information gain in GP to define a calibration ratio that appropriately reduces the posterior variance of individual local GP models.  
    
    {\bf Results:}
    The performance of GP-pro-c was evaluated using three metrics: negative log-likelihood (NLL), root mean squared error (RMSE), and expected normalised calibration error (ENCE) of the uncertainty estimates.
    Empirical experiments on four synthetic functions and six regression datasets show that the proposed calibration method enables the GP-pro-c model to achieve average reductions of 2.3\% and 12.0\% in NLL and ENCE, respectively, compared to the uncalibrated GP-pro model. 
        
    {\bf Conclusions:}
    The results demonstrate that GP-pro-c effectively mitigates the overestimation of posterior variances in product-of-experts Gaussian process models while maintaining predictive accuracy and reducing computational complexity. These findings suggest that the proposed information-based calibration method is a promising approach for improving uncertainty estimation in scalable GP models.
    We expect the GP-pro-c  to serve as a useful surrogate model in Bayesian optimisation settings  involving high-dimensional and large-scale data.
\end{abstract}


\received{10 September 2025}
\received[accepted]{15 June 2026}

\maketitle

\section{Introduction}
\label{sec:sec_intro}

Gaussian process (GP) models \cite{Rasmussen2006} are nonparametric methods widely used for nonlinear regression. 
They provide both posterior predictions and uncertainty quantification, which support principled decision-making. 
A single global GP model is defined by a mean function and a covariance function. 
Given $n$ training data points, model fitting and inference require inversion of the covariance matrix, resulting in cubic computational cost $\mathcal{O}(n^3)$ and quadratic memory requirements $\mathcal{O}(n^2)$. 
These limitations hinder the application of global GP models to large-scale problems.

A variety of approaches have been proposed to address the scalability limitations of GP models. 
These can be broadly categorised into three classes: sparse inducing-point methods, expert-based methods, and dimensionality-reduction methods. 
Sparse inducing-point approaches \citep{Nguyen2014, Liu2020} reduce computational cost by summarising the dataset with a smaller set of representative points. While effective for moderate input dimensionality, they may struggle to capture highly localised structures or scale to very large datasets. 
Expert-based approaches \cite{Cao2018,Tresp2000bcm,Deisenroth2015,Liu2020} partition the input space and aggregate predictions from multiple local models. 
Dimensionality-reduction approaches, such as the principal component-based method of \citet{Higdon2008}, address scalability from a different perspective by projecting high-dimensional outputs into a lower-dimensional subspace. This reduces the effective output dimensionality and enables efficient emulation of complex computer models. Unlike expert-based approaches, which primarily address scalability with respect to the number of observations and input structure, principal component-based methods focus on output compression; the two approaches are therefore complementary.

In this work, we focus on expert-based approaches, specifically the product-of-experts GP model, which combines a collection of local GP models \cite{Cao2014,Cao2015,Cao2018,Tresp2000bcm,Deisenroth2015,Liu2019,Liu2020}. 
The key idea is to partition the training data into $M$ subsets, each containing $\tilde{n}$ points with $\tilde{n} \ll n$, and to train a GP model on each subset. 
At test time, predictions are aggregated by multiplying the outputs of the local experts. 
This reduces the training complexity to $\mathcal{O}(M \tilde{n}^3)$ and the memory requirement to $\mathcal{O}(M(\tilde{n}^2 + \tilde{n}d))$ \cite{Liu2020}, where $M$ is the number of local GP models, $d$ denotes the input dimensionality, and assuming that each local GP uses the same number of $ \tilde{n}$ training data points.

Three notable variants of the product-of-experts framework are the Bayesian committee machine (BCM) \cite{Tresp2000bcm}, the generalised product of experts (gPoE) \cite{Cao2014,Cao2015,Cao2018}, and the robust Bayesian committee machine (rBCM) \cite{Deisenroth2015}. 
The BCM and rBCM assume a shared GP prior across all experts and require identical hyperparameters, which can limit flexibility in modelling locally varying patterns \cite{Liu2019}. 
Compared to gPoE, rBCM offers stronger Bayesian consistency and improved behaviour in data-sparse regions due to its principled prior correction. 
In contrast, gPoE provides several practical advantages: (i) it captures local patterns by allowing each expert to learn distinct hyperparameters; (ii) it enables flexible weighting of individual experts; (iii) it captures long-range correlations; and (iv) it reduces computational cost \cite{Cao2014, Cao2015, Deisenroth2015,Liu2020}. 
We note that while gPoE can adapt to data-induced heterogeneity across experts, this is not equivalent to modelling true nonstationary kernels.
For the above-mentioned practical reasons, we focus on the gPoE model.
For brevity, we use GP-pro to refer to the generalised product-of-experts model unless otherwise stated.

Despite its advantages, GP-pro models tend to overestimate predictive uncertainty (variance). 
This occurs because each local GP is trained on only a subset of the data: while this reduces computational cost, it also limits the information available to each expert, causing each local predictive variance to be larger than it would be if the GP were trained on all data.
Since GP predictive variance decreases as more training data are incorporated \cite{Desautels2014}, aggregating predictions from multiple local experts without correction propagates this overestimation to the final prediction \cite{Deisenroth2015,Szabo2019,Liu2020}. 
Such miscalibration can negatively affect downstream tasks, including Bayesian optimisation, active learning, and other uncertainty-sensitive applications.

Researchers have proposed the use of a scaling temperature to calibrate the predictive uncertainties in GP-pro models \cite{Cohen2020}. 
In this approach, the predictive variances of all local GPs are multiplied by a constant factor to adjust their magnitude and better reflect the true variability of the function. However, the method requires the user to pre-define this temperature constant, and if it is set too high or too low, the model can become overly uncertain or overconfident, meaning that the predicted variances still do not accurately match the actual uncertainty.

In this work, we address variance overestimation in GP-pro models by proposing an information-based calibration method. 
The method adjusts the posterior variance of each local expert at test time using quantities already available in standard GP-pro predictions. 
It leverages the monotonicity and submodularity of information gain to determine an appropriate calibration factor. 
This results in improved uncertainty estimates while preserving computational efficiency. 
Empirical results demonstrate that the proposed approach yields well-calibrated uncertainty without compromising predictive accuracy.

This study makes four main contributions:
\begin{itemize}
  \item An analysis of variance overestimation in GP-pro models.
  \item An information-based calibration method for uncertainty quantification in GP-pro, resulting in the GP-pro-c model.
  \item A comprehensive comparison of GP-pro and GP-pro-c across multiple data assignment methods and weighting schemes.
  \item An empirical evaluation of GP-pro-c against GP-pro, temperature-scaled GP-pro, and a global GP model in terms of predictive performance and computational cost.
\end{itemize}

The remainder of this paper is organised as follows. 
Section \ref{sec:sec_background} introduces background on global GP models, GP-pro, and relevant information-theoretic concepts. 
Section \ref{sec:sec_related_work} reviews related work. 
Section \ref{sec:sec_analyse} analyses variance overestimation in GP-pro. 
Section \ref{sec:sec_calibration} presents the proposed  information-based calibration method. 
Section \ref{sec:sec_experiments} reports experimental results, and Section \ref{sec:sec_s3_conclusion} concludes the paper.

\section{Preliminaries}
\label{sec:sec_background}

A Gaussian process (GP) is a nonparametric probabilistic model that defines a distribution over functions via a collection of random variables. 
The reader is referred to \citet{Rasmussen2006} for a comprehensive introduction. 
A GP is specified by a mean function and a covariance function, and is written as:
\begin{align*}
f(\vx) \sim GP \left( \tilde{\mu}(\vx), \, k(\vx, \vx') \right)
\end{align*}
where $f(\vx)$ is a real-valued stochastic process, and $\tilde{\mu}(\vx)$ and $k(\vx, \vx')$ denote the mean and covariance functions, respectively, defined as:
\begin{align*}
\tilde{\mu}(\vx) &= \mathbb{E}[f(\vx)] \\
k(\vx, \vx') &= \mathbb{E}\left[(f(\vx) - \tilde{\mu}(\vx))(f(\vx') - \tilde{\mu}(\vx'))\right]
\end{align*}

The hyperparameters of the prior mean function and the covariance function, denoted by $\vtheta$, are learned from data.
Given a set of $n$ training data points and their corresponding observations $ \mathcal{D} = \{ ( \vx_i, y_i ) \}_{i \leqslant n}$, where $\vx_i$ denotes an input vector with dimension $d$ and $y_i$ denotes a scalar output or target. 
The observed value $y_i$ is assumed to differ from the function values $f(\vx_i)$ by additive noise, and 
this noise is assumed to follow an independent, identically distributed Gaussian distribution with zero mean and variance $\ddot{\sigma}^2$, i.e., $y_i = f(\vx_i) + \epsilon_i$, $\epsilon_i \sim \mathcal{N}(0, \ddot{\sigma}^2)$.

The hyperparameters $\vtheta$ are estimated by maximising the log marginal likelihood:
\begin{align}
\log p( \vy | \mX, \vtheta) 
&= - \frac{1}{2} (\vy - \tilde{\vmu})^{\top} \ \left( \mK +  \ddot{\sigma}^2 \ \mI \right)^{-1} (\vy - \tilde{\vmu}) 
        - \frac{1}{2} \log \left| \mK +  \ddot{\sigma}^2 \ \mI \right| 
        - \frac{n}{2} \log 2 \pi   \label{eq:eq_gp_loglikelihood}
\end{align}
where $\tilde{\vmu}$ is the vector of mean function evaluations, and $\mK$ is the covariance matrix with entries $K_{ij} = k(\vx_i, \vx_j)$.

For a test input $\vx'$, the predictive distribution of $y' = f(\vx') + \epsilon'$ is Gaussian:
\begin{align*}
p(y' \mid \mathcal{D}, \vx') = \mathcal{N}\left( \mu(\vx'), \, \sigma^2(\vx') + \ddot{\sigma}^2 \right)
\end{align*}
where the posterior mean and variance are given by:
\begin{align}
\mu(\vx') &= \tilde{\mu}(\vx') + \vk(\vx')^{\top} \left( \mK + \ddot{\sigma}^2 \mI \right)^{-1} (\vy - \tilde{\vmu})  \label{eq:eq_gp_mu} \\
\sigma^2(\vx') &= k(\vx', \vx') - \vk(\vx')^{\top} \left( \mK + \ddot{\sigma}^2 \mI \right)^{-1} \vk(\vx')  \label{eq:eq_gp_sigma}
\end{align}
and $\vk(\vx') = [k(\vx', \vx_1), \ldots, k(\vx', \vx_n)]^\top$ denotes the covariance vector between the test input and the training inputs.

\subsection{Product-of-Experts Gaussian Process Models (GP-pro) }

The GP-pro model consists of a collection of $M > 1$ local GP models, where each local GP models a distinct region of the input domain and performs independent predictions and uncertainty estimation. The overall predictive distribution is obtained by combining the outputs of the individual local GPs through a product-of-experts formulation. 
The two essential components of a GP-pro model are (i) the data assignment method used to allocate training data to each local GP and (ii) the weighting scheme used to aggregate the posterior predictions and uncertainties provided by each expert.

\paragraph{Assigning data and training the GP-pro model} 

Let $N_{GP}$ denote the maximum number of data points assigned to each local GP model.
Given a training dataset $\mathcal{D}$, the number of local GP models is determined as $M = \mathrm{int}(|\mathcal{D}| / N_{GP})$.
Each local $GP^{(m)}$ is assigned a subset $\mathcal{D}^{(m)} \subset \mathcal{D}$.
Common data assignment methods include random partitioning, $k$-means clustering, and balltree methods \cite{Cao2014,Deisenroth2015,Cao2018,Liu2019}.

With random assignment, each local $GP^{(m)}$ is allocated the equal number of training data points, i.e., $|\mathcal{D}^{(m)}| = |\mathcal{D}| / M$. 
Data points within each  $ \mathcal{D}^{(m)}$ are sampled uniformly (with or without replacement) from $\mathcal{D}$.

With $k$-means assignment, the training inputs $\{\vx_i\}_{i \leqslant n}$ are partitioned into $M$ disjoint clusters using the $k$-means algorithm \cite{Arthur2007}. Each local GP is then trained on one cluster, resulting in potentially uneven subset sizes.

With balltree assignment, a balltree \cite{omohundro1989balltree} is constructed such that the training data points are partitioned into a nested set of balls by recursively splitting the points into two disjoint sets based on a distance metric.
In this study, each leaf node contains at least $|\mathcal{D}|/M$ data points.
Traversing from the leaf nodes to the parent nodes, $M$ disjoint subsets of data are constructed by randomly choosing $ \left| \mathcal{D}^{(m)} \right| $ data points  without replacement from individual nodes. Each subset of data $ \mathcal{D}^{(m)}$ is used to fit a local $GP^{(m)}$ model.

The random assignment method is computationally efficient, incurring only linear-time complexity in the number of data points. The $k$-means and balltree methods introduce additional overhead due to clustering or spatial indexing, but this cost is negligible relative to the cubic $\mathcal{O}(n^3)$ training cost of a global GP. 
In this study, we use the $k$-means implementation from SciPy\footnote{https://docs.scipy.org/doc/scipy/reference/generated/scipy.cluster.vq.kmeans2.html} and the balltree implementation from scikit-learn\footnote{https://scikit-learn.org/stable/modules/generated/sklearn.neighbors.BallTree.html}. Empirical runtime results later in the paper confirm that the overhead of these assignment methods is minor compared to GP training and inference.

Each local GP model $GP^{(m)}$ is trained on the respective $\mathcal{D}^{(m)}$ using a constant prior mean function and a Mat\'{e}rn-5/2 covariance function.
The hyperparameters of $GP^{(m)}$, denoted by $\vtheta^{(m)}$, are estimated by maximising the log marginal likelihood.
The marginal likelihood for each local GP expert has the same functional form as the global GP marginal likelihood in Equation (\ref{eq:eq_gp_loglikelihood}), but is evaluated independently for each local dataset $\mathcal{D}^{(m)}$ with expert-specific hyperparameters $\vtheta^{(m)}$:
\begin{align}
\log p( \vy^{(m)} | \mX^{(m)},  \vtheta^{(m)}) = &- \frac{1}{2} \left( \vy^{(m)} - \tilde{\vmu}^{(m)} \right)^{\top} \left( \mK^{(m)} + \ddot{\sigma}_{m}^2 \ \mI \right)^{-1} \left( \vy^{(m)} -  \tilde{\vmu}^{(m)} \right)  \notag \\
								    &-  \frac{1}{2} \log \left| \mK^{(m)} + \ddot{\sigma}_{m}^2 \ \mI \right| - \frac{ \left| \mathcal{D}^{(m)} \right| }{2} \log 2 \pi    \label{eq:eq_c3_1}
\end{align}
where $\tilde{\vmu}^{(m)}$ is a vector of mean function values for $\mX^{(m)}$, $\ddot{\sigma}_{m}^2$ is the noise variance, and $ \mK^{(m)} + \ddot{\sigma}_{m}^2 \ \mI$ is the covariance matrix for the noisy target $\vy^{(m)}$.

\paragraph{Aggregating posterior predictions} 

At test time, at candidate $\vx'$, each local $GP^{(m)}$ computes its posterior predictive mean $\mu_{m} (\vx')$ and variance $\sigma_{m}^2 (\vx')$ independently. 
These predictive expressions have the same functional form as the global GP predictive mean and variance in Equations (\ref{eq:eq_gp_mu}) and (\ref{eq:eq_gp_sigma}), but is evaluated independently for each local dataset $\mathcal{D}^{(m)}$ with expert-specific hyperparameters $\vtheta^{(m)}$:
\begin{align}
\mu_{m}(\vx') &=   \tilde{\mu}^{(m)}(\vx')  +  \vk^{(m)}(\vx')^{\top} \left(  \mK^{(m)} + \ddot{\sigma}_{m}^2 \ \mI  \right)^{-1} \left( \vy^{(m)} -  \tilde{\vmu}^{(m)}  \right) \\
\sigma_{m}^2(\vx') &= k^{(m)}(\vx', \vx') -  \vk^{(m)}(\vx')^{\top}  \left(  \mK^{(m)} + \ddot{\sigma}_{m}^2 \ \mI \right)^{-1} \vk^{(m)}(\vx') 
\end{align}
where $ \tilde{\mu}^{(m)}$ is the mean function, $k^{(m)}$ is the covariance function,  $\vk^{(m)}$ is the vector of covariance terms between  $\vx'$ and $\mX^{(m)}$, and $ \mK^{(m)} + \ddot{\sigma}_{m}^2 \ \mI$ is the covariance matrix.

Following \citet{Cao2014}, the aggregated predictive distribution is defined as a product of expert predictions:
\begin{align*}
p(y' | \vx', \mathcal{D} ) = \prod_{m=1}^M p^{w_m(\vx')} ( y' | \vx',  \mathcal{D}^{(m)} )
\end{align*}
This yields the aggregated predictive mean and variance:
\begin{align}
\mu(\vx') &= \sigma^2(\vx')  \sum_{m=1}^M \left( w_{m}(\vx') \  \mu_{m}(\vx') \ \frac{1}{\sigma_{m}^2(\vx')} \right) \\
\sigma^2(\vx') &= \left( \sum_{m=1}^M \left( w_{m}(\vx') \  \frac{1}{\sigma_{m}^2(\vx')}  \right)  \right)^{-1}  \label{eq:eq_c3_agg_sigma}
\end{align}

The weight $w_{m}(\vx') \in \sR^+$ reflects the reliability of the $m$-th expert at $\vx'$. The higher the weight $w_{m}(\vx')$, the more reliable the $m$-th expert will be. The weighting factor at each $\vx'$ is normalised such that $\sum_{m'=1}^{M} w_{m'}(\vx') = 1$. 
We consider three weighting schemes: entropy-based, variance-based, and uniform weighting.

The entropy-weighted scheme (ENT) \cite{Cao2014} measures reliability as the reduction in differential entropy between the prior and posterior. 
For the Gaussian distribution, the differential entropy of local $GP^{(m)}$ at candidate $\vx'$ can be computed in closed form as $H_{m} (\vx') = \frac{1}{2}(1 + \log 2 \pi \sigma_{m}^2(\vx'))$. With that, the quantity $\Delta H_{m} (\vx')$ is given as:
\begin{align*}
\Delta H_{m} (\vx')  &= \tilde{H}_{m} (\vx') - H_{m} (\vx') \\
				    &=  \frac{1}{2} \left( 1 + \log 2 \pi \tilde{\sigma}_{m}^2(\vx') \right) -  \frac{1}{2} \left( 1 + \log 2 \pi \sigma_{m}^2(\vx') \right) \\
				    &=  \frac{1}{2} \left( \log \tilde{\sigma}_{m}^2(\vx') -  \log \sigma_{m}^2(\vx') \right)
\end{align*}
where $\tilde{\sigma}_m^2(\vx') = k^{(m)}(\vx', \vx')$ is the prior variance, and $\sigma_{m}^2(\vx')$ is the posterior predictive variance of local $GP^{(m)}$ at candidate $\vx'$.  
Using the quantity $\Delta H_{m} (\vx')$, the normalised weight of a  local $GP^{(m)}$ at  candidate $\vx'$ is: 
\begin{align}
w_{m}^{\text{ENT}}  (\vx')  = \frac{ \Delta H_{m} (\vx') }{ \sum_{m'=1}^M \Delta H_{m'} (\vx') }
\end{align}
A higher difference in entropy indicates a higher reliability of a local $GP^{(m)}$, and therefore the posterior prediction provided by this local GP model should contribute more to the aggregated posterior prediction.

With the variance-weighted scheme (VAR), 
the reliability of local $GP^{(m)}$ is measured as the difference between the prior and posterior variances at candidate $\vx'$. 
The higher the difference in variance, the more reliable the local $GP^{(m)}$.
Therefore, at candidate $\vx'$, the normalised weighting factor $w^{\text{VAR}}$ of a local $GP^{(m)}$ is: 
\begin{align}
w_{m}^{\text{VAR}}  (\vx')  &= \frac{  \tilde{\sigma}_{m}^2(\vx') -  \sigma_{m}^2(\vx') }{ \sum_{m'=1}^M \left(  \tilde{\sigma}_{m'}^2(\vx') -  \sigma_{m'}^2(\vx') \right) }
\end{align}
where $ \tilde{\sigma}_{m}^2(\vx')$ is the prior variance and $\sigma_{m}^2(\vx')$ is the posterior predictive variance of the local $GP^{(m)}$ at the candidate $\vx'$. 

The variance weighting is different from the entropy weighting in that the difference in entropy is unitless, while the difference in variance carries the unit of variance \cite{Cao2014}. 
Moreover, because the variance-weighted scheme depends directly on the magnitude of the posterior variance reduction, it tends to more strongly downweight local $GP^{(m)}$ models that remain highly uncertain at the candidate $\vx'$. 
In contrast, entropy-based weighting induces a smoother penalisation, as entropy grows logarithmically with variance. 
Neither weighting scheme is universally preferable; variance weighting emphasises sensitivity to uncertainty magnitude, while entropy weighting yields smoother aggregation.
Our proposed information-based calibration method is orthogonal to this choice and can be applied in conjunction with either weighting scheme.

For comparison, the uniform-weighted scheme (UNI) assigns equal weight to all experts:
\begin{align}
w_m^{\text{UNI}}(\vx') = \frac{1}{M}
\end{align}

Figure~\ref{fig:fig_c3_glo_vs_gpoe} (right) illustrates an example GP-pro model with two local experts on the one-dimensional Ackley function, using balltree assignment and entropy-weighted aggregation.

\subsection{Information Gain for Covariance Functions}
\label{sec:sec_c2_ig}

In a GP model, the covariance function (kernel) is a fundamental component that encodes prior assumptions about the objective function.
In this section, we introduce key information-theoretic quantities that form the basis of our proposed calibration method for addressing variance overestimation in GP-pro models.

In the framework of Bayesian experimental design \cite{Chaloner1995}, the informativeness of a set of sampling points $\mathcal{A} \subset \mathcal{X}$ about the latent function $f$ is quantified by the information gain \cite{Cover2005ch2}. 
Information gain measures the reduction in uncertainty about $f$ induced by observing data.
Specifically, for a set $\mathcal{A}$, we observe noisy outputs $\vy_{\mathcal{A}} = \vf_{\mathcal{A}} + \epsilon_{\mathcal{A}}$, where $\epsilon \sim \mathcal{N}(0, \ddot{\sigma}^2)$ denotes independent Gaussian noise. Observing $\vy_{\mathcal{A}}$ reduces the uncertainty about the function values $\vf$ over the entire domain $\mathcal{X}$.

The information gain associated with $\mathcal{A}$ is defined as the mutual information between $\vy_{\mathcal{A}}$ and $\vf$:
\begin{align*}
I(\vy_{\mathcal{A}}; \vf) = H(\vf) - H(\vf \mid \vy_{\mathcal{A}}) = H(\vy_{\mathcal{A}}) - H(\vy_{\mathcal{A}} \mid \vf)
\end{align*}
For Gaussian distributions, the differential entropy is given by $H(\mathcal{N}(\mu, \Sigma)) = \frac{1}{2} \log |2\pi e \Sigma|$, and the mutual information can be expressed in terms of a log-determinant:
\begin{align}
I(\vy_{\mathcal{A}}; \vf) = I(\vy_{\mathcal{A}}; \vf_{\mathcal{A}}) 
&= H(\vy_{\mathcal{A}}) - H(\vy_{\mathcal{A}} \mid \vf) \label{eq:eq_c2_1}  \notag  \\
&= \frac{1}{2} \log \left| \mI + \ddot{\sigma}^{-2} \mK_{\mathcal{A}} \right|
\end{align}
where $\mK_{\mathcal{A}} = [k(\vx, \vx')]_{\vx, \vx' \in \mathcal{A}}$ is the covariance matrix of $\vf_{\mathcal{A}}$.
For GP models, $\vy_{\mathcal{A}}$ depends only on $\vf_{\mathcal{A}}$, which implies $H(\vy_{\mathcal{A}} \mid \vf) = H(\vy_{\mathcal{A}} \mid \vf_{\mathcal{A}})$ and hence $I(\vy_{\mathcal{A}}; \vf) = I(\vy_{\mathcal{A}}; \vf_{\mathcal{A}})$ \cite{Srinivas2012,Desautels2014}.

The information gain is both monotonic and submodular \cite{Cover2005ch2,Krause2008,Srinivas2012}. Monotonicity implies that for $\mathcal{A} \subset \mathcal{A}'$, we have
$I(\vy_{\mathcal{A}}; \vf) \leqslant I(\vy_{\mathcal{A}'}; \vf)$.
Submodularity captures a diminishing returns property: adding a new data point to a larger set yields a smaller marginal increase in information gain than adding it to a smaller set.

Let $\gamma_N$ denote the maximum information gain obtainable from any subset $\mathcal{A} \subset \mathcal{X}$ with $|\mathcal{A}| \leqslant N$, defined as 
$\gamma_N = \max_{\mathcal{A} \subset \mathcal{X}, |\mathcal{A}| \leqslant N} I(\vy_{\mathcal{A}}; \vf)$. 
Computing $\gamma_N$ exactly is NP-hard. However, a greedy algorithm provides an efficient approximation \cite{Nemhauser1978,Ko1995,Srinivas2012,Dorard2012}. At iteration $n$, the algorithm selects 
$\vx_n = \argmax_{\vx \in \mathcal{X}} I(\vy_{\mathcal{A}_{n-1} \cup \{\vx\}}; \vf)$, 
where $\mathcal{A}_{n-1} = \{\vx_1, \dots, \vx_{n-1}\}$. For GP models, this selection is equivalent to choosing the point with maximum posterior variance:
$ \vx_n = \argmax_{\vx \in \mathcal{X}} \sigma_{n-1}^2(\vx)$.

Due to submodularity, \citet{Nemhauser1978} showed that the information gain achieved by the greedy selection, denoted $I^{\text{G}}(\vy_{\mathcal{A}}; \vf)$, satisfies:
\begin{align}
I^{\text{G}}(\vy_{\mathcal{A}}; \vf) \geqslant \left(1 - \frac{1}{e}\right) 
\max_{\mathcal{A} \subset \mathcal{X}, |\mathcal{A}| \leqslant N} I(\vy_{\mathcal{A}}; \vf)
\label{eq:eq_c2_4}
\end{align}

Furthermore, \citet{Dorard2012} showed that the greedy information gain can be expressed as a sum of marginal conditional gains:
\begin{align}
I^{\text{G}}(\vy_{\mathcal{A}}; \vf) = \frac{1}{2} \sum_{n=1}^N 
\log \left( 1 + \frac{\sigma_{n-1}^2(\vx_n)}{\ddot{\sigma}^2} \right)
\label{eq:eq_c2_2}
\end{align}
where $\sigma_{n-1}^2(\vx_n)$ is the posterior variance at $\vx_n$ conditioned on $\{y_1, \dots, y_{n-1}\}$.
A proof is provided in Appendix~\ref{appendix_info_gain}.

The maximum information gain $\gamma_N$ depends on both the kernel and the input space \cite{Srinivas2012}. 
Using the greedy bound in Equation (\ref{eq:eq_c2_4}),  \citet{Srinivas2012} derived highly problem-specific bounds on $\gamma_N$, which are simple analytical expressions of $N$ and the power of $k$ (covariance function), for three most commonly used kernels. 
For a linear kernel, $\gamma_N =  \mathcal{O}( d \log N ) $. 
For the squared exponential kernel, $\gamma_N =  \mathcal{O}( ( \log N)^{d+1} ) $. 
For the Mat\'{e}rn kernel with $ \nu > 1$,  $\gamma_N =  \mathcal{O}( N^{\frac{d(d+1)}{2\nu + d(d+1)}} ( \log N) ) +  \mathcal{O}( N^0 )$.

\section{Related Work}
\label{sec:sec_related_work}

The GP-pro model depends on both a data assignment method for distributing data among local GPs and a weighting scheme for aggregating their posterior predictions. 
\citet{Cao2014} conducted a comparative study of random, $k$-means, and balltree assignment methods combined with an entropy-based weighting scheme, reporting that the balltree assignment achieved slightly better performance. 
Most subsequent work \cite{Deisenroth2015,Liu2016,Liu2020} adopts random assignment with entropy-based weighting, while \citet{Cohen2020} considered $k$-means assignment with variance-based weighting. 
To the best of our knowledge, no systematic comparison has been conducted across different combinations of assignment methods and weighting schemes. 
In this study, we perform a comprehensive evaluation of GP-pro variants (with and without calibration) using three assignment methods (random, $k$-means, and balltree) and three weighting schemes (entropy, variance, and uniform), examining both predictive performance and computational overhead.

Uncertainty calibration in GP models has been studied in several contexts. 
In computer model calibration, discrepancy-based approaches introduce an explicit model–data discrepancy term to account for simulator bias (e.g., \citet{Higdon2008}). 
These methods are particularly effective for physical simulations with high-dimensional outputs, but they address structural model mismatch rather than miscalibration arising from approximate inference. 
In contrast, scalable GP models based on local or distributed approximations, such as product-of-experts formulations, exhibit miscalibration due to the aggregation of multiple local posteriors. 
While these models alleviate cubic computational costs, they often produce inaccurate uncertainty estimates. 
The present work addresses this form of miscalibration by proposing an information-based variance calibration method for GP-pro models.

To mitigate overestimated variances in GP-pro, \citet{Cohen2020} proposed a weight calibration approach based on a temperature-scaled softmax function:
\begin{align*}
\tilde{w}_m(\vx') \propto \exp ( - T \psi_m (\vx') ), \quad \sum_{m=1}^M \tilde{w}_m(\vx') = 1
\end{align*}
Here, $T$ is an inverse temperature parameter that controls the concentration of weights, amplifying stronger experts while downweighting weaker ones. 
The functional $\psi_m(\vx')$ describes the level of confidence of the $m$-th expert at the test point $\vx'$, typically defined using the posterior variance or differential entropy.
However, this approach calibrates the aggregation weights rather than directly correcting the overestimated variances of individual experts. 
Moreover, the temperature parameter must be specified a priori, and poor choices can lead to undesirable behaviour: large $T$ values overly concentrate weight on a small subset of experts, while small $T$ values produce overly diffuse weights. 
Both cases can degrade predictive accuracy and uncertainty calibration. 
Additionally, the formulation in \citet{Cohen2020} requires all experts to share the same kernel specification and hyperparameters. 
While this constraint regularises the model and reduces overfitting, it also limits expressiveness, making the model less effective in heteroskedastic and non-stationary settings \cite{Cao2015}.

Related work on information gain in GP models is primarily found in Bayesian optimisation (BO).  
\citet{Srinivas2010,Srinivas2012} used maximum information gain to derive confidence bounds for global GP models in sequential BO. 
\citet{Desautels2014} employed information gain to address underestimated variances in global GP models used in asynchronous batch BO settings.

Motivated by these developments, we propose a method that leverages information gain bounds (specifically for the Mat\'{e}rn kernel) to address variance overestimation in product-of-experts GP models. 
Given $M$ local GP experts trained on data subsets, the key challenge is to adjust their posterior variances such that the aggregated predictive variance accurately reflects the global data distribution. 
To this end, we define a pointwise calibration ratio that can be applied to each local expert at test time using only quantities available in standard GP-pro predictive equations. 
This formulation establishes the problem setting addressed by our method.

\section{Analysing Uncertainty Quantification in Product-of-Experts Gaussian Process Models }
\label{sec:sec_analyse}

In a GP-pro model with $M$ local GPs, each local expert $GP^{(m)}$ is trained on a subset of the data $\mathcal{D}^{(m)}$, while the remaining data $\mathcal{D} \setminus \mathcal{D}^{(m)}$ are effectively unavailable to that expert. 
Training on smaller subsets reduces computational cost; however, it introduces an important side effect: the predictive variance $\sigma_m^2(\vx)$ is systematically overestimated (i.e., overly conservative). 
This follows from the property that, for any fixed input $\vx$, the GP posterior variance is non-increasing with respect to the number of training points \cite{Desautels2014}. 
In other words, as more training data are observed, the shape of $f$ is better understood, leading to reduced uncertainty. 

\begin{figure}
\centering
\includegraphics[width=1.0\textwidth]{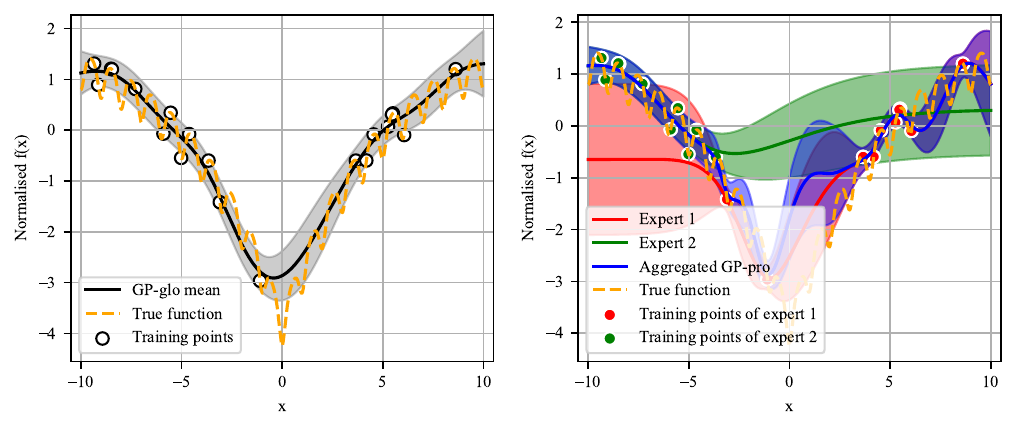}
\caption{Illustration of posterior predictions produced by a single global GP model (GP-glo) and a product-of-experts GP model (GP-pro) for the one-dimensional Ackley function. 
The white dots represent the training data points. 
The plot on the left shows the posterior prediction of the GP-glo model, with the shaded region indicating the standard deviation at each input value $x$. The plot on the right shows a GP-pro model comprising two local experts. The balltree assignment method is used to allocate training data points to the local GP models. The green and red curves, together with their corresponding shaded regions, represent the posterior predictions and standard deviations of the two local GP models. The GP-pro model employs an entropy-weighted aggregation scheme to combine the predictions of the local GP models. The blue curve and shaded region represent the aggregated posterior prediction and its associated standard deviation. Compared with the GP-glo model, the GP-pro model exhibits overestimated posterior variance, as evidenced by the wider blue uncertainty intervals.}
\Description{ Two line graphs (left and right) show the Ackley function values, with the y-axis ranging from −4 to 2 and the x-axis representing input values from −10 to 10 in increments of 5. The left graph presents the function values and their corresponding confidence intervals predicted by a single global Gaussian process (GP) model. The right graph presents the function values and their corresponding confidence intervals predicted by a product of two local GP models. The confidence intervals produced by the product of local GP models are noticeably wider than those of the single global GP model, illustrating the issue of variance overestimation that can arise in product-of-Gaussian-process models.
}
\label{fig:fig_c3_glo_vs_gpoe}
\end{figure}

When aggregating predictions across all local experts $GP^{(m)}$, these inflated variances propagate into the combined posterior (see Figure \ref{fig:fig_c3_glo_vs_gpoe}). 
Consequently, the aggregated predictive variance $\sigma^2(\vx)$ remains overly conservative, and the posterior reflected by $\sigma^2(\vx)$ is `conservative' about the algorithm's actual state of knowledge of the function. 
Figure \ref{fig:fig_c3_glo_vs_gpoe} illustrates this effect.

To quantitatively evaluate the predictive accuracy and uncertainty quality of the GP-pro model, this study uses three metrics: negative log-likelihood (NLL), root mean squared error (RMSE) and expected normalised calibration error (ENCE). 
These metrics have been proposed and used by researchers  \cite{Levi2020,Tran2020} in the literature on uncertainty quantification.

We use root mean squared error (RMSE) to measure point prediction accuracy.
RMSE is used because it is sensitive to outliers and is therefore a good measure of the worst-case accuracy \cite{Tran2020}.

To assess uncertainty calibration, we adopt the regression calibration definition of \citet{Levi2020}. 
Let $\mu(\vx)$ and $\sigma^2(\vx)$ denote the predicted mean and variance. A model is well calibrated if:
\begin{align*}
\forall \sigma: \ \E_{\vx, y} \left[ (\mu(\vx) - y)^2 \mid \sigma^2(\vx) = \sigma^2 \right] = \sigma^2
\end{align*}
This condition states that, for predictions with a given uncertainty level, the expected squared error should match the predicted variance $\sigma^2$.

In practice, calibration is evaluated via binning. 
Let $\sigma_t$ be the standard deviation of the predicted output probability density function (PDF) $p_t$, and assume  that the examples are ordered by increasing values of $\sigma_t$, and the number of bins, $J$, divides the number of examples, $T$.  
The indices of the examples are divided into $J$ bins, $\{B_j\}^J_{j=1}$, such that
$B_j = \{ (j - 1) \cdot \frac{T}{J} + 1, \ldots, j \cdot \frac{T}{J} \}$. 
Each bin therefore corresponds to an interval in the standard deviation axis: $ [ \min_{t \in B_j} \{ \sigma_t \},  \max_{t \in B_j} \{ \sigma_t \} ] $. The intervals are non-overlapping, and their boundary values are increasing.
Following \citet{Levi2020}, we use $J = 10$ bins.

For each bin $j$, we compute the root of the mean variance (RMV) and the empirical root mean square error (RMSE): 
\begin{align*}
RMV(j) &= \sqrt{ \frac{1}{|B_j|} \sum_{t \in B_j} \sigma_t^2 } \\
RMSE(j) &= \sqrt{ \frac{1}{|B_j|} \sum_{t \in B_j} (y_t - \hat{y}_t)^2 }
\end{align*}
where $\hat{y}_t$ is the predicted mean. 
The expected normalised calibration error (ENCE) \cite{Levi2020} is then defined as:
\begin{align*}
ENCE = \frac{1}{J} \sum_{j=1}^{J} \frac{|RMV(j) - RMSE(j)|}{RMV(j)}
\end{align*}
ENCE averages the calibration error in each bin, normalised by the bin’s mean predicted variance, since a larger variance is expected to have larger errors.

\begin{figure}
    \centering 

\begin{subfigure}{0.35\textwidth}
  \includegraphics[width=\linewidth]{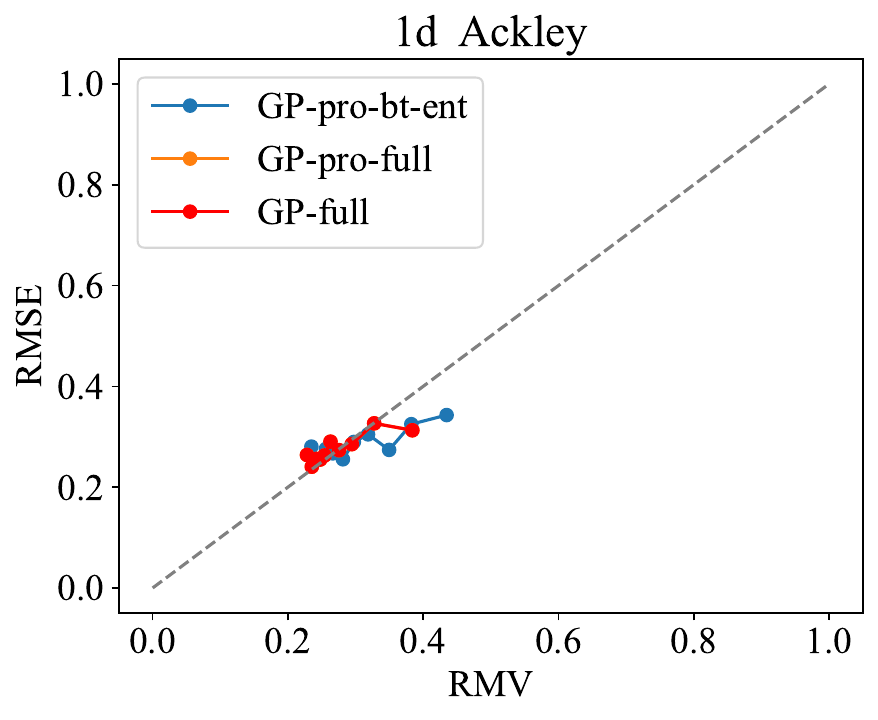}
   \caption{$1d$ Ackley}
  \label{fig:1}
\end{subfigure}\hfil 
\begin{subfigure}{0.35\textwidth}
  \includegraphics[width=\linewidth]{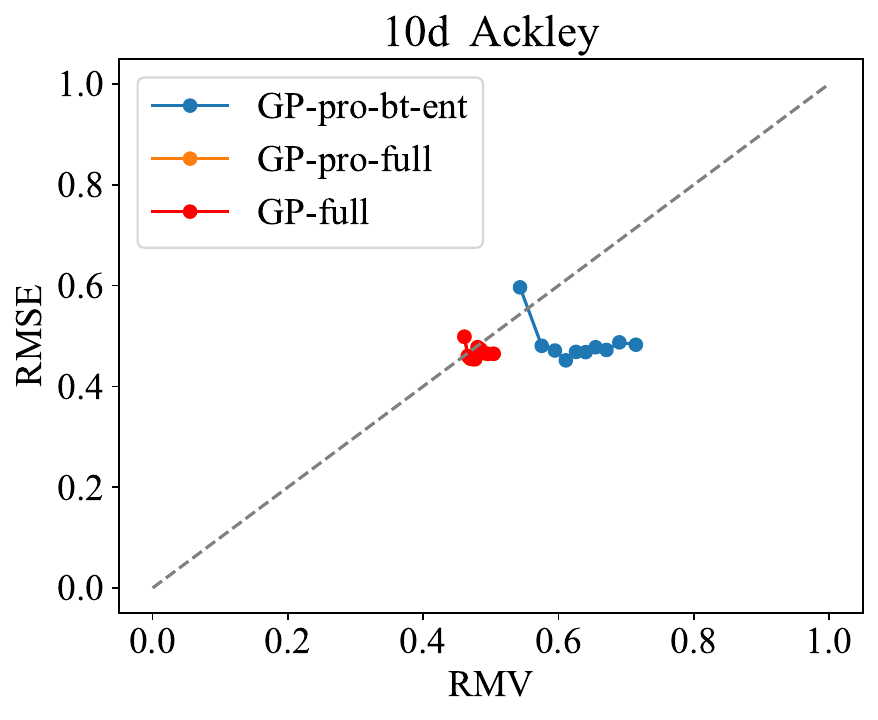}
  \caption{$10d$ Ackley}
  \label{fig:2}
\end{subfigure}\hfil 

\medskip
\medskip
\begin{subfigure}{0.35\textwidth}
   \includegraphics[width=\linewidth]{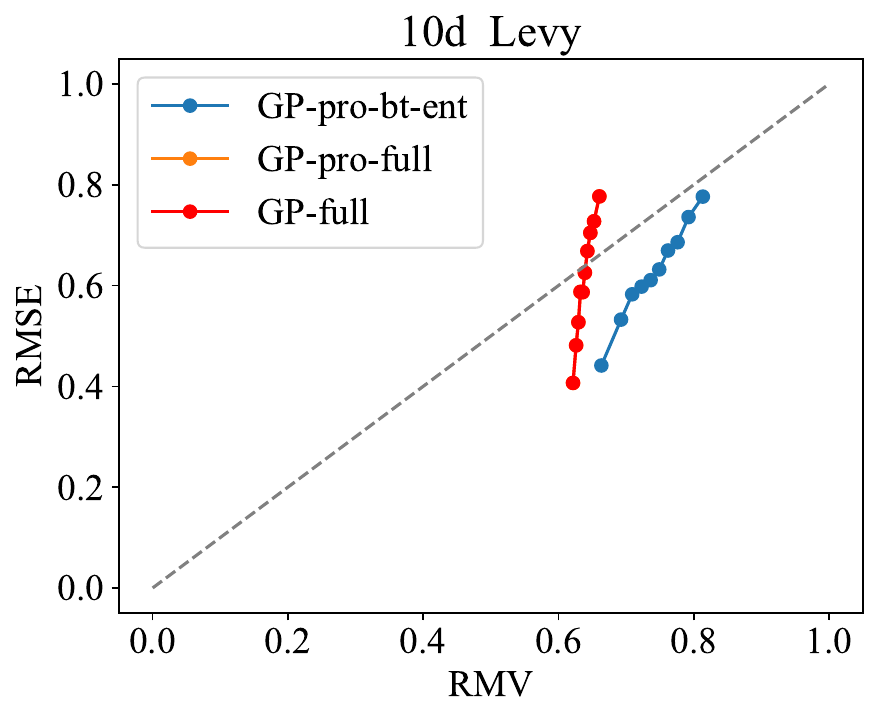}
    \caption{$10d$ Levy}
  \label{fig:4}
\end{subfigure}\hfil 
\begin{subfigure}{0.35\textwidth}
  \includegraphics[width=\linewidth]{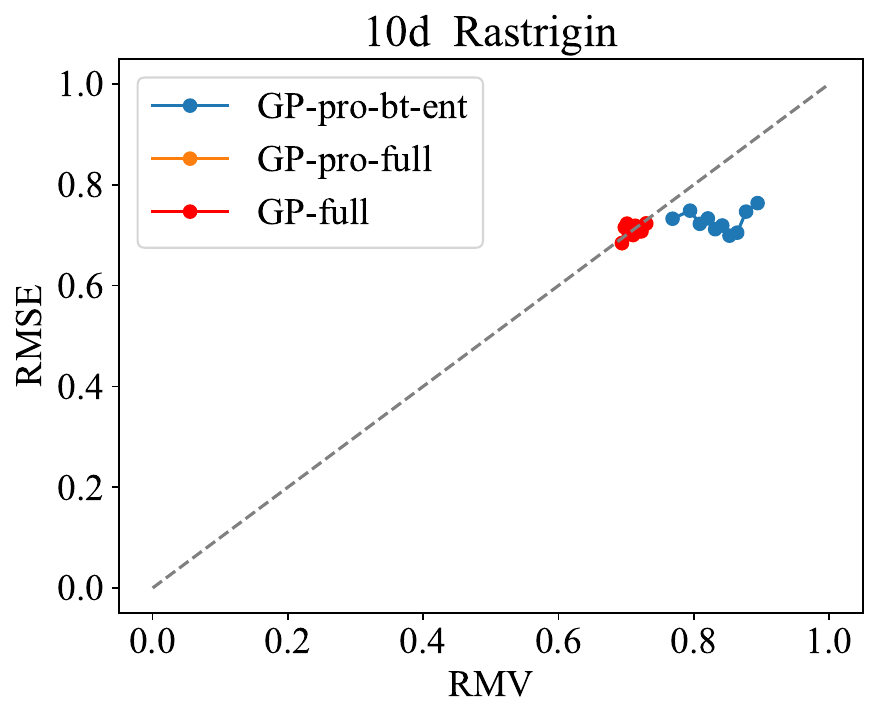}
   \caption{$10d$ Rastrigin}
  \label{fig:5}
\end{subfigure}\hfil 

\caption{Reliability diagrams for the GP-pro, GP-pro-full, and GP-full models on four synthetic functions. Well-calibrated models yield curves closer to the ideal diagonal line. The curves of GP-pro-full overlap with those of GP-full, as expected. For the $10d$ Levy function, GP-full also exhibits noticeable deviation from the diagonal line, which can be attributed to kernel misspecification and the high-dimensional, nonstationary nature of the function. In contrast, GP-pro shows consistently larger deviations across all functions, indicating systematic overestimation of posterior variances in the GP-pro models. 
}
\Description{ Four line graphs, labelled (a), (b), (c), and (d), show reliability diagrams for three types of Gaussian process models evaluated on the one-dimensional Ackley function, the 10-dimensional Ackley function, the 10-dimensional Levy function, and the 10-dimensional Rastrigin function, respectively.
In each graph, the horizontal axis represents the Root Mean Variance (RMV), ranging from 0 to 1 in increments of 0.2, and the vertical axis represents the Root Mean Square Error (RMSE), also ranging from 0 to 1 in increments of 0.2.
Each graph contains four lines: an ideal dashed diagonal line extending from the lower-left corner to the upper-right corner, and three model-specific lines corresponding to (i) a product-of-experts Gaussian process model, (ii) a product-of-experts Gaussian process model in which each local model was trained on the entire training dataset, and (iii) a single global Gaussian process model.
Across all four graphs, the single global Gaussian process model produces a line that is closest to the ideal diagonal, indicating better calibration. In contrast, the product-of-experts Gaussian process model produces a line that deviates noticeably from the diagonal, reflecting poorer calibration.
}
\label{fig:fig_reliability_analyse}
\end{figure}

For visual assessment, we use reliability diagrams \cite{Levi2020}, which plot RMSE$(j)$ against RMV$(j)$, as shown in Figure \ref{fig:fig_reliability_analyse}. 
The diagram is to show that for a perfectly calibrated model, the RMV and the observed RMSE should be approximately equal; hence, the plot should be close to the identity function (i.e., the ideal diagonal line). 

We also evaluate models using the negative log-likelihood (NLL) on the test set:
\begin{align*}
NLL = - \frac{1}{N} \sum_{i=1}^N \log \ p\left(y_i \mid \mathcal{N}(\hat{y}_i, \hat{\sigma}_i^2)\right)
\end{align*}
where $y_i$ is the true value, $\hat{y}_i$ and $\hat{\sigma}_i^2$ are the predicted mean and variance, and $N$ is the number of test points. 
Lower NLL indicates better overall performance.
NLL is used because it provides an overall assessment that captures both predictive accuracy and uncertainty quality \cite{Tran2020}.

In summary, RMSE evaluates point prediction accuracy, NLL jointly measures accuracy and uncertainty quality, and ENCE specifically assesses calibration.

Table \ref{tab:tab_rmse_nll_ence_analyse} and Figure \ref{fig:fig_reliability_analyse} compare GP-pro, GP-pro-full, and GP-full on four synthetic functions: $1d$ Ackley, $10d$ Ackley, $10d$ Levy, and $10d$ Rastrigin. 
The GP-pro model uses balltree assignment, entropy-based weighting, and 200 data points per expert. 
The GP-pro-full model is a diagnostic baseline in which each expert is trained on the full dataset. 
GP-full denotes a single global GP.
Table \ref{tab:tab_rmse_nll_ence_analyse} lists the performance metrics, Figure \ref{fig:fig_reliability_analyse} 
shows the reliability diagrams for the GP models, and Table  \ref{tab:tab_train_test_time_analyse}  lists the computational overheads incurred by the models. 

The results show that all models achieve similar RMSE values. 
However, GP-pro exhibit higher NLL and ENCE than GP-pro-full and GP-full. 
As expected, GP-pro-full matches GP-full, since all experts are trained on identical data and therefore produce identical posteriors. 
As shown in the reliability diagrams in Figure \ref{fig:fig_reliability_analyse}, the GP-pro-full and GP-full models exhibit curves that approach the ideal diagonal line, whereas the GP-pro models shows curves that clearly deviate from the diagonal line.
These results indicate overestimated posterior variances in the GP-pro model.

Although GP-pro-full provides well-calibrated uncertainty, it incurs significantly higher computational cost. 
Specifically, GP-pro scales as $\mathcal{O}(M \tilde{n}^3)$, GP-full as $\mathcal{O}(n^3)$, and GP-pro-full as $\mathcal{O}(M n^3)$, where $M$ is the number of experts, $\tilde{n}$ is the number of data points per expert, $n$ is the size of the training set and $ \tilde{n} \ll n$.

Overall, this analysis shows that GP-pro successfully reduces computational complexity but suffers from systematic overestimation of predictive variance, motivating the need for calibration.

\begin{table}
\caption{Average RMSE, NLL, and ENCE results for the GP-pro, GP-pro-full, and GP-full models on four synthetic functions. GP-pro-full and GP-full yield identical results, as expected, since GP-pro-full trains each expert on the entire training dataset, resulting in identical posteriors whose product recovers the global GP solution. GP-pro exhibits higher NLL and ENCE than GP-pro-full and GP-full, indicating variance overestimation in the GP-pro model.} \label{tab:tab_rmse_nll_ence_analyse}
\centering
\begin{scriptsize}
\begin{tabular}{@{}lrrccccccccc@{}}
    \toprule
    \textbf{Dataset} &
    \textbf{Size} &
    \textbf{Dim} &
      \multicolumn{3}{c}{\textbf{GP-pro-bt-ent}} &
      \multicolumn{3}{c}{\textbf{GP-pro-full}} &
      \multicolumn{3}{c}{\textbf{GP-full}}  \\
      \cmidrule(r){4-6}\cmidrule(r){7-9}\cmidrule(l){10-12}
      & & & {RMSE$\downarrow$} & {NLL$\downarrow$} & {ENCE$\downarrow$} & {RMSE$\downarrow$} & {NLL$\downarrow$} & {ENCE$\downarrow$} & {RMSE$\downarrow$} & {NLL$\downarrow$} & {ENCE$\downarrow$}  \\
      \midrule
    Ackley           & 500  &  1      & 0.300 & 0.267 & 0.296    & 0.290 & 0.204 & 0.252    & 0.290 & 0.204 & 0.252    \\ 
    Ackley           & 5000  &  10   &  0.487 & 0.775 & 0.245     & 0.468 & 0.663 & 0.057    & 0.468 & 0.663 & 0.057   \\ 
    Levy              & 5000  &  10   & 0.681 & 1.021 & 0.160    & 0.653 & 0.982 & 0.137    & 0.653 & 0.982 & 0.137  \\ 
    Rastrigin        & 5000  &  10   & 0.745 & 1.138 & 0.127    & 0.713 & 1.081 & 0.039    & 0.713 & 1.081 & 0.039  \\ 
    \bottomrule
    Average $\downarrow$ &   &                    & 0.553 & 0.800 & 0.207           & 0.531 & 0.732 & 0.121          & 0.531 & 0.732 & 0.121   \\
    \bottomrule
  \end{tabular}
  \end{scriptsize}
  \end{table}

\begin{table}
\caption{Average training and test times (in seconds) across four synthetic functions for the GP-pro, GP-pro-full, and GP-full models.}
 \label{tab:tab_train_test_time_analyse}
\centering
\begin{footnotesize}
\begin{tabular}{@{}lrrSSSS@{}}
    \toprule
    \textbf{Dataset} &
    \textbf{Size} &
    \textbf{Dim} &
      \multicolumn{1}{c}{\textbf{GP-pro}} &
      \multicolumn{1}{c}{\textbf{GP-pro-full}} &
      \multicolumn{1}{c}{\textbf{GP-full}}  \\
       \cmidrule(r){4-4}\cmidrule(r){5-5}\cmidrule(r){6-6}\cmidrule(l){7-7}
      & & & {training, test time} & {training, test time} & {training, test time}   \\
      \midrule
    Ackley           & 500  &  1       & {0.666, 0.018}  & {0.677, 0.021}  & {0.569, 0.013}       \\ 
    Ackley           & 5000  &  10   & {3.101, 1.425}  & {963.947, 13.745}  & {47.905, 0.885}    \\ 
    Levy              & 5000  &  10   & {3.380, 1.450}  & {982.690, 14.161}  & {46.986, 0.892}     \\ 
    Rastrigin       & 5000  &  10   & {3.119, 1.452}  & {972.641, 14.850}  & {50.469, 1.053}      \\ 
    \bottomrule
    Average $\downarrow$ &   &                    & {2.566, 1.086}  & {729.988, 10.694}  & {36.482, 0.711}        \\
    \bottomrule
\end{tabular}
  \end{footnotesize}
\end{table}

\section{Information-Based Uncertainty Calibration in Product-of-Experts Gaussian Process Models }
\label{sec:sec_calibration}

To calibrate uncertainty in GP-pro models, the central challenge is to quantify the appropriate magnitude of variance reduction. 
We propose an information-based variance calibration method that expresses this reduction in terms of conditional mutual information.

Let $n = \left|  \mathcal{D}  \right|$ and $n_m = \left|  \mathcal{D}^{(m)}  \right|$.   
For a local expert $GP^{(m)}$ and any input $\vx$, a natural measure of variance overestimation is the ratio $\frac{\sigma_{m, 1:n} (\vx) }{ \sigma_{m, 1:n_m} (\vx) }$, where $\sigma_{m,1:n}(\vx)$ denotes the posterior standard deviation conditioned on the full dataset  $ \mathcal{D} = \{ (\vx_i, y_i) \}_{1 \leqslant i \leqslant n}$, and $\sigma_{m,1:n_m}(\vx)$ denotes the posterior standard deviation conditioned on the local subset $ \mathcal{D}^{(m)} = \{ (\vx_i, y_i) \}_{1 \leqslant i \leqslant n_m}$. 
In a local  $GP^{(m)}$, the subset of data not assigned to $GP^{(m)}$ is denoted by $\mathcal{D} \setminus \mathcal{D}^{(m)} =  \{ (\vx_i, y_i) \}_{n_m + 1 \leqslant i \leqslant n} $. 
The  ratio $\frac{\sigma_{m, 1:n} (\vx) }{ \sigma_{m, 1:n_m} (\vx) }$ is related to $I \left( f(\vx) ; \vy_{m, n_m+1 : n } | \vy_{m, 1:n_m} \right)$, that is, the conditional mutual information with respect to $f(\vx)$ resulting from observations $\vy_{m, n_m+1 : n}$, given previous observations $\vy_{m, 1:n_m}$. The relation is shown below.

\begin{lemma} \label{lem:lemma1}
The ratio of the standard deviation of the posterior over $f(\vx)$, conditioned on $ \mathcal{D} $, to that conditioned on $ \mathcal{D}^{(m)}$ is:
\begin{align}
\frac{\sigma_{m, 1:n} (\vx) }{ \sigma_{m, 1:n_m} (\vx) } = \exp \Big( - I \left( f(\vx) ;  \vy_{m, n_m+1 : n } | \vy_{m, 1:n_m} \right) \Big)
\end{align}
where $\exp$ is the natural exponential function.
\end{lemma}

\begin{proof}
Since Gaussian process regression yields Gaussian posterior distributions, the conditional distribution $f(\vx) | \vy$ is Gaussian with variance given by the GP posterior variance. The differential entropy of a univariate Gaussian random variable with variance $\sigma^2$ is $\frac{1}{2} \log ( 2 \pi e \sigma^2  ) $. The lemma therefore follows from:
\begin{align*}
- I( f(\vx) ; \vy_{m,  n_m+1 : n} | \vy_{m, 1:n_m} ) &= - \left( H( f(\vx) | \vy_{m, 1:n_m} ) -   H( f(\vx) | \vy_{m, 1:n_m}, \vy_{m, n_m+1 : n} )\right)   \notag \\
	&= - \left( H( f(\vx) | \vy_{m, 1:n_m} ) -   H( f(\vx) | \vy_{m, 1:n} )\right)   \notag \\
	&= - \left( \frac{1}{2} \log ( 2 \pi e \sigma^2_{m, 1:n_m} (\vx) ) - \frac{1}{2} \log ( 2 \pi e \sigma^2_{m, 1:n} (\vx)  ) \right)  \notag \\
	&= \log \frac{ \sigma_{m, 1:n} (\vx) }{ \sigma_{m, 1:n_m} (\vx) }
\end{align*}
\end{proof}

The ratio $\frac{\sigma_{m, 1:n} (\vx) }{ \sigma_{m, 1:n_m} (\vx) }$ satisfies: 
\begin{align}
0 \ \  < \ \   \frac{\sigma_{m, 1:n} (\vx) }{ \sigma_{m, 1:n_m} (\vx) } = \exp \Big( - I \left( f(\vx) ;  \vy_{m, n_m+1 : n } | \vy_{m, 1:n_m} \right) \Big)  \ \  \leqslant  \ \ 1
\end{align}
where the upper bound 1 corresponds to no calibration. 
As proved by \citet{Dorard2012} and \citet{Srinivas2012} (see Section \ref{sec:sec_c2_ig}), 
$I \left( f(\vx) ;  \vy_{m, n_m+1 : n } | \vy_{m, 1:n_m} \right)$ can be calculated as the sum of the marginal conditional information gain of observations $\vy_{m, n_m+1 : n }$:
\begin{align}
I \left( f(\vx) ;  \vy_{m, n_m+1 : n } | \vy_{m, 1:n_m} \right) &=  \frac{1}{2} \sum_{i=n_m+1}^n \log \left( 1 + \frac{ \sigma_{m, i-1}^2(\vx) } { \ddot{\sigma}_m^2 }  \right)    \label{eq:eq_cc_ig_all} 
\end{align}

Evaluating $I \left( f(\vx) ;  \vy_{m, n_m+1 : n } | \vy_{m, 1:n_m} \right)$ exactly would require access to the full dataset, 
which would defeat the purpose of the distributed GP-pro framework whose objective is to avoid the cubic computational complexity of global GP inference.
To obtain a tractable approximation, we exploit the monotonicity and submodularity of information gain in GP. In particular, the cumulative information gain from the unseen observations can be lower-bounded by a scaled version of the largest marginal contribution.
Let $\Delta_i = \frac{1}{2} \log \left( 1 + \frac{ \sigma_{m, i-1}^2(\vx) } { \ddot{\sigma}_m^2 }  \right)$, then
\[
I \left( f(\vx) ;  \vy_{m, n_m+1 : n } | \vy_{m, 1:n_m} \right) = \sum_{i=n_m+1}^n \Delta_i
\]
In GP, information gain is submodular \citep{Srinivas2012}.
Submodularity implies diminishing returns, i.e. the marginal gains are nonincreasing as additional observations are incorporated:
\[
 \Delta_{n_m + 1}  \geqslant  \Delta_{n_m + 2}  \geqslant  \cdots  \geqslant  \Delta_{n} 
\]
Thus the largest marginal gain occurs at the first element, i.e., $\underset{i \in n_m+1 : n }{\max}  \Delta_i = \Delta_{n_m + 1}$.

Since all marginal gains are nonnegative, this yields the conservative lower bound:
\begin{align}
I \left( f(\vx) ;  \vy_{m, n_m+1 : n } | \vy_{m, 1:n_m} \right) \geqslant  \frac{e-1}{e} \  \frac{1}{2} \log \left( 1 + \frac{ \sigma_{m,n_m}^2(\vx) } { \ddot{\sigma}_m^2 }  \right)   \label{eq:eq_cc_ig_lb} 
\end{align}
Although Equation (\ref{eq:eq_c2_4}) is commonly derived in the context of greedy subset selection, we use it here only as a general inequality for monotone submodular functions to obtain a conservative lower bound on cumulative information gain. The derivation therefore does not assume that the dataset itself is generated by a greedy algorithm.

Substituting this bound into the variance ratio expression in Lemma \ref{lem:lemma1} yields:
\begin{align}
\frac{\sigma_{m, 1:n} (\vx) }{ \sigma_{m, 1:n_m} (\vx) } = \exp \Big( - I \left( f(\vx) ;  \vy_{m, n_m+1 : n } | \vy_{m, 1:n_m} \right) \Big)  \leqslant  \exp \ \left( -  \  \frac{e-1}{e}  \  \frac{1}{2} \log \left( 1 + \frac{ \sigma_{m,n_m}^2(\vx) } { \ddot{\sigma}_m^2 }  \right)  \right)  
\end{align}
Note that the posterior variance $\sigma_{m}^2(\vx)$ computed by the local expert $GP^{(m)}$ is exactly the variance conditioned on its local dataset 
$\mathcal{D}^{(m)}$, i.e.,  
$ \sigma_{m}^2(\vx) = \sigma_{m, n_m}^2 (\vx)$.
We therefore use the shorter notation $\sigma_{m}^2(\vx)$ in the following derivation.

We define the calibration ratio for any $ \vx \in \mathcal{X}$ at a local $GP^{(m)}$ as:
\begin{align}
 \alpha_{m}(\vx) = \exp \ \left( -  \  \frac{e-1}{e}  \  \frac{1}{2} \log \left( 1 + \frac{ \sigma_{m}^2(\vx) } { \ddot{\sigma}_m^2 }  \right)  \right)  \label{eq:eq_c4_cc_ratio}
\end{align}
The proposed calibration ratio does not reduce the posterior variance more aggressively than the reduction implied by the lower bound on conditional mutual information. As a result, the calibrated variance
\begin{align}
\hat{\sigma}_{m}^2(\vx) =  \alpha_{m} (\vx) \  \cdot \ \sigma_{m}^2(\vx)
\end{align}
remains a conservative estimate of the variance that would be obtained if the additional observations in $\mathcal{D} \setminus \mathcal{D}^{(m)}$ were incorporated.

The calibration ratio $\alpha_{m}(\vx)$ implicitly captures the potential contribution of the missing global data while remaining computable using only quantities available to the local expert, i.e.,  
the posterior variance $\sigma_{m}^2(\vx) $, which is already computed as part of the standard GP-pro predictive equations. Consequently, the calibration ratio $\alpha_{m}(\vx)$ can be evaluated efficiently at test time for arbitrary inputs $\vx$ without requiring access to other experts' datasets or additional training-time computation. In this sense, the posterior variance predicted by the local expert serves as a compact summary of the information already incorporated by $GP^{(m)}$, while the bound provides a conservative estimate of the additional information that could be contributed by the missing observations.

The factor $\frac{e-1}{e}$ introduces a controlled degree of conservativeness.
In the context of uncertainty calibration, this conservativeness is intentional: an overly aggressive reduction of posterior variance can lead to 
numerically unstable predictions. The proposed bound therefore acts as a safeguard against over-calibration.
While tighter, data-dependent bounds on conditional information gain may further improve calibration accuracy, exploring such alternatives would require additional assumptions or adaptive estimation strategies and is left for future work. Empirically, the current bound already yields consistent improvements in NLL and ENCE across diverse datasets, as reported in Section \ref{sec:sec_experiments}.

In the proposed calibration rule, the information gain term corresponds to the largest marginal information gain contributed by a single observation, as discussed above. In practice, we normalise this quantity so that its magnitude does not exceed one. This choice is consistent with the analysis of  \citet{Srinivas2012}, which shows that under standard assumptions with unit prior variance, the marginal information gain contributed by a single observation is bounded by a constant of order one for commonly used kernels such as the Mat\'{e}rn kernel. Since our calibration uses only the largest marginal information gain term, adopting a normalised scale with an upper bound of one provides a convenient and stable numerical reference without affecting the functional form of the calibration rule. 
Although the derivation relies on standard assumptions regarding normalised information gain and the reference to Mat\'{e}rn kernels, the calibration itself depends only on posterior variances and noise levels that are already available at test time, and does not otherwise rely on kernel-specific properties. We therefore expect the method to extend to other commonly used stationary kernels, although a formal analysis of such extensions is left for future work.

\paragraph{Aggregating posterior predictions.}  

During test time, at a candidate input $\vx'$, each local $GP^{(m)}$ independently computes the posterior predictive mean $\mu_{m} (\vx')$ and variance $\sigma_{m}^2 (\vx')$ using the standard GP-pro predictive equations.
For completeness, we restate the standard GP-pro predictive equations below:
\begin{align*}
\mu_{m}(\vx') &=   \tilde{\mu}^{(m)}(\vx')  +  \vk^{(m)}(\vx')^{\top} \left(  \mK^{(m)} + \ddot{\sigma}_{m}^2 \ \mI  \right)^{-1} \left( \vy^{(m)} -  \tilde{\vmu}^{(m)}  \right) \\
\sigma_{m}^2(\vx') &= k^{(m)}(\vx', \vx') -  \vk^{(m)}(\vx')^{\top}  \left(  \mK^{(m)} + \ddot{\sigma}_{m}^2 \ \mI \right)^{-1} \vk^{(m)}(\vx') 
\end{align*}

With the calibrated variance $\hat{\sigma}_{m}^2(\vx') =  \alpha_{m} (\vx')   \cdot  \sigma_{m}^2(\vx')$, the aggregated predictive mean  $\mu$ and variance $\sigma^2$ at the candidate $\vx'$  are derived as:
\begin{align}
\mu(\vx') &= \sigma^2(\vx')  \sum_{m=1}^M \left( w_{m}(\vx') \  \mu_{m}(\vx') \ \frac{1}{ \hat{\sigma}_{m}^2(\vx')} \right)   \label{eq:eq_c4_agg_mu} \\
\sigma^2(\vx') &= \left( \sum_{m=1}^M \left( w_{m}(\vx') \  \frac{1}{ \hat{\sigma}_{m}^2(\vx')}  \right)  \right)^{-1}  \label{eq:eq_c4_agg_sigma}
\end{align}
Although Equations (\ref{eq:eq_c4_agg_mu}) and (\ref{eq:eq_c4_agg_sigma}) retain the algebraic structure of the standard GP-pro aggregation, the predictive uncertainty entering the aggregation is fundamentally different. In the proposed model, each expert contributes a calibrated posterior variance $\hat{\sigma}_{m}^2(\vx')$ obtained from the information-based shrinkage rule, rather than the raw posterior variance produced by the local GP.
This modification corrects the systematic variance inflation that arises when experts are trained on disjoint subsets of data. Importantly, the aggregation rule itself remains unchanged, which preserves the computational efficiency and scalability of the original GP-pro framework while improving the reliability of its uncertainty estimates.

\begin{figure}
\centering
\includegraphics[width=1.0\textwidth]{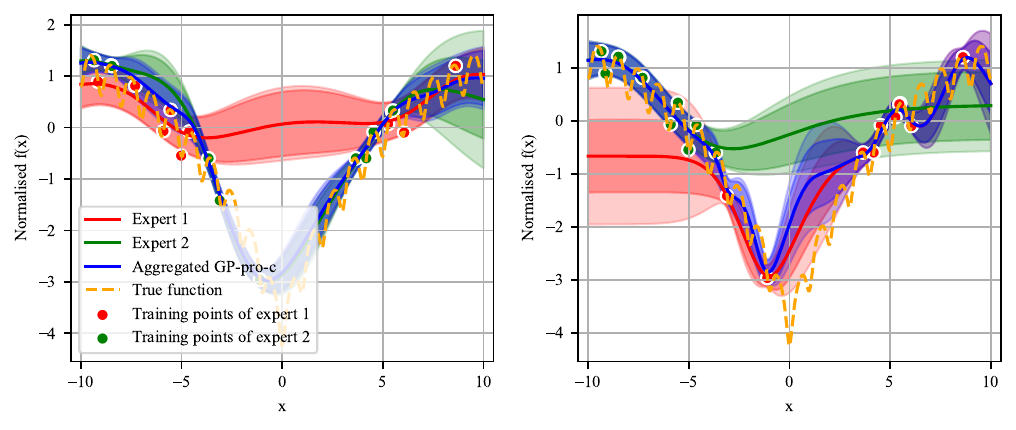}
\caption{Illustration of posterior predictions before and after applying the calibration ratio to GP-pro-c models comprising two local experts for the one-dimensional Ackley function. The GP-pro-c model on the left uses the random assignment method, whereas the model on the right uses balltree assignment. Both GP-pro-c models employ an entropy-weighted aggregation scheme. The posterior predictions of the two local GP models are represented by the green and red curves, with the corresponding light green and light red shaded regions indicating the standard deviation at each input value $x$. The blue curve and shaded region represent the aggregated posterior prediction. The darker green, red, and blue shaded regions indicate the reduced standard deviations obtained after applying the calibration ratio.}
\Description{Two line graphs, shown side by side, illustrate posterior predictions for the one-dimensional Ackley function. In both graphs, the y-axis shows function values ranging from -4 to 2, and the x-axis shows input values from -10 to 10 in increments of 2.
Each graph contains three prediction curves and their corresponding uncertainty intervals. The first and second curves represent the posterior mean predictions of two local Gaussian process models, together with their uncertainty intervals before and after applying the calibration ratio. The third curve represents the aggregated posterior prediction obtained by combining the outputs of the two local Gaussian process models, along with its corresponding uncertainty interval. The darker shaded regions indicate the calibrated uncertainties after applying the calibration ratio, while the lighter shaded regions indicate the original uncertainties before calibration.}
\label{fig:fig_c3_gp_gpoe_c}
\end{figure}

Figure \ref{fig:fig_c3_gp_gpoe_c} illustrates the effect of calibration on posterior predictions for the Ackley function using GP-pro-c models comprising two local experts. 
The calibrated model reduces predictive variance while maintaining accurate mean predictions.

Table \ref{tab:tab_gp_models_time} summarises training and test-time complexity for global GP, vanilla GP-pro and GP-pro-c with variance calibration.
GP-pro-c retains efficient product-of-experts aggregation with calibrated posterior variances. Each local GP requires $\mathcal{O}(n_m^3)$ training, and test-time mean and variance computations are $\mathcal{O}(n_m)$ and $\mathcal{O}(n_m^2)$ per expert, respectively. The calibration ratio $\alpha_{m}(\vx)$ is computed directly from the already-available posterior variance, introducing negligible extra computation, and preserving the efficiency advantage over a global GP trained on all $n$ points.

\begin{table}
\centering
\caption{Computational complexity comparison.}
\begin{scriptsize}
\begin{tabular}[t]{llllll}
\toprule
\textbf{Method}   &  \textbf{Training}           &  \textbf{Test-time}   &  \textbf{Test-time}   & \textbf{Notes}  \\
	                   &   \textbf{complexity}     &   \textbf{mean}        &  \textbf{variance}    & \\
\midrule

\textbf{Global GP}                                                                 &  $\mathcal{O}(n^3)$     &   $\mathcal{O}(n)$          & $\mathcal{O}(n^2)$  &  Full dataset, accurate variance       \\
\textbf{Vanilla GP-pro ($M$ experts, $n_m$ points each) }   &  $\mathcal{O}(M n_m^3)$      &   $\mathcal{O}(M n_m)$      & $\mathcal{O}(M n_m^2)$    & Distributed experts, standard PoE  \\
\textbf{GP-pro-c (proposed) }                            &  $\mathcal{O}(M n_m^3)$      &   $\mathcal{O}(M n_m)$      & $\mathcal{O}(M n_m^2)$    &   Adds calibrated variance via $\alpha_{m}(\vx)$ with negligible cost    \\
\bottomrule
\label{tab:tab_gp_models_time}
\end{tabular}
\end{scriptsize}
\end{table}

\section{Experiments}
\label{sec:sec_experiments}

This section presents the empirical evaluation of the GP-pro-c model.
The implementation is publicly available at \url{https://github.com/yhong123/gp-pro-c}.

All experiments were conducted using GPyTorch \cite{Gardner2018} for GP regression. 
Prior to model fitting, input features were rescaled to $[0,1]^d$, and outputs were standardised. 
Each GP was parameterised with a Mat\'{e}rn-$5/2$ covariance function with automatic relevance determination (ARD) and a constant mean function. 
Hyperparameters were learned by maximising the log marginal likelihood.

Experiments were conducted on four synthetic functions with known global minima: the $10d$ Ackley (5000 data points), $10d$ Levy (5000 data points), $10d$ Rastrigin  (5000 data points) and $50d$ Rosenbrock  (50000 data points) functions.
All function values were corrupted with additive Gaussian noise $\mathcal{N}(0, 0.25)$.

We also consider six real-world regression datasets: Concrete ($8d$ feature space, 1030 data points), Airfoil ($5d$, 1503 data points), Power ($4d$, 9568 data points), Parkinson ($20d$, 5875 data points), Kinetic ($8d$, 40000 data points) and Protein ($9d$, 45730 data points).

Each experiment on a function or dataset was repeated ten times with different random seeds.
In each run, the data points in a function or dataset were partitioned such that $50\%$ were used as training points and $50\%$ as test points.

The performance of each GP model was measured using  three performance metrics: root mean squared error (RMSE), expected normalised calibration error (ENCE) and negative log-likelihood (NLL).
Details of these metrics are provided in Section \ref{sec:sec_analyse}.

All experiments were conducted on a single Apple M2 processor (3.49 GHz) with 8 GB of memory.

\subsection{Comparing Data Assignment Methods and Weighting Schemes}

The first set of experiments examines the impact of the proposed information-based calibration across variants of GP-pro-c models using different data assignment methods and weighting schemes. 
Specifically, we consider random, $k$-means, and balltree assignment methods, combined with entropy-, variance-, and uniform-weighted schemes, with 200 data points per expert. 
These models are compared against corresponding GP-pro variants without calibration.

Table \ref{tab:tab_c3_gp_models} summarises the model configurations and their labels. 
Each model name (e.g., rd-ent-c) encodes its design choices: 
`rd',  `kx' , and `bt' denote random, $k$-means, and balltree assignment methods, respectively, while `ent', `var', and `uni' denote entropy-, variance-, and uniform-weighted schemes. 
The suffix `c'  indicates the use of information-based variance calibration.

\begin{table}
\centering
\caption{Summary of the models used in the experiments for the comparison and evaluation of the GP-pro-c and GP-pro models.}
\begin{scriptsize}
\begin{tabular}[t]{llllll}
\toprule
Model   &  Data assignment method    &  Weighting scheme   &  Uncertainty calibration   &  \\
\midrule

bt-ent                &  balltree     &   entropy          & no       \\
bt-var                &  balltree      &   variance           & no      \\
bt-uni               &  balltree    &   uniform          & no       \\

kx-ent                 &  $k$-means     &   entropy          & no       \\
kx-var                &  $k$-means      &   variance           & no      \\
kx-uni                  &  $k$-means     &   uniform          & no       \\

rd-ent              &  random     &   entropy         & no        \\
rd-var               &  random     &   variance         & no        \\
rd-uni                &  random     &   uniform         & no        \\

bt-ent-c             &  balltree     &   entropy         & yes        \\
bt-var-c              &  balltree     &   variance         & yes        \\
bt-uni-c               &  balltree     &   uniform         & yes        \\

kx-ent-c                &  $k$-means     &   entropy          & yes       \\
kx-var-c                 &  $k$-means      &   variance           & yes      \\
kx-uni-c               &  $k$-means     &   uniform          & yes       \\

rd-ent-c             &  random     &   entropy         & yes        \\
rd-var-c               &  random     &   variance         & yes        \\
rd-uni-c              &  random     &   uniform         & yes        \\

\bottomrule
\label{tab:tab_c3_gp_models}
\end{tabular}
\end{scriptsize}
\end{table}

Tables \ref{tab:tab_rmse}-\ref{tab:tab_ence} report the performance metrics for each GP model. 
Overall, across all data assignment methods and weighting schemes, GP-pro-c achieves RMSE values comparable to those of GP-pro. 
However, GP-pro-c consistently attains lower NLL and ENCE scores, indicating improved uncertainty calibration. 
These results suggest that the proposed information-based calibration effectively mitigates overestimation of predictive variance in local experts, regardless of the assignment or weighting method.

\begin{table}
\caption{Average RMSE values across four synthetic functions and six benchmark datasets using balltree, $k$-means, and random assignment methods, as well as entropy-, variance-, and uniform-weighted schemes for the GP-pro-c and GP-pro models. GP-pro-c models with information-based calibration achieve RMSE values comparable to those of GP-pro models, indicating that information-based calibration improves uncertainty quantification without affecting predictive accuracy.}
 \label{tab:tab_rmse}
\begin{scriptsize}

\begin{subtable}{1\textwidth}
\sisetup{table-format=-1.2}   
\centering
   \begin{tabular}{@{}lrrSSSSSSSSS@{}}
      \toprule
      \textbf{Dataset} & \textbf{Size} & \textbf{Dim} & \textbf{bt-ent-c} & \textbf{bt-var-c} & \textbf{bt-uni-c} & \textbf{kx-ent-c} & \textbf{kx-var-c} & \textbf{kx-uni-c} & \textbf{rd-ent-c} & \textbf{rd-var-c} & \textbf{rd-uni-c} \\ 
      \midrule
      Ackley & 5000 & 10          & 0.487  & 0.490  & 0.490  & 0.497  & 0.502  & 0.500  & 0.471  & 0.471  & 0.471  \\ 
      Levy & 5000 & 10             & 0.681  & 0.685  & 0.697  & 0.679  & 0.681  & 0.686  & 0.670  & 0.670  & 0.670  \\ 
      Rastrigin & 5000 & 10        & 0.745  & 0.745  & 0.755  & 0.746  & 0.747  & 0.751  & 0.737  & 0.736  & 0.738  \\ 
      Rosenbrock & 50000 & 50    & 0.610  & 0.612  & 0.624  & 0.691  & 0.687  & 0.690  & 0.586  & 0.584  & 0.586   \\
      Concrete & 1030 &  8          & 0.336  & 0.340  & 0.338  & 0.331  & 0.335  & 0.338  & 0.352  & 0.353  & 0.352  \\ 
      Airfoil & 1503 &  5                & 0.283  & 0.291  & 0.295  & 0.294  & 0.296  & 0.310  & 0.314  & 0.316  & 0.317 \\ 
      Parkinsons & 5875 &  20   & 0.027  & 0.031  & 0.094  & 0.042  & 0.063  & 0.172  & 0.042  & 0.047  & 0.048   \\ 
      Power & 9568 &  4             & 0.249  & 0.263  & 0.312  & 0.252  & 0.269  & 0.488  & 0.245  & 0.245  & 0.245  \\ 
      Kin40K & 40000 &  8         & 0.384  & 0.457  & 0.500  & 0.365  & 0.452  & 0.503  & 0.466  & 0.474  & 0.486   \\ 
      Protein & 45730 &  9          & 0.664  & 0.688  & 0.888  & 0.725  & 0.704  & 1.021  & 0.813  & 0.815  & 0.814  \\ 
      \bottomrule
      Average RMSE $\downarrow$ &   &  & 0.447  & 0.460  & 0.499  & 0.462  & 0.474  & 0.546  & 0.470  & 0.471  & 0.473  \\
       \bottomrule
   \end{tabular}
   \caption{GP-pro-c models}\label{tab:sub_first}
\end{subtable}

\bigskip
\begin{subtable}{1\textwidth}
\sisetup{table-format=-1.2}   
\centering
   \begin{tabular}{@{}lrrSSSSSSSSS@{}}
      \toprule
      \textbf{Dataset} & \textbf{Size} & \textbf{Dim} & \textbf{bt-ent} & \textbf{bt-var} & \textbf{bt-uni} & \textbf{kx-ent} & \textbf{kx-var} & \textbf{kx-uni} & \textbf{rd-ent} & \textbf{rd-var} & \textbf{rd-uni} \\ 
      \midrule
      Ackley & 5000 & 10        & 0.487  & 0.490  & 0.489  & 0.497  & 0.502  & 0.500  & 0.471  & 0.471  & 0.471   \\ 
      Levy & 5000 & 10            & 0.681  & 0.684  & 0.695  & 0.679  & 0.680  & 0.685  & 0.670  & 0.670  & 0.670  \\ 
      Rastrigin & 5000 & 10      & 0.745  & 0.745  & 0.753  & 0.746  & 0.747  & 0.750  & 0.737  & 0.736  & 0.738   \\ 
      Rosenbrock & 50000 & 50 & 0.610  & 0.612  & 0.623  & 0.693  & 0.689  & 0.692  & 0.587  & 0.585  & 0.587   \\
      Concrete & 1030 &  8      & 0.336  & 0.340  & 0.338  & 0.331  & 0.335  & 0.337  & 0.352  & 0.353  & 0.352   \\ 
      Airfoil & 1503 &  5            & 0.286  & 0.294  & 0.297  & 0.293  & 0.295  & 0.307  & 0.314  & 0.316  & 0.317   \\ 
      Parkinsons & 5875 &  20  & 0.027  & 0.031  & 0.073  & 0.040  & 0.059  & 0.140  & 0.042  & 0.047  & 0.048   \\ 
      Power & 9568 &  4           & 0.249  & 0.262  & 0.309  & 0.252  & 0.268  & 0.484  & 0.245  & 0.245  & 0.245   \\ 
      Kin40K & 40000 &  8       & 0.375  & 0.446  & 0.486  & 0.352  & 0.436  & 0.484  & 0.464  & 0.472  & 0.484   \\ 
      Protein & 45730 &  9       & 0.663  & 0.687  & 0.886  & 0.727  & 0.703  & 1.027  & 0.813  & 0.815  & 0.814   \\ 
      \bottomrule
      Average RMSE $\downarrow$ &   &  & 0.446  & 0.459  & 0.495  & 0.461  & 0.471  & 0.541  & 0.470  & 0.471  & 0.473   \\
       \bottomrule
   \end{tabular}
   \caption{GP-pro models}\label{tab:sub_first}
\end{subtable}

 \end{scriptsize}
\end{table}

\begin{table}
\caption{Average NLL values across four synthetic functions and six benchmark datasets using balltree, $k$-means, and random assignment methods, as well as entropy-, variance-, and uniform-weighted schemes for the GP-pro-c and GP-pro models. GP-pro-c models with information-based calibration achieve lower NLL values than GP-pro models, demonstrating that information-based calibration improves uncertainty quantification without affecting predictive accuracy.} \label{tab:tab_nll}
\begin{scriptsize}

\begin{subtable}{1\textwidth}
\sisetup{table-format=-1.2}   
\centering
   \begin{tabular}{@{}lrrSSSSSSSSS@{}}
      \toprule
      \textbf{Dataset} & \textbf{Size} & \textbf{Dim} & \textbf{bt-ent-c} & \textbf{bt-var-c} & \textbf{bt-uni-c} & \textbf{kx-ent-c} & \textbf{kx-var-c} & \textbf{kx-uni-c} & \textbf{rd-ent-c} & \textbf{rd-var-c} & \textbf{rd-uni-c} \\ 
      \midrule
      Ackley & 5000 & 10             & 0.742  & 0.755  & 0.770  & 0.769  & 0.786  & 0.800  & 0.691  & 0.691  & 0.692   \\ 
      Levy & 5000 & 10                & 1.012  & 1.021  & 1.043  & 1.015  & 1.021  & 1.034  & 0.983  & 0.982  & 0.982  \\ 
      Rastrigin & 5000 & 10           & 1.124  & 1.125  & 1.141  & 1.127  & 1.128  & 1.137  & 1.118  & 1.116  & 1.118  \\ 
      Rosenbrock & 50000 & 50   & 0.918  & 0.923  & 0.944  & 1.038  & 1.031  & 1.035  & 0.874  & 0.871  & 0.874   \\
      Concrete & 1030 &  8           & 0.287  & 0.304  & 0.316  & 0.253  & 0.269  & 0.304  & 0.331  & 0.337  & 0.334  \\ 
      Airfoil & 1503 &  5               & 0.093  & 0.133  & 0.189  & 0.045  & 0.072  & 0.137  & 0.217  & 0.228  & 0.234  \\ 
      Parkinsons & 5875 &  20    & -1.374  & -1.315  & -0.730  & -1.316  & -1.117  & -0.331  & -1.114  & -1.063  & -1.057  \\ 
      Power & 9568 &  4             & 0.034  & 0.082  & 0.232  & 0.038  & 0.103  & 0.784  & 0.015  & 0.015  & 0.015  \\ 
      Kin40K & 40000 &  8         & 0.466  & 0.626  & 0.724  & 0.424  & 0.622  & 0.737  & 0.647  & 0.662  & 0.683   \\ 
      Protein & 45730 &  9         & 0.932  & 0.989  & 1.306  & 1.113  & 1.018  & 1.491  & 1.220  & 1.222  & 1.221  \\ 
      \bottomrule
      Average NLL $\downarrow$ &   & & 0.423  & 0.464  & 0.593  & 0.451  & 0.493  & 0.713  & 0.498  & 0.506  & 0.510  \\
       \bottomrule
   \end{tabular}
   \caption{GP-pro-c models}\label{tab:sub_first}
\end{subtable}

\bigskip
\begin{subtable}{1\textwidth}
\sisetup{table-format=-1.2}   
\centering
   \begin{tabular}{@{}lrrSSSSSSSSS@{}}
      \toprule
      \textbf{Dataset} & \textbf{Size} & \textbf{Dim} & \textbf{bt-ent} & \textbf{bt-var} & \textbf{bt-uni} & \textbf{kx-ent} & \textbf{kx-var} & \textbf{kx-uni} & \textbf{rd-ent} & \textbf{rd-var} & \textbf{rd-uni} \\ 
      \midrule
      Ackley & 5000 & 10           & 0.775  & 0.792  & 0.812  & 0.804  & 0.825  & 0.845  & 0.707  & 0.709  & 0.710  \\ 
      Levy & 5000  & 10            & 1.021  & 1.031  & 1.054  & 1.031  & 1.041  & 1.056  & 0.980  & 0.979  & 0.980  \\ 
      Rastrigin & 5000 & 10      & 1.138  & 1.141  & 1.159  & 1.147  & 1.151  & 1.164  & 1.113  & 1.113  & 1.116   \\ 
      Rosenbrock & 50000 & 50  & 0.929  & 0.934  & 0.953  & 1.028  & 1.023  & 1.027  & 0.881  & 0.879  & 0.882   \\
      Concrete & 1030 &  8        & 0.287  & 0.305  & 0.317  & 0.254  & 0.271  & 0.307  & 0.331  & 0.337  & 0.333  \\ 
      Airfoil & 1503 &  5             & 0.099  & 0.139  & 0.196  & 0.047  & 0.074  & 0.141  & 0.219  & 0.230  & 0.236  \\ 
      Parkinsons & 5875 &  20  & -1.352  & -1.293  & -0.748  & -1.304  & -1.113  & -0.410  & -1.113  & -1.062  & -1.056   \\ 
      Power & 9568 &  4             & 0.034  & 0.082  & 0.222  & 0.038  & 0.103  & 0.747  & 0.015  & 0.015  & 0.015  \\ 
      Kin40K & 40000 &  8         & 0.471  & 0.631  & 0.730  & 0.431  & 0.629  & 0.746  & 0.661  & 0.676  & 0.697   \\ 
      Protein & 45730 &  9         & 0.930  & 0.988  & 1.305  & 1.106  & 1.013  & 1.487  & 1.219  & 1.221  & 1.220  \\ 
      \bottomrule
      Average NLL $\downarrow$ &   &  & 0.433  & 0.475  & 0.600  & 0.458  & 0.502  & 0.711  & 0.501  & 0.510  & 0.513   \\
       \bottomrule
   \end{tabular}
   \caption{GP-pro models}\label{tab:sub_first}
\end{subtable}

 \end{scriptsize}
\end{table}

\begin{table}
\caption{Average ENCE values across four synthetic functions and six benchmark datasets using balltree, $k$-means, and random assignment methods, as well as entropy-, variance-, and uniform-weighted schemes for the GP-pro-c and GP-pro models. GP-pro-c models with information-based calibration achieve lower ENCE values than GP-pro models, demonstrating improved uncertainty quantification.} \label{tab:tab_ence}
\begin{scriptsize}

\begin{subtable}{1\textwidth}
\sisetup{table-format=-1.2}   
\centering
   \begin{tabular}{@{}lrrSSSSSSSSS@{}}
      \toprule
      \textbf{Dataset} & \textbf{Size} & \textbf{Dim} & \textbf{bt-ent-c} & \textbf{bt-var-c} & \textbf{bt-uni-c} & \textbf{kx-ent-c} & \textbf{kx-var-c} & \textbf{kx-uni-c} & \textbf{rd-ent-c} & \textbf{rd-var-c} & \textbf{rd-uni-c} \\ 
      \midrule
      Ackley & 5000 & 10            & 0.190  & 0.207  & 0.236  & 0.206  & 0.224  & 0.259  & 0.120  & 0.122  & 0.126   \\ 
      Levy & 5000 & 10              & 0.103  & 0.114  & 0.113  & 0.159  & 0.174  & 0.186  & 0.098  & 0.099  & 0.100  \\ 
      Rastrigin & 5000 & 10       & 0.060  & 0.064  & 0.084  & 0.065  & 0.070  & 0.089  & 0.083  & 0.079  & 0.076  \\ 
      Rosenbrock & 50000 & 50  & 0.116  & 0.104  & 0.081  & 0.188  & 0.181  & 0.174  & 0.203  & 0.206  & 0.205   \\
      Concrete & 1030 &  8        & 0.162  & 0.151  & 0.178  & 0.157  & 0.174  & 0.217  & 0.167  & 0.165  & 0.162   \\ 
      Airfoil & 1503 &  5             & 0.190  & 0.187  & 0.153  & 0.140  & 0.178  & 0.252  & 0.151  & 0.157  & 0.162   \\ 
      Parkinsons & 5875 &  20  & 0.750  & 0.735  & 0.543  & 0.635  & 0.528  & 0.233  & 0.689  & 0.678  & 0.678   \\ 
      Power & 9568 &  4            & 0.069  & 0.081  & 0.092  & 0.052  & 0.103  & 0.312  & 0.059  & 0.059  & 0.057   \\ 
      Kin40K & 40000 &  8         & 0.306  & 0.277  & 0.268  & 0.372  & 0.331  & 0.311  & 0.178  & 0.175  & 0.169  \\ 
      Protein & 45730 &  9         & 0.097  & 0.076  & 0.072  & 0.299  & 0.095  & 0.210  & 0.085  & 0.082  & 0.081   \\ 
      \bottomrule
      Average ENCE $\downarrow$ &   &  & 0.204  & 0.200  & 0.182  & 0.227  & 0.206  & 0.224  & 0.183  & 0.182  & 0.182  \\
       \bottomrule
   \end{tabular}
   \caption{GP-pro-c models}\label{tab:sub_first}
\end{subtable}

\bigskip
\begin{subtable}{1\textwidth}
\sisetup{table-format=-1.2}   
\centering
   \begin{tabular}{@{}lrrSSSSSSSSS@{}}
      \toprule
      \textbf{Dataset} & \textbf{Size} & \textbf{Dim} & \textbf{bt-ent} & \textbf{bt-var} & \textbf{bt-uni} & \textbf{kx-ent} & \textbf{kx-var} & \textbf{kx-uni} & \textbf{rd-ent} & \textbf{rd-var} & \textbf{rd-uni} \\ 
      \midrule
      Ackley & 5000 & 10          & 0.245  & 0.263  & 0.292  & 0.260  & 0.279  & 0.314  & 0.166  & 0.172  & 0.175  \\ 
      Levy & 5000 & 10             & 0.160  & 0.172  & 0.179  & 0.214  & 0.232  & 0.252  & 0.103  & 0.104  & 0.107   \\ 
      Rastrigin & 5000 & 10       & 0.127  & 0.138  & 0.153  & 0.153  & 0.164  & 0.182  & 0.064  & 0.065  & 0.070  \\ 
      Rosenbrock  & 50000 & 50  & 0.167  & 0.163  & 0.145  & 0.191  & 0.187  & 0.180  & 0.240  & 0.242  & 0.240   \\
      Concrete & 1030 &  8       & 0.166  & 0.150  & 0.176  & 0.155  & 0.178  & 0.230  & 0.166  & 0.164  & 0.161   \\ 
      Airfoil & 1503 &  5             & 0.191  & 0.188  & 0.155  & 0.143  & 0.183  & 0.263  & 0.159  & 0.163  & 0.166  \\ 
      Parkinsons & 5875 &  20  & 0.759  & 0.747  & 0.657  & 0.662  & 0.570  & 0.400  & 0.690  & 0.678  & 0.678  \\ 
      Power & 9568 &  4            & 0.071  & 0.085  & 0.098  & 0.059  & 0.122  & 0.286  & 0.059  & 0.059  & 0.057   \\ 
      Kin40K & 40000 &  8         & 0.348  & 0.326  & 0.323  & 0.420  & 0.391  & 0.377  & 0.218  & 0.217  & 0.211  \\ 
      Protein & 45730 &  9         & 0.086  & 0.074  & 0.072  & 0.281  & 0.086  & 0.183  & 0.056  & 0.053  & 0.057  \\ 
      \bottomrule
      Average ENCE $\downarrow$ &   &  & 0.232  & 0.231  & 0.225  & 0.254  & 0.239  & 0.267  & 0.192  & 0.192  & 0.192  \\
       \bottomrule
   \end{tabular}
   \caption{GP-pro models}\label{tab:sub_first}
\end{subtable}

 \end{scriptsize}
\end{table}

Among GP-pro-c variants, the combination of balltree assignment and entropy-based weighting performs best, achieving the lowest RMSE, NLL, and ENCE. 
This observation is consistent with prior findings by \citet{Cao2014,Cao2015} for GP-pro models without calibration.

The balltree and $k$-means assignment methods partition training data points based on similarity measures. Data points with close similarities are grouped and used to fit individual local GP models. 
These lead to better modelling of the input domain and yield better performance results in comparison to the random assignment method.
Between the balltree and $k$-means, the balltree method is more beneficial because the allocation is done so that each local GP model has a nearly equivalent number of training data points.
In contrast, by using the $k$-means assignment, each local GP model can be allocated a different number of data points. 
This may result in an  undesirable condition where some local GPs have few data points while others are allocated a large number of data points, which will incur higher computational overheads.

The choice of assignment method influences predictive accuracy and uncertainty calibration in distinct ways. Random assignment constructs weakly correlated local GP experts, leading to conservative predictive variances that tend to match empirical errors and therefore achieve lower ENCE. However, the lack of spatial locality degrades posterior mean estimates, resulting in higher RMSE and NLL. In contrast, balltree and $k$-means assignments preserve locality and improve predictive accuracy, but induce greater information redundancy across experts. This redundancy amplifies variance misestimation in the uncalibrated GP-pro model, making uncertainty calibration more challenging and yielding higher ENCE despite superior predictive performance. These observations highlight that ENCE should be interpreted jointly with RMSE and NLL rather than in isolation

It is worth noting that lower absolute ENCE values do not necessarily imply larger uncertainty corrections. While random assignment achieves consistently lower ENCE due to weaker correlations among local experts, locality-aware assignments such as balltree and $k$-means introduce greater information redundancy, resulting in more severe variance overestimation in the uncalibrated GP-pro model. As a result, the proposed information-based calibration produces larger uncertainty reductions for balltree and $k$-means assignments, even though their final ENCE values may remain higher than those of random assignment.

The entropy-weighted scheme computes the weighting factors of each local GP model based on the entropy difference between the prior and posterior distributions at a test point. 
 \citet{Cao2015} explained that the entropy-weighted scheme is beneficial 
because maximising the entropy differences corresponds to minimising the KL divergence between the distribution of individual local GPs and the 
unknown ground-truth  aggregating distribution.
 \citet{Cao2015} formulated the product of a collection of local GP models as a logarithmic opinion pool of experts \cite{Heskes1997}, where the decision of weighting factors for each local GP can be optimised by minimising the KL divergence of the opinion pool, which can be decomposed into the KL divergences of individual local GPs.
Within the entropy-weighted scheme, the prior entropy reflects the uncertainty before observing the training data, while the posterior entropy reflects the uncertainty after observing the training data. A large entropy difference corresponds to a small KL divergence and entails a higher reliability of a local GP model. 
Thus, by using the entropy-weighted scheme,  
higher weights are given to local GPs that are more reliable (larger entropy difference) in their predictions, and lower weights are given to local GPs that are less reliable (lower entropy difference). These lead to better aggregation of the posterior predictions.

The variance-weighted scheme, which computes the weighting factors based on the variance difference, works in a similar way to the entropy-weighted scheme. 
A consistent pattern emerges when comparing variance- and entropy-weighted schemes. Entropy weighting achieves slightly better predictive accuracy, as measured by RMSE and NLL, likely due to its smoother aggregation of local experts. In contrast, variance-based weighting yields marginally lower ENCE, indicating improved uncertainty calibration. This behaviour aligns with the theoretical distinction between the two schemes: variance weighting more strongly penalises uncertain experts, which benefits calibration, while entropy weighting provides a more conservative aggregation that favours predictive accuracy.

In contrast, the uniform-weighted scheme assigns equal weights to all experts and consistently performs worse. 
This suggests that assuming equal reliability across experts is suboptimal for aggregating posterior predictions.

\begin{figure}
    \centering 

\begin{subfigure}{0.3\textwidth}
  \includegraphics[width=\linewidth]{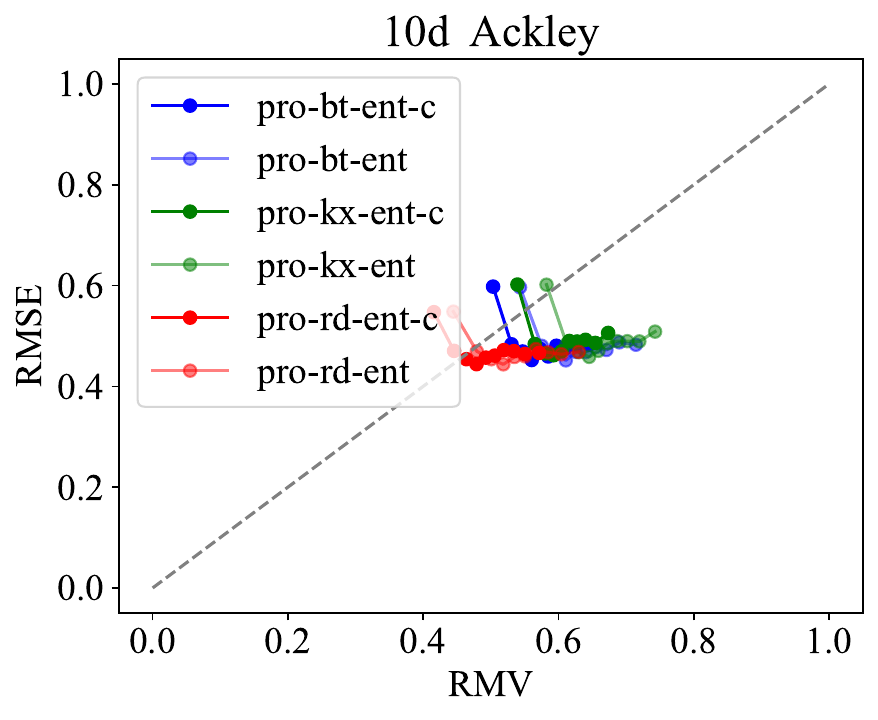}
  \label{fig:1}
\end{subfigure}\hfil 
\begin{subfigure}{0.3\textwidth}
  \includegraphics[width=\linewidth]{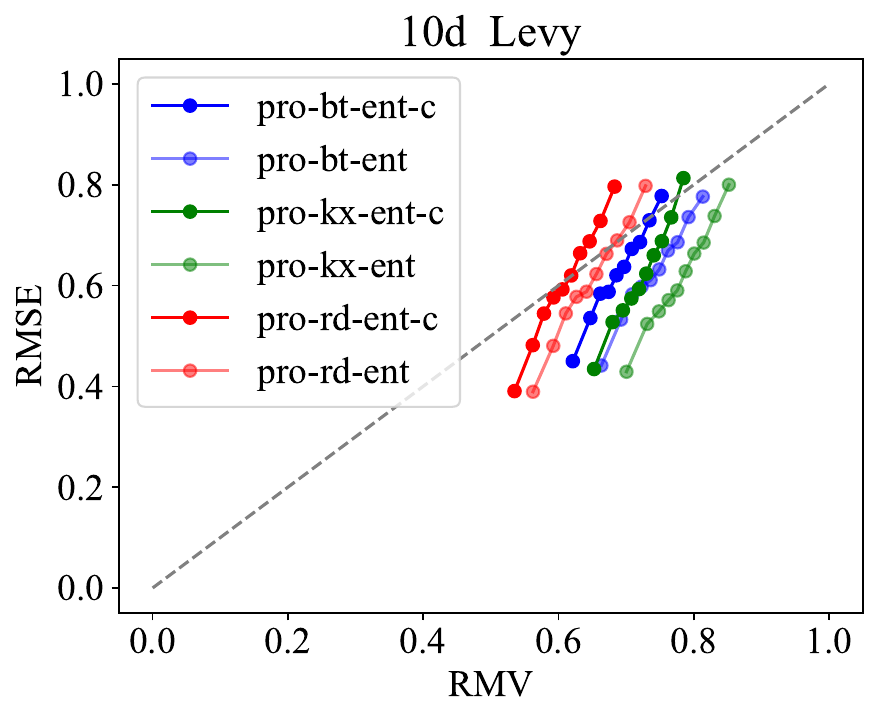}
  \label{fig:2}
\end{subfigure}\hfil 
\begin{subfigure}{0.3\textwidth}
   \includegraphics[width=\linewidth]{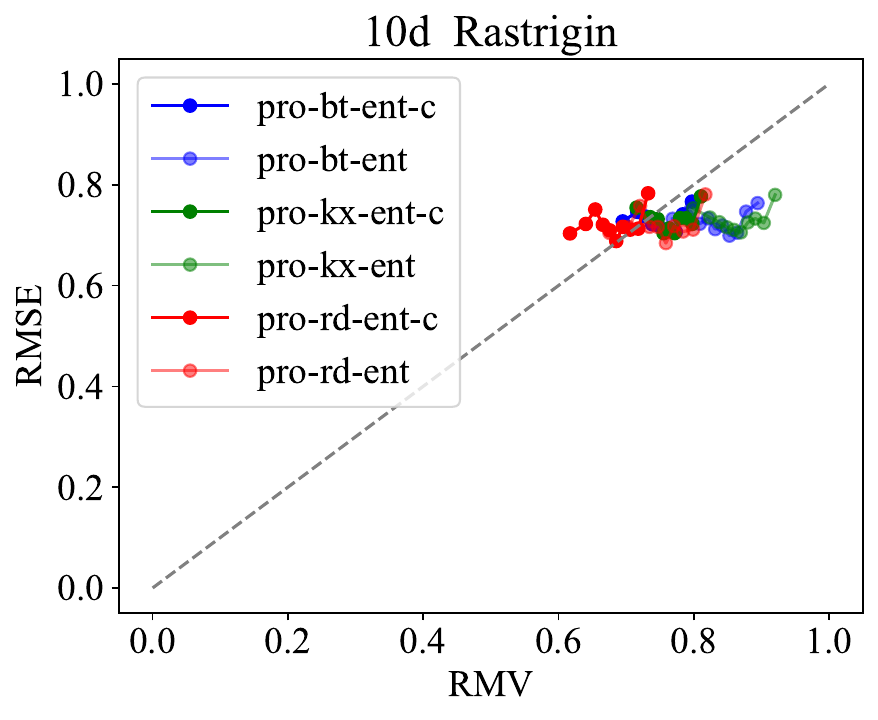}
   \label{fig:2}
\end{subfigure}\hfil 

\begin{subfigure}{0.3\textwidth}
  \includegraphics[width=\linewidth]{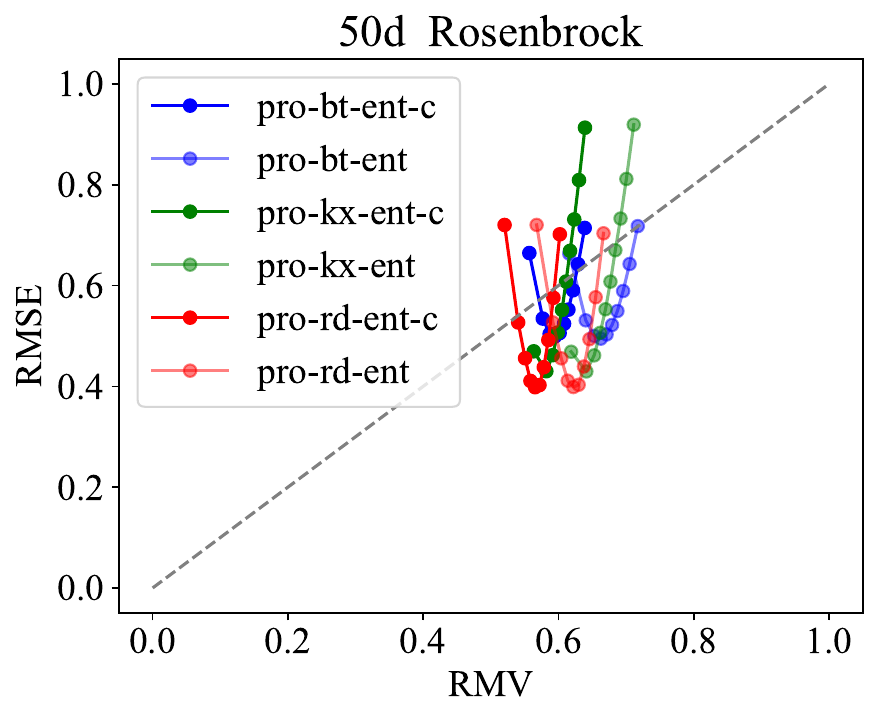}
  \label{fig:5}
\end{subfigure}\hfil 
\begin{subfigure}{0.3\textwidth}
  \includegraphics[width=\linewidth]{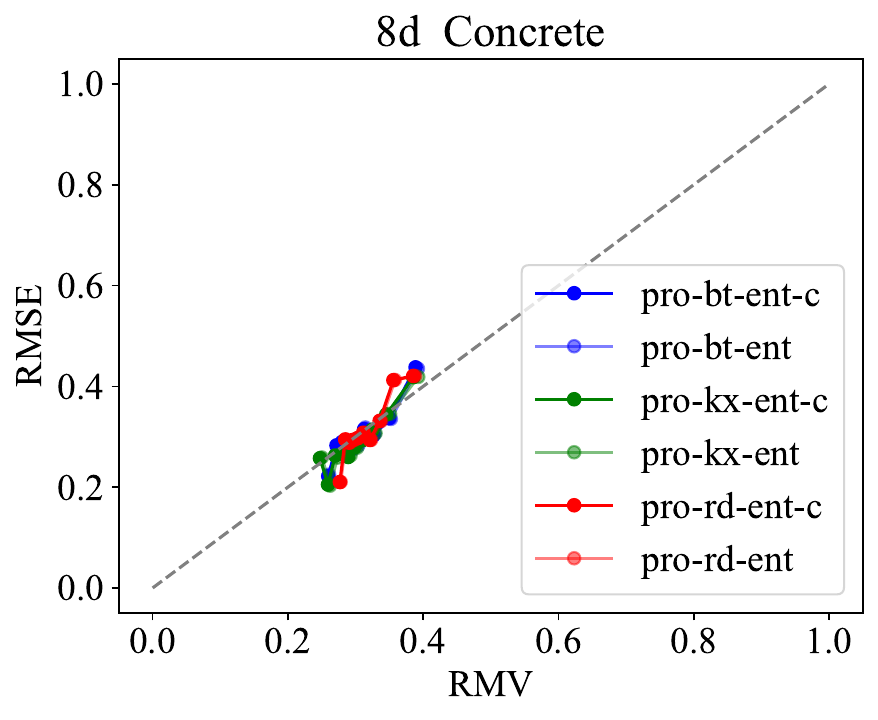}
   \label{fig:1}
\end{subfigure}\hfil 
\begin{subfigure}{0.3\textwidth}
  \includegraphics[width=\linewidth]{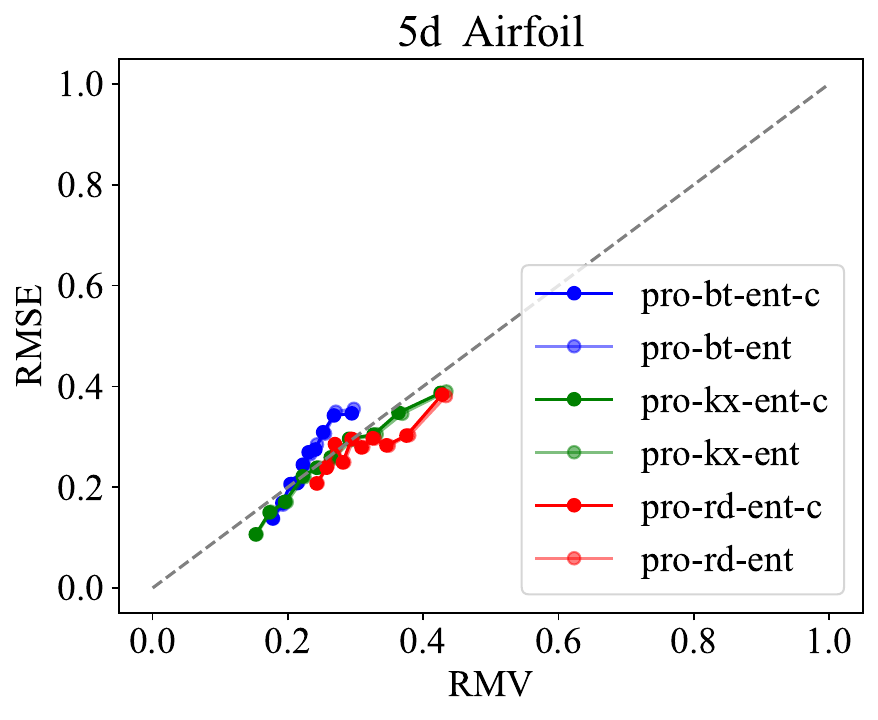}
  \label{fig:2}
\end{subfigure}\hfil 

\begin{subfigure}{0.3\textwidth}
  \includegraphics[width=\linewidth]{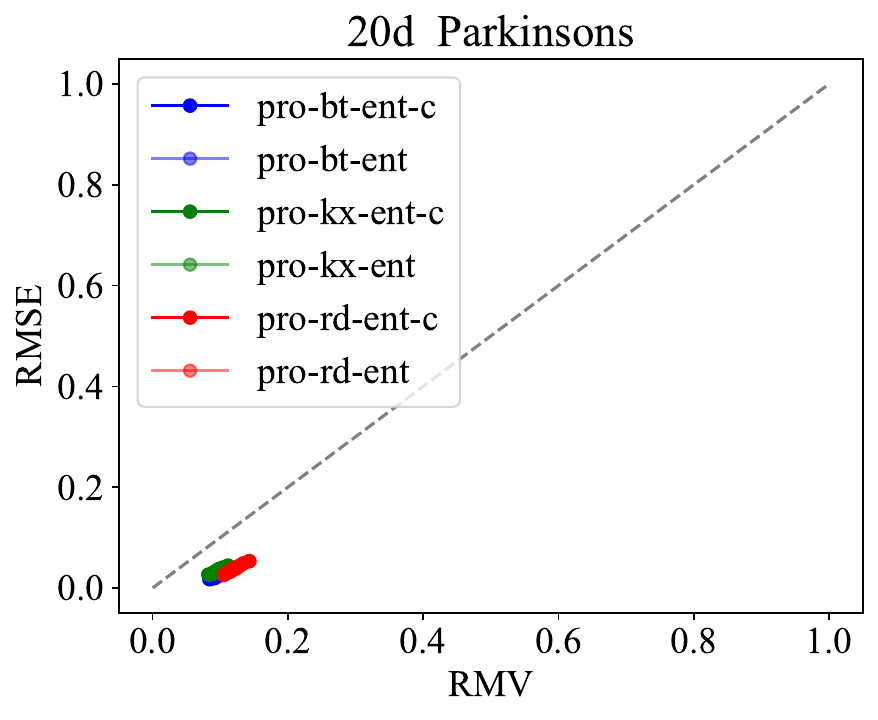}
  \label{fig:4}
\end{subfigure}\hfil 
\begin{subfigure}{0.3\textwidth}
  \includegraphics[width=\linewidth]{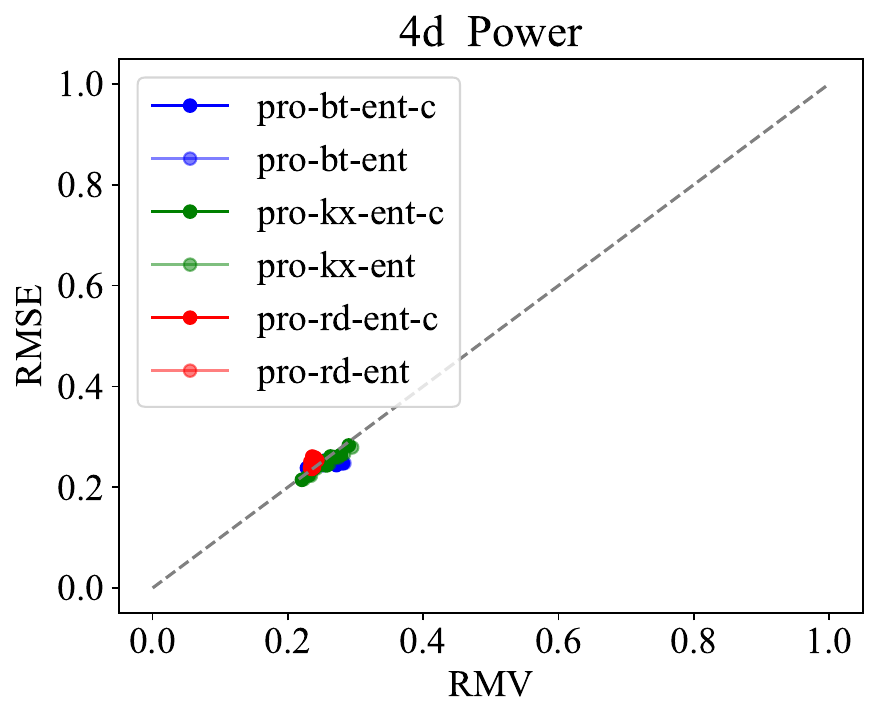}
  \label{fig:5}
\end{subfigure}\hfil 
\begin{subfigure}{0.3\textwidth}
  \includegraphics[width=\linewidth]{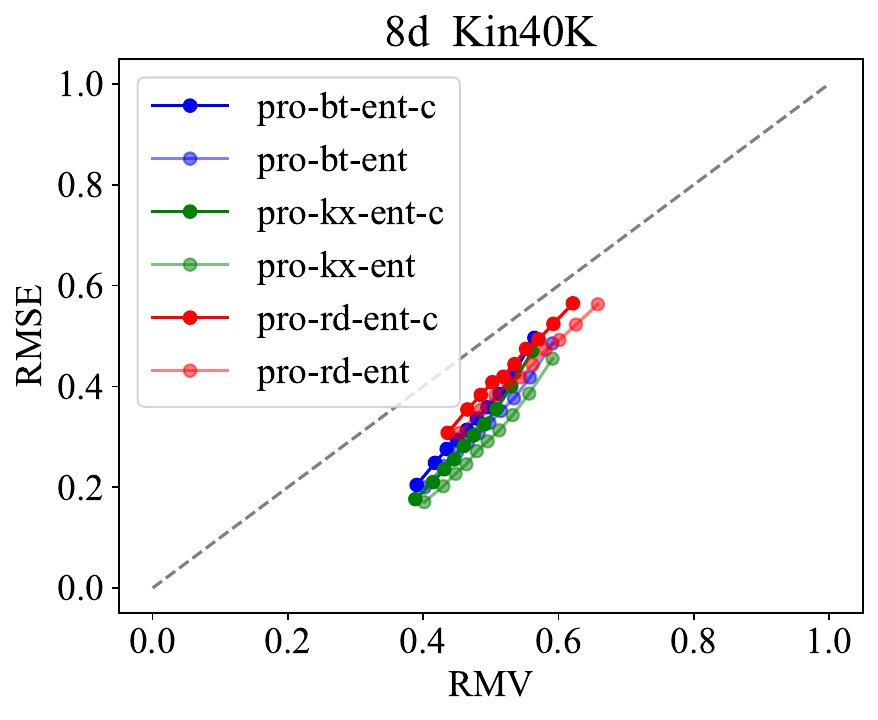}
  \label{fig:4}
\end{subfigure}\hfil 

\begin{subfigure}{0.3\textwidth}
  \includegraphics[width=\linewidth]{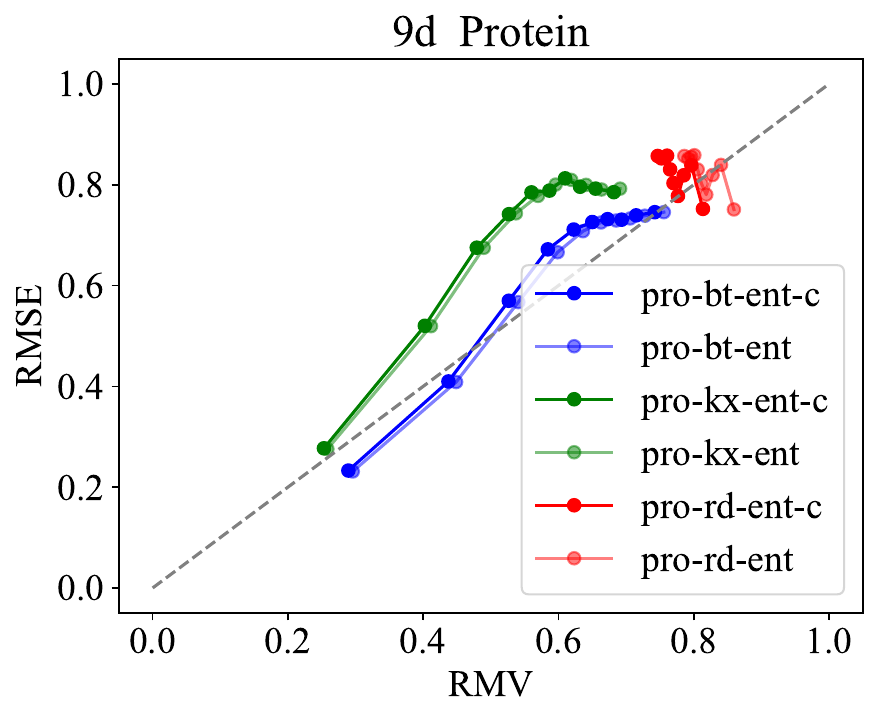}
  \label{fig:5}
\end{subfigure}\hfil 

\caption{Reliability diagrams comparing GP-pro-c and GP-pro models using balltree, $k$-means, and random assignment methods across four synthetic functions and six regression datasets. GP-pro-c consistently yields curves closer to the ideal diagonal than GP-pro, indicating improved uncertainty calibration. The degree of deviation from the diagonal varies across datasets: smoother or lower-dimensional problems (e.g., Concrete, Airfoil, and Power) show near-diagonal curves, whereas noisier or higher-dimensional problems (e.g., Levy, Rosenbrock, Parkinsons, and Kin40K) show larger residual deviations.}

\Description{Ten line graphs show reliability diagrams for six variants of product-of-experts Gaussian process models evaluated on the 10-dimensional Ackley function, 10-dimensional Levy function, 10-dimensional Rastrigin function, 50-dimensional Rosenbrock function, the 8-dimensional Concrete dataset, the 5-dimensional Airfoil dataset, the 20-dimensional Parkinsons dataset, the 4-dimensional Power dataset, the 8-dimensional Kin40K dataset, and the 9-dimensional Protein dataset.

In each graph, the horizontal axis represents the Root Mean Variance (RMV), ranging from 0 to 1 in increments of 0.2, and the vertical axis represents the Root Mean Square Error (RMSE), also ranging from 0 to 1 in increments of 0.2.

Each graph contains seven lines: an ideal dashed diagonal line extending from the lower-left corner to the upper-right corner, and six model-specific lines corresponding to (i) a product-of-experts Gaussian process model using the balltree assignment method, an entropy-weighted aggregation scheme, and uncertainty calibration; (ii) the same model without uncertainty calibration; (iii) a product-of-experts Gaussian-process model using the $k$-means assignment method, an entropy-weighted aggregation scheme, and uncertainty calibration; (iv) the same model without uncertainty calibration; (v) a product-of-experts Gaussian process model using random assignment, an entropy-weighted aggregation scheme, and uncertainty calibration; and (vi) the same model without uncertainty calibration.

Across all graphs, the calibrated product-of-experts Gaussian process model using the balltree assignment method generally produces a curve that lies closest to the ideal diagonal, indicating better uncertainty calibration than the other model variants. Curves corresponding to uncalibrated models tend to deviate more substantially from the diagonal.}

\label{fig:fig_reliability_btkxrd}
\end{figure}

Figure \ref{fig:fig_reliability_btkxrd} presents the reliability diagrams for GP-pro-c and GP-pro models under different data assignment methods, including balltree,  $k$-means, and random assignment. 
Across all cases, GP-pro-c consistently exhibits reliability curves closer to the ideal diagonal line than GP-pro, demonstrating improved calibration. 
Deviations from the diagonal vary across datasets: smoother or lower-dimensional datasets (e.g., Concrete, Airfoil, Power) show near-diagonal curves, whereas noisier or higher-dimensional datasets (e.g., Levy, Rosenbrock, Parkinsons, Kin40K) exhibit larger residual deviations.

Figure \ref{fig:fig_reliability_bt} in Appendix \ref{appendix_exp_results} further reports reliability diagrams for GP-pro-c and GP-pro using the balltree assignment while comparing entropy-, variance-, and uniform-weighted aggregation schemes. The results show that, irrespective of the weighting scheme, GP-pro-c achieves curves closer to the diagonal. Among the GP-pro-c variants, the combination of balltree assignment and entropy weighting consistently produces reliability curves closest to the diagonal, indicating more accurate uncertainty quantification.

Table  \ref{tab:tab_train_time} lists the training and test time required by the GP-pro-c and GP-pro models.
GP-pro-c exhibits computational costs comparable to GP-pro, indicating that the proposed calibration introduces only negligible overhead.

Overall, the results demonstrate that the information-based calibration method improves uncertainty quantification in GP-pro-c without compromising predictive accuracy or computational efficiency.

\begin{table}
\caption{Average training and test times (in seconds) across four synthetic functions and six regression datasets using balltree, $k$-means, and random assignment methods, as well as entropy- and variance-weighted schemes (the uniform-weighted scheme is omitted for brevity) for the GP-pro-c and GP-pro models. Variants of GP-pro-c have training and test times similar to those of GP-pro.}
 \label{tab:tab_train_time}
\begin{scriptsize}
\begin{subtable}{1\textwidth}
\sisetup{table-format=-1.2}   
\centering
   \begin{tabular}{@{}lrrSSSSSS@{}}
      \toprule
      \textbf{Dataset} & \textbf{Size} & \textbf{Dim} & \textbf{bt-ent-c} & \textbf{bt-var-c}  & \textbf{kx-ent-c} & \textbf{kx-var-c}  & \textbf{rd-ent-c} & \textbf{rd-var-c}  \\ 
      & & & {training, test time} &       &  &      &  &        \\
      \midrule
      Ackley & 5000 & 10            & {3.101, 1.427}  & {3.101, 1.427}    & {3.265, 1.396}  & {3.265, 1.395}    & {3.166, 1.427}  & {3.166, 1.427}    \\ 
      Levy & 5000 & 10              & {3.380, 1.452}  & {3.380, 1.452}    & {3.479, 1.352}  & {3.479, 1.352}    & {3.413, 1.459}  & {3.413, 1.459}    \\ 
      Rastrigin & 5000 & 10        & {3.119, 1.454}  & {3.119, 1.454}    & {3.371, 1.435}  & {3.371, 1.435}    & {3.069, 1.449}  & {3.069, 1.449}     \\ 
      Rosenbrock  & 50000 & 50  & {32.522, 85.483}  & {32.522, 85.467}    & {33.917, 84.008}  & {33.917, 83.992}    & {30.327, 81.578}  & {30.327, 81.565}   \\
      Concrete & 1030 &  8        & {1.263, 0.037}  & {1.263, 0.037}    & {1.238, 0.031}  & {1.238, 0.031}    & {1.301, 0.033}  & {1.301, 0.033}    \\ 
      Airfoil & 1503 &  5              & {1.359, 0.066}  & {1.359, 0.066}    & {1.367, 0.061}  & {1.367, 0.061}    & {1.282, 0.063}  & {1.282, 0.063}    \\ 
      Parkinsons & 5875 &  20  & {4.282, 2.581}  & {4.282, 2.581}   & {4.514, 2.724}  & {4.514, 2.723}    & {3.745, 2.463}  & {3.745, 2.463}    \\ 
      Power & 9568 &  4            & {5.779, 0.538}  & {5.779, 0.537}    & {6.151, 0.578}  & {6.151, 0.577}   & {5.495, 0.502}  & {5.495, 0.502}   \\ 
      Kin40K & 40000 &  8         & {22.336, 41.573}  & {22.336, 41.558}   & {24.403, 42.255}  & {24.403, 42.242}    & {20.518, 41.063}  & {20.518, 41.055}    \\ 
      Protein & 45730 &  9         & {27.192, 58.806}  & {27.192, 58.793}    & {32.110, 63.094}  & {32.110, 63.068}    & {24.176, 58.166}  & {24.176, 58.150}    \\ 
      \bottomrule
      Average time $\downarrow$  &   &               & {10.433, 19.342}  & {10.433, 19.337}    & {11.381, 19.693}  & {11.381, 19.688}    & {9.649, 18.820}  & {9.649, 18.816}   \\
       \bottomrule
   \end{tabular}
   \caption{Training and test time for GP-pro-c models}\label{tab:sub_first}
\end{subtable}

\bigskip
\begin{subtable}{1\textwidth}
\sisetup{table-format=-1.2}   
\centering
   \begin{tabular}{@{}lrrcccccc@{}}
      \toprule
      \textbf{Dataset} & \textbf{Size} & \textbf{Dim} & \textbf{bt-ent} & \textbf{bt-var}   & \textbf{kx-ent} & \textbf{kx-var}   & \textbf{rd-ent} & \textbf{rd-var}  \\ 
      & & & {training, test time} &      &   &      &   &       \\
      \midrule
      Ackley & 5000 & 10           & {3.101, 1.425}  & {3.101, 1.425}    & {3.265, 1.394}  & {3.265, 1.394}    & {3.166, 1.425}  & {3.166, 1.425}    \\ 
      Levy & 5000 & 10              & {3.380, 1.450}  & {3.380, 1.450}    & {3.479, 1.350}  & {3.479, 1.350}    & {3.413, 1.457}  & {3.413, 1.457}     \\ 
      Rastrigin & 5000 & 10       & {3.119, 1.452}  & {3.119, 1.452}    & {3.371, 1.433}  & {3.371, 1.433}    & {3.069, 1.447}  & {3.069, 1.447}    \\ 
      Rosenbrock & 50000 & 50  & {32.522, 85.375}  & {32.522, 85.356}    & {33.917, 83.896}  & {33.917, 83.879}    & {30.327, 81.470}  & {30.327, 81.457}    \\
      Concrete & 1030 &  8          & {1.263, 0.036}  & {1.263, 0.036}    & {1.238, 0.031}  & {1.238, 0.031}    & {1.301, 0.033}  & {1.301, 0.033}   \\ 
      Airfoil & 1503 &  5               & {1.359, 0.065}  & {1.359, 0.065}    & {1.367, 0.060}  & {1.367, 0.060}    & {1.282, 0.062}  & {1.282, 0.062}    \\ 
      Parkinsons & 5875 &  20  & {4.282, 2.579}  & {4.282, 2.579}    & {4.514, 2.721}  & {4.514, 2.721}    & {3.745, 2.461}  & {3.745, 2.460}    \\ 
      Power & 9568 &  4            & {5.779, 0.533}  & {5.779, 0.532}    & {6.151, 0.573}  & {6.151, 0.573}    & {5.495, 0.498}  & {5.495, 0.497}    \\ 
      Kin40K & 40000 &  8        & {22.336, 41.503}  & {22.336, 41.489}    & {24.403, 42.187}  & {24.403, 42.172}    & {20.518, 40.994}  & {20.518, 40.987}   \\ 
      Protein & 45730 &  9       & {27.192, 58.716}  & {27.192, 58.701}    & {32.110, 62.996}  & {32.110, 62.982}    & {24.176, 58.073}  & {24.176, 58.055}   \\ 
      \bottomrule
      Average time $\downarrow$  &   &           & {10.433, 19.313}  & {10.433, 19.309}    & {11.381, 19.664}  & {11.381, 19.659}    & {9.649, 18.792}  & {9.649, 18.788}   \\
       \bottomrule
   \end{tabular}
   \caption{Training and test time for GP-pro models}\label{tab:sub_first}
\end{subtable}

 \end{scriptsize}
\end{table}

\subsection{Comparing the Number of Training Data Points per Local Gaussian Process Model }

The second set of experiments examines the impact of varying the number of training data points per local GP on predictive performance and computational cost of the GP-pro-c models. 
Specifically, we consider models with 100, 200, 300, 400, 500, and 600 data points per local expert.

As discussed in Section \ref{sec:sec_intro}, product-of-experts GP models reduce the computational overhead of a single global GP from cubic complexity $ \mathcal{O}(n^3)$ to $ \mathcal{O}(M \tilde{n}^3)$, where $n$ is the number of training data points, $\tilde{n}$ is the number of training data points under each local GP, and $ \tilde{n} \ll n$.
Within a local GP model, the more data points allocated, the more accurate the model is, while also incurring higher computational overhead.

\begin{figure}
    \centering 

\medskip
\begin{subfigure}{0.4\textwidth}
  \includegraphics[width=\linewidth]{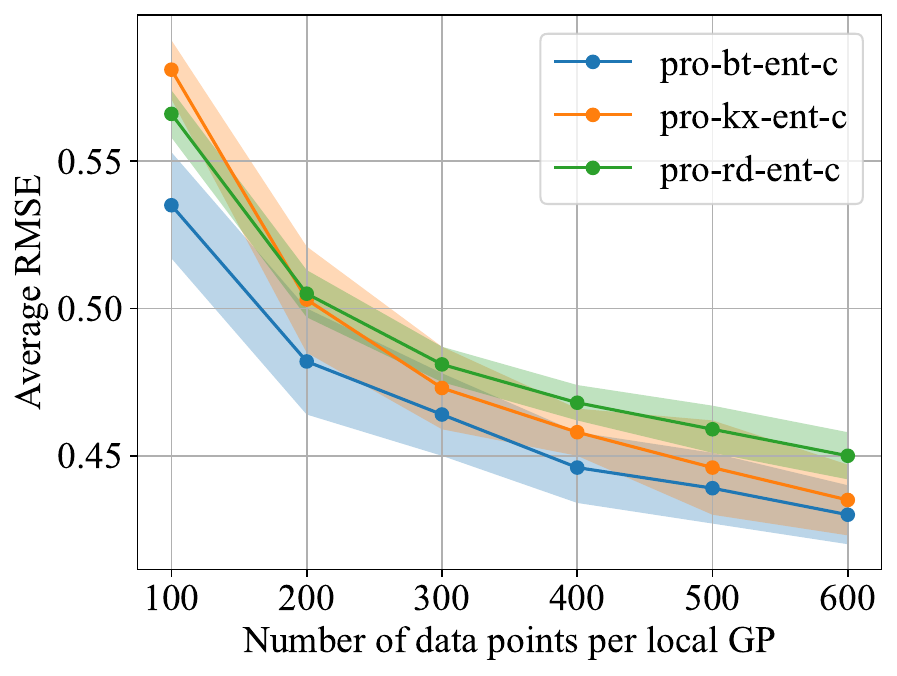}
  \caption{RMSE}
  \label{fig:4}
\end{subfigure}\hfil 
\begin{subfigure}{0.4\textwidth}
  \includegraphics[width=\linewidth]{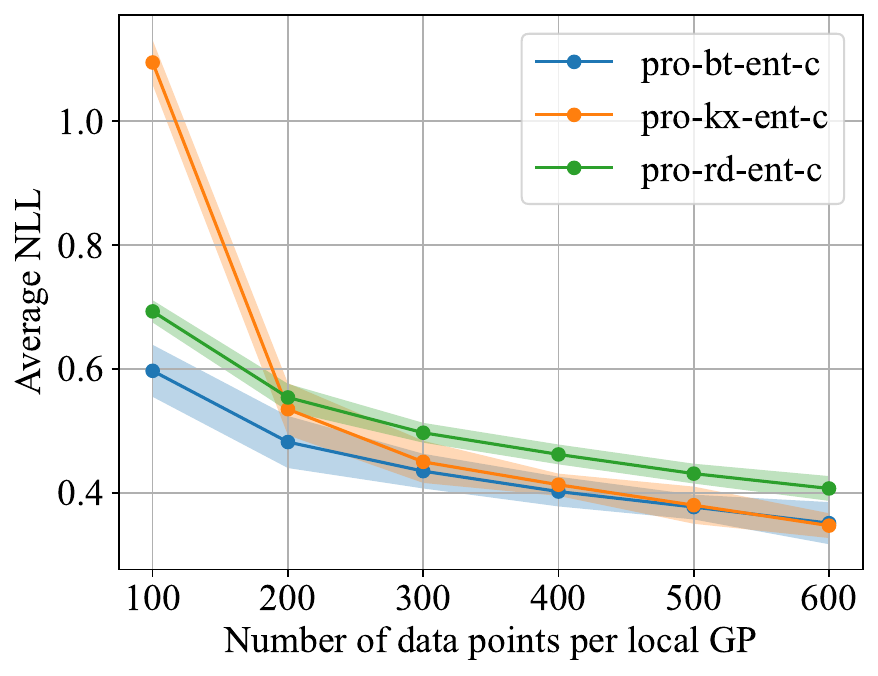}
  \caption{NLL}
  \label{fig:5}
\end{subfigure}\hfil 

\bigskip
\begin{subfigure}{0.4\textwidth}
  \includegraphics[width=\linewidth]{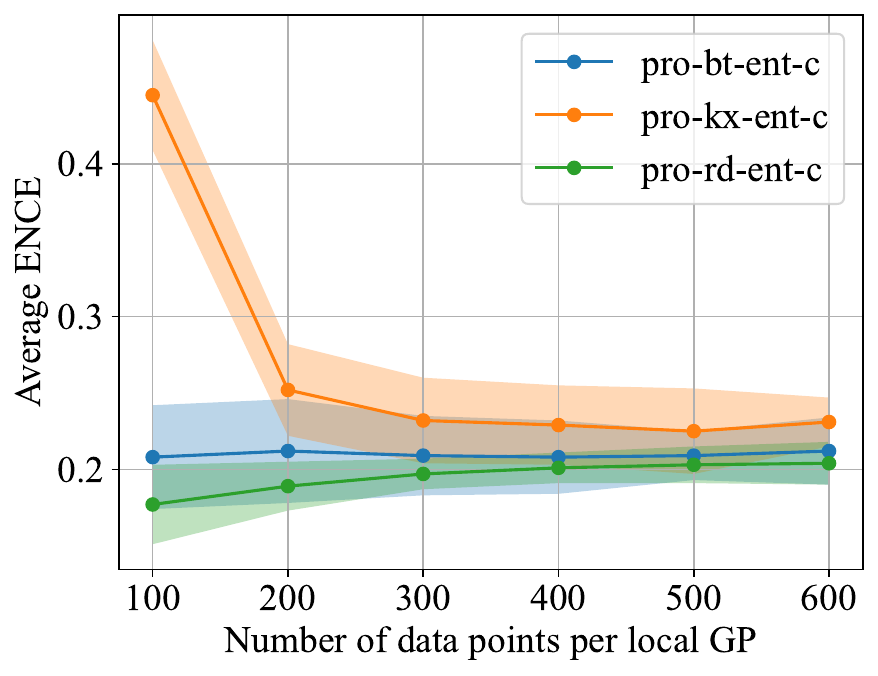}
  \caption{ENCE}
  \label{fig:4}
\end{subfigure}\hfil 
\begin{subfigure}{0.4\textwidth}
  \includegraphics[width=\linewidth]{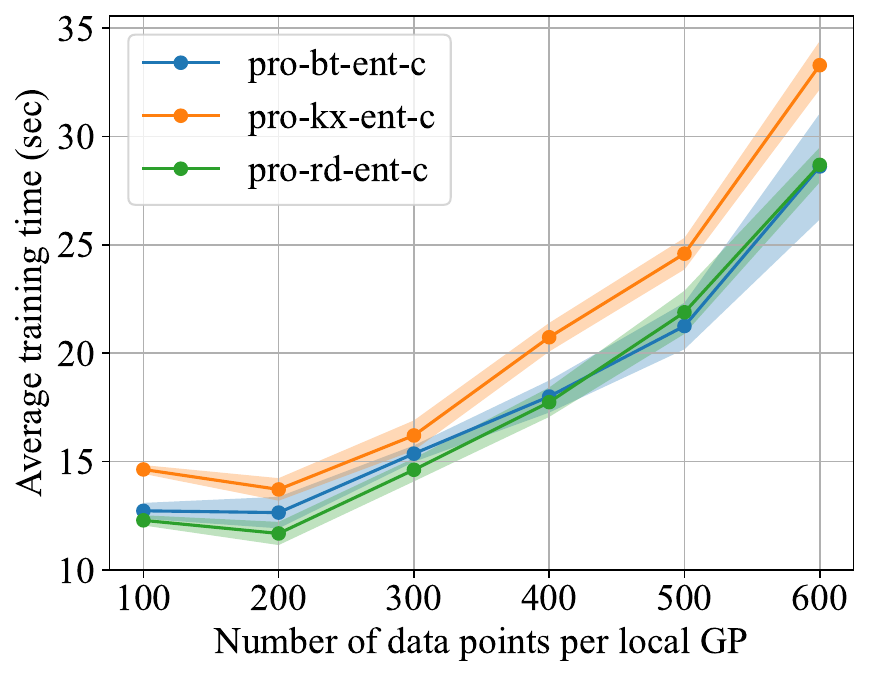}
  \caption{Training time}
  \label{fig:5}
\end{subfigure}\hfil 

\caption{ Average NLL, RMSE, ENCE, and training time for GP-pro-c models using balltree, $k$-means, and random assignment methods with 100-600 data points per local GP, evaluated on four synthetic functions and four benchmark datasets. Random assignment yields lower ENCE but higher RMSE and NLL due to conservative, weakly correlated local GPs. Balltree and $k$-means preserve locality, yielding lower RMSE and NLL but higher ENCE. $k$-means performs poorly with small cluster sizes because small, high-dimensional clusters can be heterogeneous and poorly conditioned; performance improves sharply as cluster size increases. $k$-means can also incur slightly higher computational overhead due to uneven cluster sizes, whereas balltree and random assignments are more balanced. Overall, balltree assignment with 200-400 points per local GP provides a good trade-off between accuracy, calibration, and computational cost.
}
\Description{Four line graphs labelled (a), (b), (c), and (d).

In all graphs, the horizontal x-axis shows the number of data points per local Gaussian process, ranging from 100 to 600 in increments of 100. Each graph contains three lines representing product-of-experts Gaussian process models constructed using the balltree, $k$-means, and random assignment methods, respectively. All models use the entropy-weighted scheme with uncertainty calibration.

In (a), the vertical y-axis shows the average Root Mean Square Error (RMSE), ranging from 0.40 to 0.60 in increments of 0.05. All three lines decrease consistently as the number of data points increases. The balltree line remains the lowest throughout. The $k$-means line is initially the highest at 100 data points, but from 200 data points onward it becomes the second lowest, overtaking the random assignment line.

In (b), the vertical y-axis shows the average Negative Log-Likelihood (NLL), ranging from 0.2 to 1.2 in increments of 0.2. All three lines decrease consistently. The balltree line remains the lowest throughout. The $k$-means line is initially the highest at 100 data points, but from 200 data points onward it becomes the second lowest, overtaking the random assignment line.

In (c), the vertical y-axis shows the average Expected Normalised Calibration Error (ENCE), ranging from 0.1 to 0.5 in increments of 0.1. The balltree and random assignment lines remain the lowest, at approximately 0.2 ENCE, throughout the graph. The $k$-means line begins at 0.45 ENCE with 100 data points, then decreases sharply to approximately 0.25 ENCE at 200 data points and remains the highest thereafter.

In (d), the vertical y-axis shows the average training time in seconds, ranging from 10 to 35 in increments of 5. All three lines increase exponentially as the number of data points grows. The balltree and random assignment lines remain the lowest throughout, while the $k$-means line remains the highest.}

\label{fig:fig_rmse_nll_ence_time}
\end{figure}

Figure \ref{fig:fig_rmse_nll_ence_time} shows the average RMSE, NLL, ENCE scores, and training time for GP-pro-c models using balltree, $k$-means, and random assignment methods with 100, 200, 300, 400, 500, and 600 data points per local GP, evaluated on four synthetic functions and four benchmark datasets ($10d$ Ackley, $10d$ Levy, $10d$ Rastrigin, $50d$ Rosenbrock, $4d$ Power, $20d$ Parkinson, $8d$ Kinetic, and $9d$ Protein). 
RMSE and NLL decrease monotonically as the number of points per local GP increases, while training time increases accordingly. 
Models with 100 points per local GP exhibit the highest RMSE and NLL but the lowest training time, whereas models with 600 points achieve the best predictive accuracy at the cost of significantly higher computational overhead. 
Models with 200–400 points per local GP strike a favourable balance between predictive performance and computational cost.

Across different local GP sizes, it is observed that random assignment achieves lower absolute ENCE because weakly correlated local experts produce conservative predictive variances that tend to match empirical errors, despite degraded posterior mean accuracy. This results in higher RMSE and NLL but lower ENCE. 
In contrast, balltree and $k$-means preserve locality and yield more accurate posterior means, reducing RMSE and NLL, while inducing higher information redundancy across experts that makes uncertainty calibration more challenging  
and leads to higher ENCE.

Notably, $k$-means performs poorly when the number of points per local GP is small, because small clusters in high-dimensional spaces can be heterogeneous and poorly conditioned. This leads to unstable local GP estimates and inflated NLL and ENCE, which improves sharply once each expert has sufficient data.

In terms of computational cost, $k$-means can be slightly more expensive than random or balltree assignment due to imbalanced cluster sizes. 
In contrast, balltree and random assignment tend to produce more balanced partitions, leading to more stable computational requirements.

Overall, balltree assignment with 200-400 data points per local GP provides the best trade-off between predictive accuracy, uncertainty calibration, and computational efficiency.

\subsection{Comparison with Baseline Methods }

The third set of experiments compares the performance and computational cost of GP-pro-c against three baselines: 
(i) GP-pro without calibration, 
(ii) GP-pro with softmax weight calibration \cite{Cohen2020}, denoted GP-pro-sm, and 
(iii) a single global GP, denoted GP-full. 
All GP-pro and GP-pro-c variants use balltree assignment, entropy-based weighting, and 200 data points per local expert.

\begin{figure}
    \centering 

\begin{subfigure}{0.3\textwidth}
  \includegraphics[width=\linewidth]{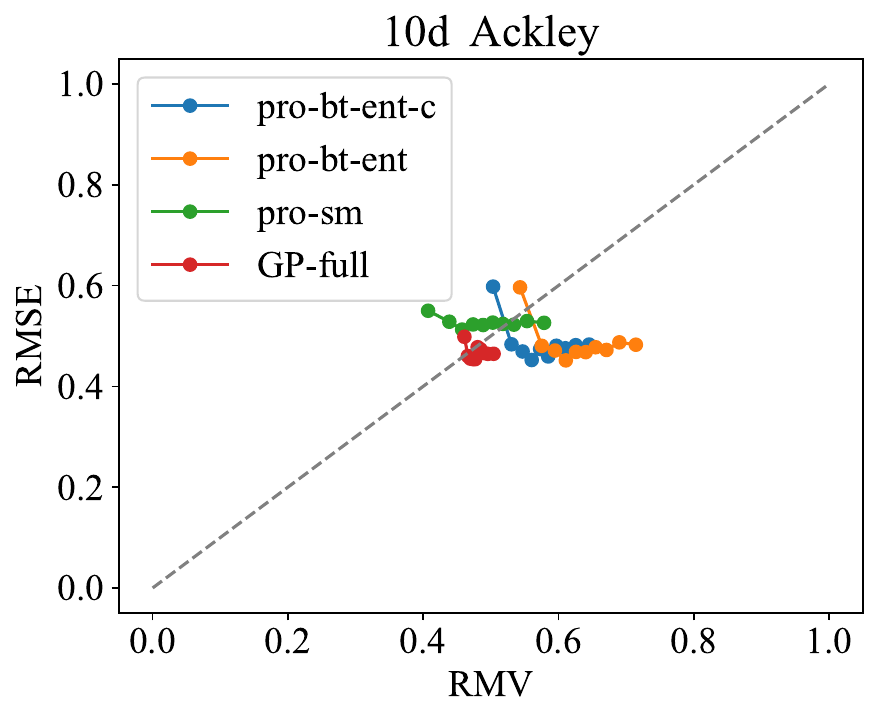}
\end{subfigure}\hfil 
\begin{subfigure}{0.3\textwidth}
  \includegraphics[width=\linewidth]{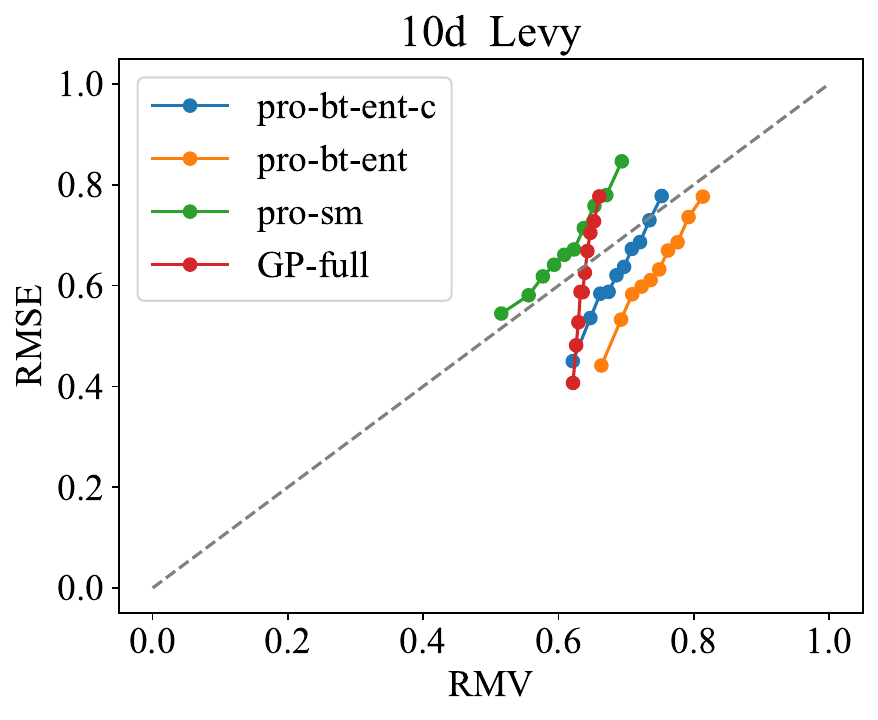}
\end{subfigure}\hfil 
\begin{subfigure}{0.3\textwidth}
   \includegraphics[width=\linewidth]{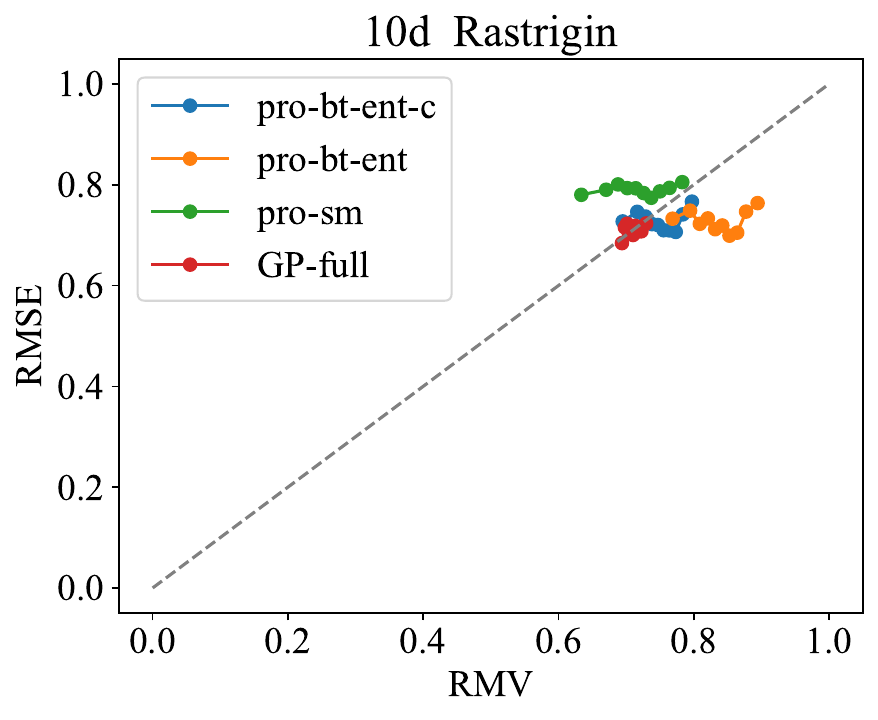}
\end{subfigure}\hfil

\begin{subfigure}{0.3\textwidth}
  \includegraphics[width=\linewidth]{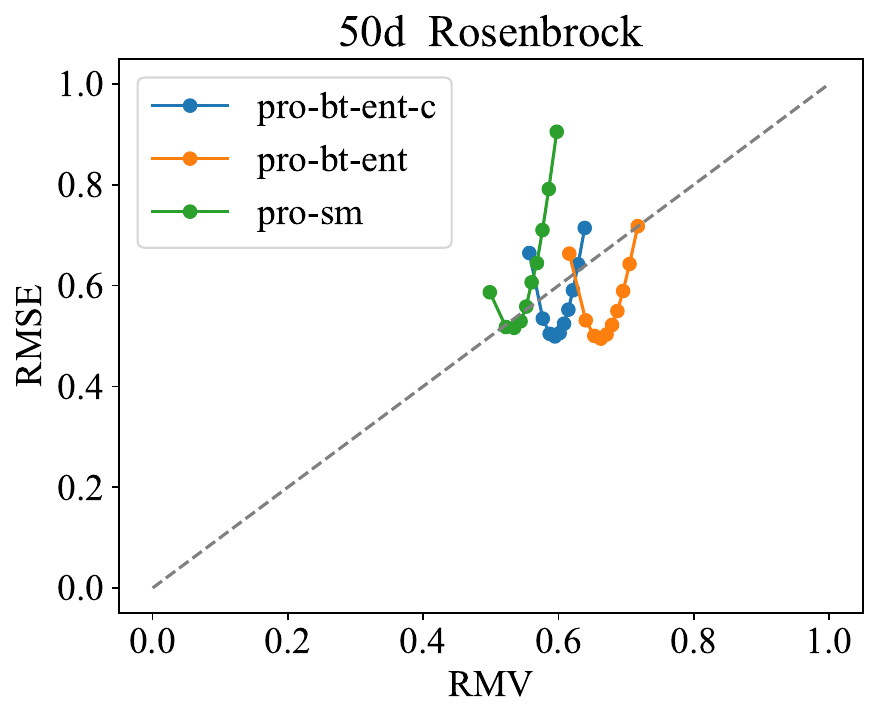}
\end{subfigure}\hfil 
\begin{subfigure}{0.3\textwidth}
  \includegraphics[width=\linewidth]{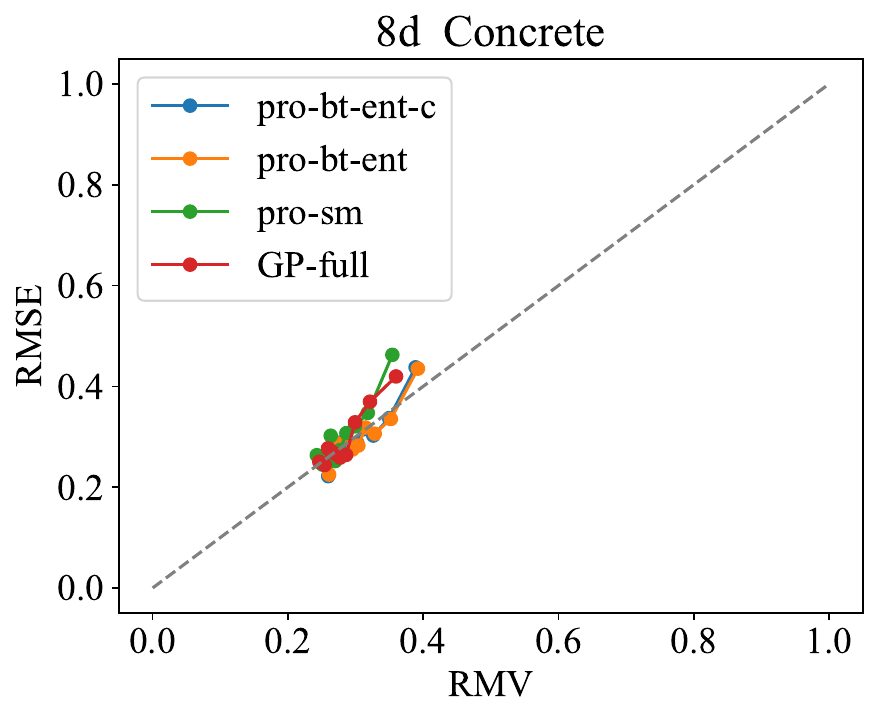}
\end{subfigure}\hfil 
\begin{subfigure}{0.3\textwidth}
  \includegraphics[width=\linewidth]{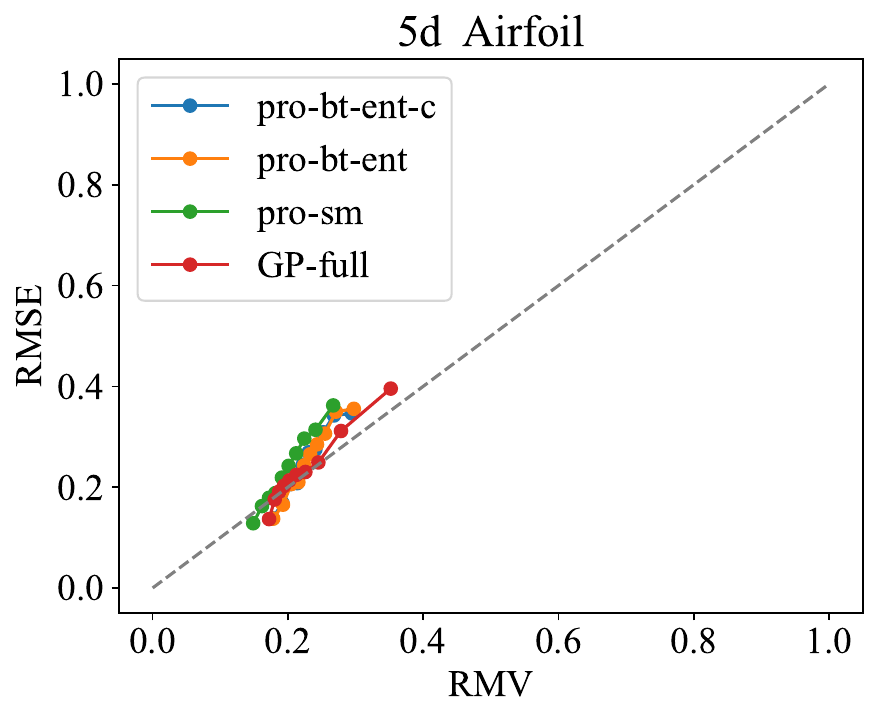}
\end{subfigure}\hfil 

\begin{subfigure}{0.3\textwidth}
  \includegraphics[width=\linewidth]{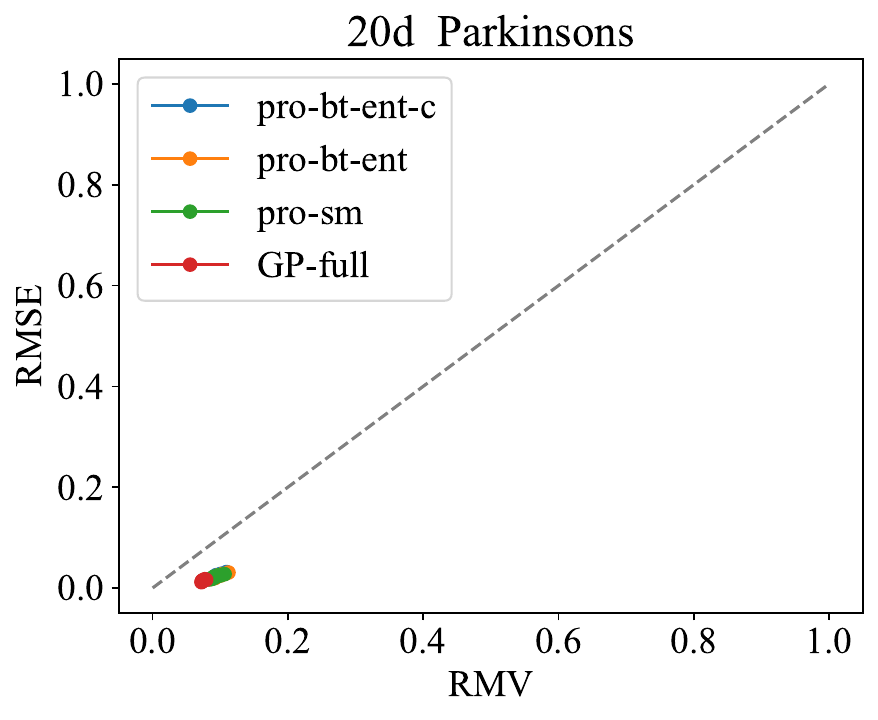}
\end{subfigure}\hfil 
\begin{subfigure}{0.3\textwidth}
  \includegraphics[width=\linewidth]{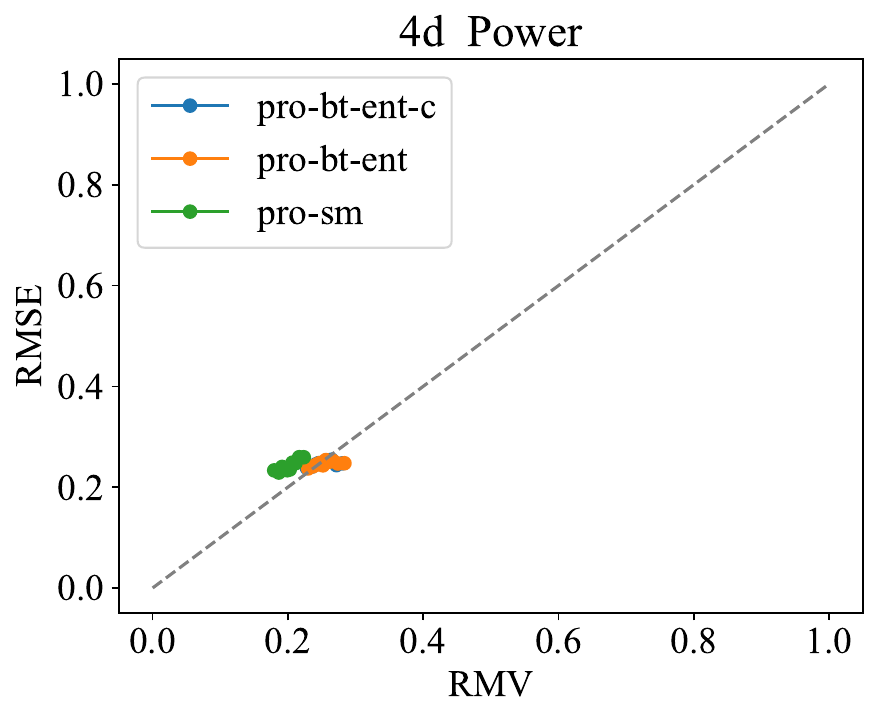}
\end{subfigure}\hfil 
\begin{subfigure}{0.3\textwidth}
  \includegraphics[width=\linewidth]{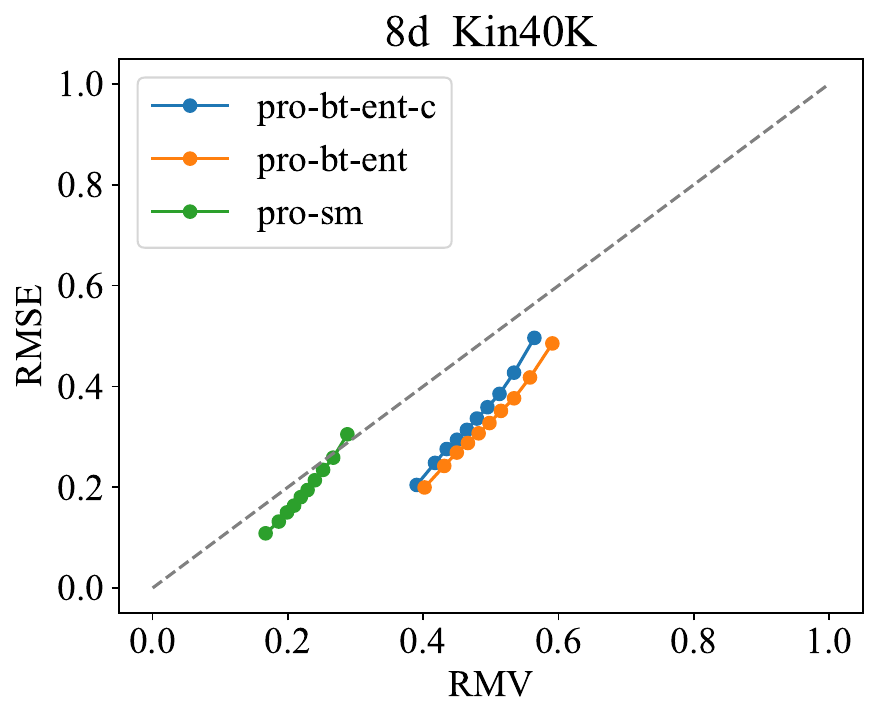}
\end{subfigure}\hfil 

\begin{subfigure}{0.3\textwidth}
  \includegraphics[width=\linewidth]{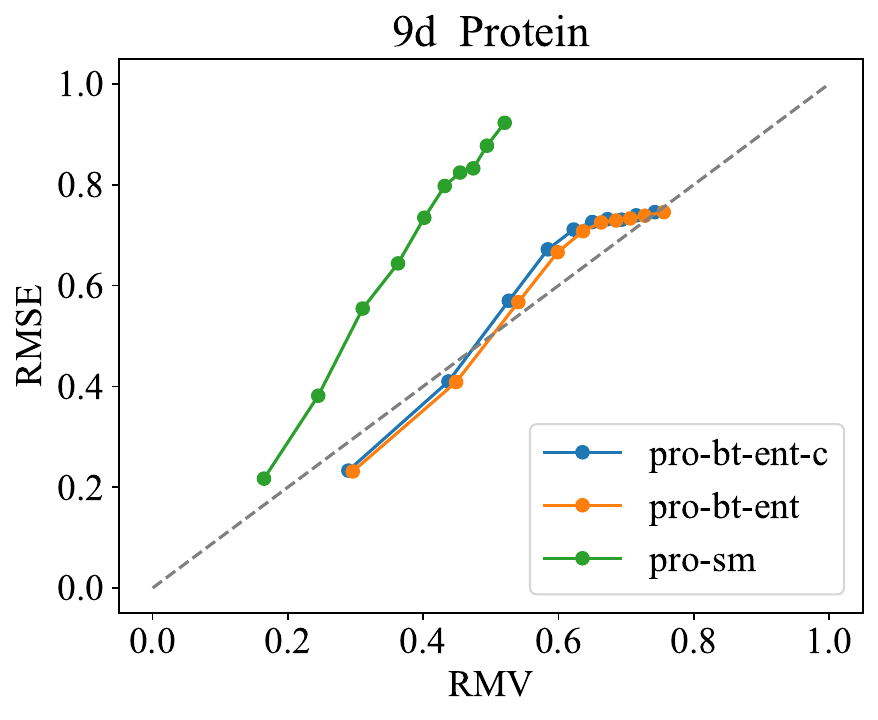}
\end{subfigure}\hfil 

\caption{Reliability diagrams for the GP-pro-c, GP-pro, GP-pro-sm, and GP-full models on four synthetic benchmark functions and six regression datasets.
The GP-full models consistently exhibit curves closest to the diagonal.
The GP-pro-c models demonstrate improved calibration compared to GP-pro across most datasets. On the Kin40K dataset, the GP-pro-sm model exhibits a curve closest to the diagonal. However, on the Protein dataset, GP-pro-sm exhibits noticeable deviation from the diagonal, whereas GP-pro-c maintains more stable calibration behaviour across datasets.
}
\Description{Ten line graphs showing reliability diagrams for six types of Gaussian process models on the 10-dimensional Ackley function, 10-dimensional Levy function, 10-dimensional Rastrigin function, 50-dimensional Rosenbrock function, 8-dimensional Concrete dataset, 5-dimensional Airfoil dataset, 20-dimensional Parkinsons dataset, 4-dimensional Power dataset, 8-dimensional Kin40K dataset, and 9-dimensional Protein dataset.

On each line graph, the horizontal x-axis shows Root Mean Variance ranging from 0 to 1 in increments of 0.2, and the vertical y-axis shows Root Mean Square Error ranging from 0 to 1 in increments of 0.2.

Each plot contains five lines: a dashed diagonal line from the lower-left to the upper-right corner representing perfect calibration, and four curves corresponding to (i) a product-of-experts Gaussian process model using the balltree assignment method with entropy-weighted scheme and uncertainty calibration, (ii) the same model without uncertainty calibration, (iii) the same model with softmax weight calibration, and (iv) a single global Gaussian process model.

In the Ackley, Levy, Rastrigin, Concrete, Airfoil, and Parkinsons datasets, the single global Gaussian process model produces curves closest to the ideal diagonal line.

Across all datasets except Kin40K, the balltree-based product-of-experts Gaussian process model with entropy weighting and uncertainty calibration yields curves closer to the ideal diagonal line than the corresponding model with softmax weight calibration.}

\label{fig:fig_reliability_baseline}
\end{figure}

Table \ref{tab:tab_rmse_nll_ence_baseline} reports performance metrics, Figure \ref{fig:fig_reliability_baseline} shows reliability diagrams, and Table \ref{tab:tab_train_test_time_baseline} summarises training and test times across four synthetic functions and six benchmark datasets.

Overall, GP-pro-c achieves lower RMSE, NLL, and ENCE than both GP-pro and GP-pro-sm. 
In addition, its reliability curves are consistently closer to the ideal diagonal, indicating improved uncertainty calibration. 
These results demonstrate that the information-based calibration method is doing a good job in maintaining accuracy performance while yielding better uncertainty calibration compared to the softmax weight calibration used in the GP-pro-sm models.

\begin{table}
\caption{Average RMSE, NLL, and ENCE values across four synthetic functions and six regression benchmark datasets, comparing GP-pro-c models with uncertainty calibration, GP-pro models without calibration, GP-pro-sm models using softmax-weight calibration, and a single global GP (GP-full). GP-pro-c models achieve lower NLL and ENCE scores, indicating that the proposed information-based calibration maintains predictive accuracy while improving uncertainty calibration.}
 \label{tab:tab_rmse_nll_ence_baseline}
\centering
\begin{scriptsize}
\begin{tabular}{@{}lrrcccccccccccc@{}}
    \toprule
    \textbf{Dataset} &
    \textbf{Size} &
    \textbf{Dim} &
      \multicolumn{3}{c}{\textbf{GP-pro-bt-ent-c}} &
      \multicolumn{3}{c}{\textbf{GP-pro-bt-ent}} &
      \multicolumn{3}{c}{\textbf{GP-pro-sm}} &
      \multicolumn{3}{c}{\textbf{GP-full}}  \\
      \cmidrule(r){4-6}\cmidrule(r){7-9}\cmidrule(l){10-12}\cmidrule(l){13-15}
      & & & {RMSE$\downarrow$} & {NLL$\downarrow$} & {ENCE$\downarrow$} & {RMSE$\downarrow$} & {NLL$\downarrow$} & {ENCE$\downarrow$} & {RMSE$\downarrow$} & {NLL$\downarrow$} & {ENCE$\downarrow$} & {RMSE$\downarrow$} & {NLL$\downarrow$} & {ENCE$\downarrow$}  \\
      \midrule
    Ackley           & 5000  &  10   & 0.487 & 0.742 & 0.190    & 0.487 & 0.775 & 0.245    & 0.527 & 0.801 & 0.115    & 0.468 & 0.663 & 0.057   \\ 
    Levy              & 5000  &  10   & 0.681 & 1.012 & 0.103    & 0.681 & 1.021 & 0.160    & 0.731 & 1.097 & 0.115    & 0.653 & 0.982 & 0.137   \\ 
    Rastrigin        & 5000  &  10   & 0.745 & 1.124 & 0.060    & 0.745 & 1.138 & 0.127    & 0.801 & 1.211 & 0.116    & 0.713 & 1.081 & 0.039  \\ 
    Rosenbrock   & 50000  &  50   & 0.610 & 0.918 & 0.116    & 0.610 & 0.929 & 0.167    & 0.711 & 1.106 & 0.162     & N/A  & N/A  & N/A   \\ 
    Concrete      & 1030  &  8   & 0.336 & 0.287 & 0.162    & 0.336 & 0.287 & 0.166    & 0.341 & 0.298 & 0.186    & 0.335 & 0.282 & 0.190 \\
    Airfoil           & 1503  &  5   & 0.283 & 0.093 & 0.190    & 0.286 & 0.099 & 0.191    & 0.276 & 0.063 & 0.227    & 0.280 & 0.022 & 0.176  \\
    Parkinsons  & 5875  &  20   & 0.027 & -1.374 & 0.750    & 0.027 & -1.352 & 0.759    & 0.025 & -1.387 & 0.761    & 0.016 & -1.640 & 0.805  \\
    Power         & 9568  &  4    & 0.249 & 0.034 & 0.069    & 0.249 & 0.034 & 0.071    & 0.244 & 0.050 & 0.206    & N/A  & N/A  & N/A    \\
    Kin40K  & 40000  &  8       & 0.384 & 0.466 & 0.306    & 0.375 & 0.471 & 0.348    & 0.229 & -0.154 & 0.172       & N/A  & N/A  & N/A   \\
    Protein  & 45730  &  9      & 0.664 & 0.932 & 0.097    & 0.663 & 0.930 & 0.086    & 0.741 & 1.462 & 0.725      & N/A  & N/A  & N/A    \\
    \bottomrule
    Average $\downarrow$ &   &                    & \textbf{0.447} & \textbf{0.423} & \textbf{0.204}    & \textbf{0.446} & 0.433 & 0.232    & 0.463 & 0.455 & 0.278     & N/A  & N/A  & N/A    \\
    
    \bottomrule
  \end{tabular}
  
  \end{scriptsize}
 \end{table}

It is important to note that the softmax calibration method used in the GP-pro-sm model calibrates the aggregation weights rather than the uncertainty quantification of individual local experts, and relies on a global temperature parameter to control the degree of calibration. 
In our experiments, the temperature was fixed at 100, following \citet{Cohen2020}. 
With this setting, GP-pro-sm achieves strong performance on datasets such as $8d$ Kin40K, where it outperforms both the uncalibrated GP-pro and the proposed GP-pro-c models in terms of RMSE, NLL and ENCE. 
However, on other datasets such as $9d$ Protein, the same temperature leads to degraded performance, with GP-pro-sm exhibiting higher RMSE, NLL and ENCE than GP-pro. 
GP-pro-sm behaves very differently on Kin40K and Protein because softmax calibration amplifies differences among local experts. Kin40K benefits from aggressive reweighting due to smoothness and expert agreement, whereas Protein suffers because expert heterogeneity and noise cause unreliable experts to dominate when the same temperature is used.
These observations indicate that the effectiveness of softmax-based calibration is highly dataset-dependent and sensitive to the choice of temperature.
In contrast, the proposed information-based calibration method provides more consistent performance across datasets without requiring dataset-specific hyperparameter tuning.

While \citet{Cohen2020} use a fixed temperature for softmax calibration, they do not propose automatic tuning; in principle, the temperature could be chosen via validation metrics such as negative log-likelihood or ENCE, but this adds additional hyperparameter optimisation and computational cost. 
Moreover, a single global temperature may be insufficient to capture the heterogeneous behaviour of local GP experts, particularly in high-dimensional or large-scale settings.
In contrast, GP-pro-c provides a robust, hyperparameter-free calibration of uncertainties.

In terms of computational cost, GP-pro-c and GP-pro-sm exhibit similar overhead. 
In contrast, the cost of GP-full increases rapidly with dataset size and dimensionality due to its cubic complexity. 
As a result, GP-full is omitted from experiments on datasets with more than 5,000 training points due to prohibitive runtime and memory requirements.

\begin{table}
\caption{Average training and test times (in seconds) across four synthetic functions and six regression benchmark datasets for GP-pro-c models with uncertainty calibration, GP-pro models without calibration, GP-pro-sm models using softmax-weight calibration, and a single global GP (GP-full).} \label{tab:tab_train_test_time_baseline}
\centering
\begin{scriptsize}
\begin{tabular}{@{}lrrSSSS@{}}
    \toprule
    \textbf{Dataset} &
    \textbf{Size} &
    \textbf{Dim} &
      \multicolumn{1}{c}{\textbf{GP-pro-bt-ent-c}} &
      \multicolumn{1}{c}{\textbf{GP-pro-bt-ent}} &
      \multicolumn{1}{c}{\textbf{GP-pro-sm}} &
      \multicolumn{1}{c}{\textbf{GP-full}}  \\
      \cmidrule(r){4-4}\cmidrule(r){5-5}\cmidrule(r){6-6}\cmidrule(l){7-7}
      & & & {training, test time} & {training, test time} & {training, test time} & {training, test time}   \\
      \midrule
    Ackley           & 5000  &  10    & {3.101, 1.427}  & {3.101, 1.425}  & {3.101, 1.425}  & {47.905, 0.885}  \\ 
    Levy              & 5000  &  10   & {3.380, 1.452}  & {3.380, 1.450}  & {3.380, 1.450}  & {46.986, 0.892} \\ 
    Rastrigin        & 5000  &  10    & {3.119, 1.454}  & {3.119, 1.452}  & {3.119, 1.453}  & {50.469, 1.053}  \\ 
    Rosenbrock   & 50000  &  50   & {32.522, 85.483}  & {32.522, 85.375}  & {32.522, 85.695}    &  N/A   \\ 
   Concrete      & 1030  &  8   & {1.263, 0.037}  & {1.263, 0.036}  & {1.263, 0.037}  & {1.717, 0.033} \\
    Airfoil           & 1503  &  5    & {1.359, 0.066}  & {1.359, 0.065}  & {1.359, 0.066}  & {2.870, 0.072}   \\
    Parkinsons  & 5875  &  20  & {4.282, 2.581}  & {4.282, 2.579}  & {4.282, 2.580}  & {77.394, 1.595}  \\
    Power         & 9568  &  4   & {5.779, 0.538}  & {5.779, 0.533}  & {5.779, 0.532}       &  N/A  \\
    Kin40K  & 40000  &  8         & {27.192, 58.806}  & {27.192, 58.716}  & {27.192, 59.187}   &  N/A   \\
    Protein  & 45730  &  9       & {27.430, 59.042}  & {27.430, 58.952}  & {27.430, 59.441}     &  N/A  \\
    \bottomrule
    Average $\downarrow$ &   &                    & {10.433, 19.342}  & {10.433, 19.313}  & {10.433, 19.410} &  N/A  \\ 
     \bottomrule
  \end{tabular}
  \end{scriptsize}
\end{table}

\section{Discussion and Conclusions}
\label{sec:sec_s3_conclusion}

This work addresses the overestimation of predictive uncertainty in product-of-experts Gaussian process (GP-pro) models, which arises from training local experts on disjoint data subsets. 
The proposed information-based variance calibration method is therefore tailored to GP-pro models. 
While this focus limits immediate applicability to a specific class of GP architectures, it enables us to exploit structural properties unique to the product-of-experts formulation, most notably, the discrepancy between the information available to individual experts and that available to the full dataset. 
This discrepancy admits a principled characterisation in terms of conditional mutual information, which yields a theoretically grounded calibration ratio for reducing overestimated posterior variances.

The proposed calibration preserves the computational advantages and aggregation structure of GP-pro while improving uncertainty estimates and maintaining predictive accuracy. 
Empirical results show consistent reductions in NLL and ENCE across synthetic functions and real-world regression datasets. 
On average, GP-pro-c reduces NLL and ENCE by 2.3\% and 12.0\%, respectively, compared to GP-pro without calibration. 
In terms of computational efficiency, GP-pro-c mitigates the cubic complexity of standard GP regression by operating on local experts. 
For datasets where a global GP (GP-glo) is tractable, GP-pro-c achieves an average reduction of 72.5\% in computational cost. 
For larger datasets, GP-glo becomes infeasible due to runtime or memory constraints, whereas GP-pro-c remains tractable, highlighting its practical scalability.

Among the GP-pro-c variants, the combination of balltree assignment and entropy-based weighting performs best overall. 
Empirical results further suggest that using 200-400 data points per expert provides a favourable balance between predictive accuracy, calibration quality, and computational cost.

The results presented in this study demonstrate consistent, if moderate, improvements in expected normalised calibration error (ENCE) for the GP-pro-c model compared to the baseline GP-pro. 
This is expected, as the proposed method applies a controlled correction based on information gain rather than aggressively reshaping the posterior distribution. 
Importantly, these improvements are achieved without degrading predictive accuracy, as reflected by stable RMSE values across all experiments. 
Furthermore, it is important to note that ENCE is inherently bounded below by model misspecification and observational noise. 
Reliability diagrams support these findings: although GP-pro-c does not uniformly outperform GP-pro across all probability levels, it consistently moves predictive uncertainty closer to ideal calibration in regions where GP-pro exhibits inflated variance. 
Overall, the method improves uncertainty calibration in a conservative and robust manner.

Several directions for future work remain. 
First, extending the method to alternative kernel classes is a promising avenue. 
Although the derivation relies on standard assumptions related to normalised information gain (with reference to Mat\'{e}rn kernels), the calibration rule itself depends only on posterior variances and noise levels. 
This suggests that the approach may generalise to other stationary kernels, although formal analysis is left for future work. 
Second, incorporating joint calibration across experts and explicitly modelling dependencies between local GPs may further reduce variance overestimation, particularly in overlapping regions. 
Third, establishing theoretical guarantees for posterior variance calibration would strengthen the methodological foundation of GP-pro-c.

Another promising direction is the integration of GP-pro-c with dimensionality-reduction approaches, such as principal component-based methods \cite{Higdon2008}. 
These approaches are complementary: GP-pro-c addresses scalability and uncertainty calibration in large input spaces, while dimensionality reduction targets high-dimensional outputs. 
Combining these strategies could enable efficient modelling of problems with both large input datasets and high-dimensional outputs.

Finally, integrating GP-pro-c into Bayesian optimisation (BO) pipelines represents a natural direction for future work. 
Gaussian processes are widely used as surrogate models in BO \cite{Brochu2010,Shahriari2016p,Frazier2018,Garnett2023}, yet their scalability remains a major challenge. Moreover, BO is particularly sensitive to the quality of uncertainty estimates, as posterior variances directly determine the behaviour of acquisition functions. 
Overestimated uncertainties can encourage excessive exploration, leading to less efficient optimisation. 
By combining scalability with improved uncertainty calibration, GP-pro-c is well positioned to address these challenges, potentially enabling more reliable acquisition decisions and greater optimisation efficiency. 
Although a comprehensive evaluation within BO is beyond the scope of the present paper, the calibration improvements demonstrated here provide a strong foundation for future investigation of GP-pro-c as a scalable and reliable surrogate model for BO.

\begin{acks}
The first author gratefully acknowledges Rasul Tutunov, Haitham Bou-Ammar, and Sye Loong Keoh for their valuable feedback and insightful discussions during the early development of the information-based calibration method. The authors also thank the anonymous reviewers for their constructive comments and suggestions, which have helped improve the clarity and quality of this paper.
\end{acks}

\printbibliography

\appendix

\section{Information Gain}
\label{appendix_info_gain}

Equation (\ref{eq:eq_c2_2}) shows that the information gain for the data points in the set $\mathcal{A} = \{ \vx_1, \vx_2, \dots, \vx_N \}$ can be expressed in terms of predictive variances. This result was established by \citet{Srinivas2012} and \citet{Dorard2012} as follows.

In Equation (\ref{eq:eq_c2_1}), the first entropy term, $H(\vy_{\mathcal{A}})$ or equivalently $H(\vy_N)$, can be expressed recursively as
$H(\vy_N) = H(\vy_{N-1}) + H(y_N \mid \vy_{N-1})$. Conditioned on $\vy_{N-1}$, the inputs $(\vx_n)_{1 \leqslant n \leqslant N}$ are deterministic, and thus
$f(\vx) \sim \mathcal{N}(0, \sigma_{N-1}^2(\vx))$ for all $\vx$.

Here, note that $\vx_1, \dots, \vx_N$ are deterministic given $\vy_{N-1}$, and that the conditional variance $\sigma_{N-1}^2(\vx_N)$ does not depend on $\vy_{N-1}$. Meanwhile,
$y_N = f(\vx_N) + \epsilon_N$ is the sum of two independent zero-mean Gaussian random variables: one with variance $\sigma_{N-1}^2(\vx_N)$ and the other with variance $\ddot{\sigma}^2$ (the observation noise variance).

By applying the entropy formula $H(\mathcal{N}(\mu, \Sigma)) = \frac{1}{2} \log |2 \pi e \Sigma|$, we obtain
\[
H(y_N \mid \vy_{N-1}) = \frac{1}{2} \log \left( 2 \pi e \left( \sigma_{N-1}^2(\vx_N) + \ddot{\sigma}^2 \right) \right)
\]
Thus, expressing $H(\vy_N)$ recursively yields
\begin{align*}
H(\vy_N) = \frac{1}{2} \sum_{n=1}^{N} \log \left( 2 \pi e \left( \ddot{\sigma}^2 + \sigma_{n-1}^2(\vx_n) \right) \right)
\end{align*}

For the second term in Equation (\ref{eq:eq_c2_1}), namely $H(\vy_{\mathcal{A}} \mid \vf)$ or equivalently $H(\vy_N \mid \vf)$, note that conditioned on $\vf$, the vector $\vy_N$ follows a zero-mean Gaussian distribution with covariance matrix $\ddot{\sigma}^2 \mI_N$. Therefore,
\begin{align*}
H(\vy_N \mid \vf)
&= \frac{1}{2} \log \left| 2 \pi e \ddot{\sigma}^2 \mI_N \right| \\
&= \frac{1}{2} \sum_{n=1}^{N} \log \left( 2 \pi e \ddot{\sigma}^2 \right)
\end{align*}

Combining these two terms, we obtain
\begin{align*}
I(\vy_{\mathcal{A}}; \vf)
&= H(\vy_N) - H(\vy_N \mid \vf) \\
&= \frac{1}{2} \sum_{n=1}^{N} \log \left( 2 \pi e \left( \ddot{\sigma}^2 + \sigma_{n-1}^2(\vx_n) \right) \right)
- \frac{1}{2} \sum_{n=1}^{N} \log \left( 2 \pi e \ddot{\sigma}^2 \right) \\
&=  \frac{1}{2} \sum_{n=1}^{N} \log \left(  \frac{  2 \pi e \ddot{\sigma}^2 + 2 \pi e  \sigma_{n-1}^2(\vx_n)  }{ 2 \pi e \ddot{\sigma}^2 }  \right) \\
&= \frac{1}{2} \sum_{n=1}^{N} \log \left( 1 + \frac{\sigma_{n-1}^2(\vx_n)}{\ddot{\sigma}^2} \right)
\end{align*}

Here, $\sigma_{n-1}^2(\vx_n)$ denotes the posterior variance of $f(\vx_n)$ given observations $\{ y_1, \dots, y_{n-1} \} \subseteq \vy_{\mathcal{A}}$. As given in Equation (\ref{eq:eq_gp_sigma}), this quantity does not depend on the observed values themselves.

Since the sum can therefore be evaluated without observing the data, it is possible to compute the mutual information that would be gained from any hypothetical set of observations \cite{Srinivas2012,Desautels2014}.

\section{Additional Experimental Results}
\label{appendix_exp_results}
 
\begin{figure}
    \centering 

\begin{subfigure}{0.3\textwidth}
  \includegraphics[width=\linewidth]{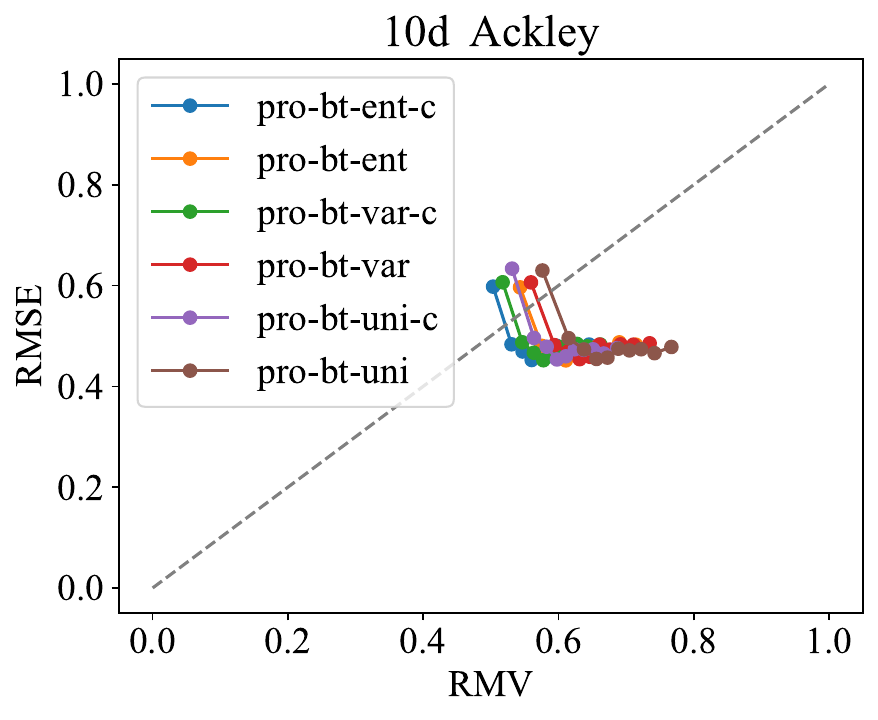}
\end{subfigure}\hfil 
\begin{subfigure}{0.3\textwidth}
  \includegraphics[width=\linewidth]{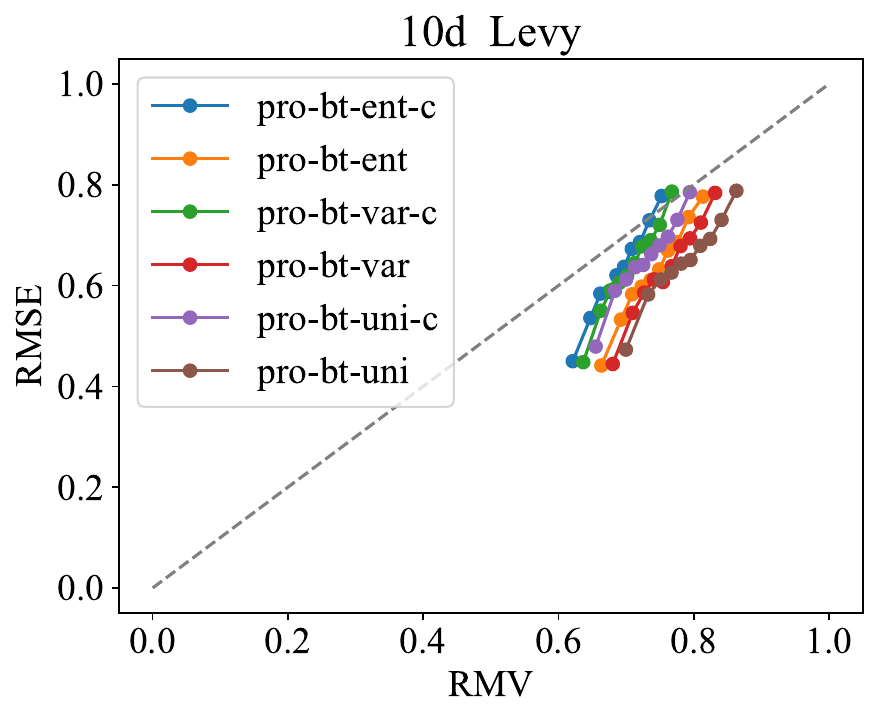}
\end{subfigure}\hfil 
\begin{subfigure}{0.3\textwidth}
  \includegraphics[width=\linewidth]{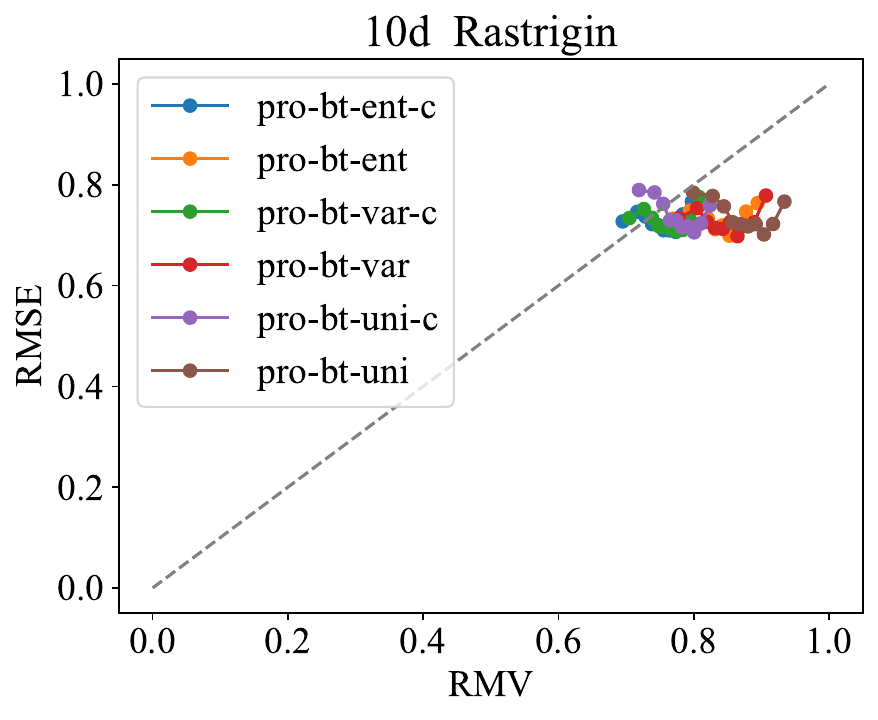}
\end{subfigure}\hfil 

\begin{subfigure}{0.3\textwidth}
  \includegraphics[width=\linewidth]{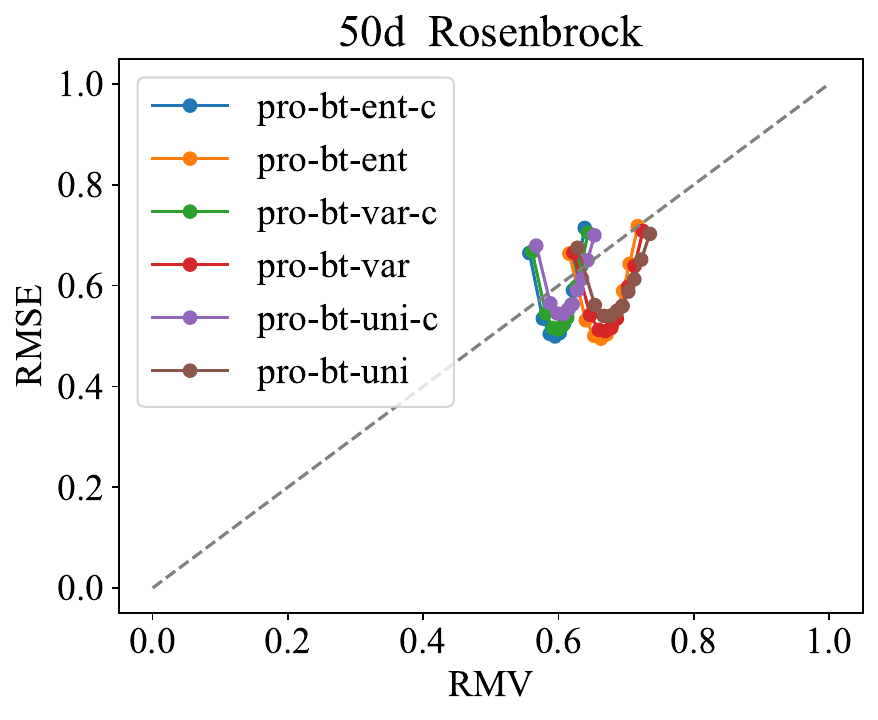}
\end{subfigure}\hfil 
\begin{subfigure}{0.3\textwidth}
  \includegraphics[width=\linewidth]{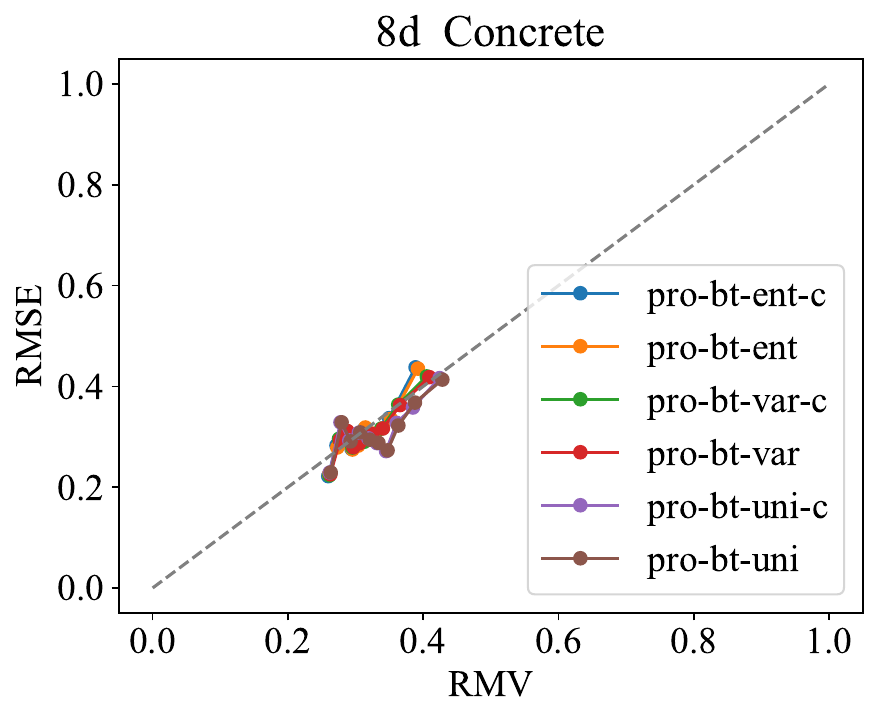}
\end{subfigure}\hfil 
\begin{subfigure}{0.3\textwidth}
  \includegraphics[width=\linewidth]{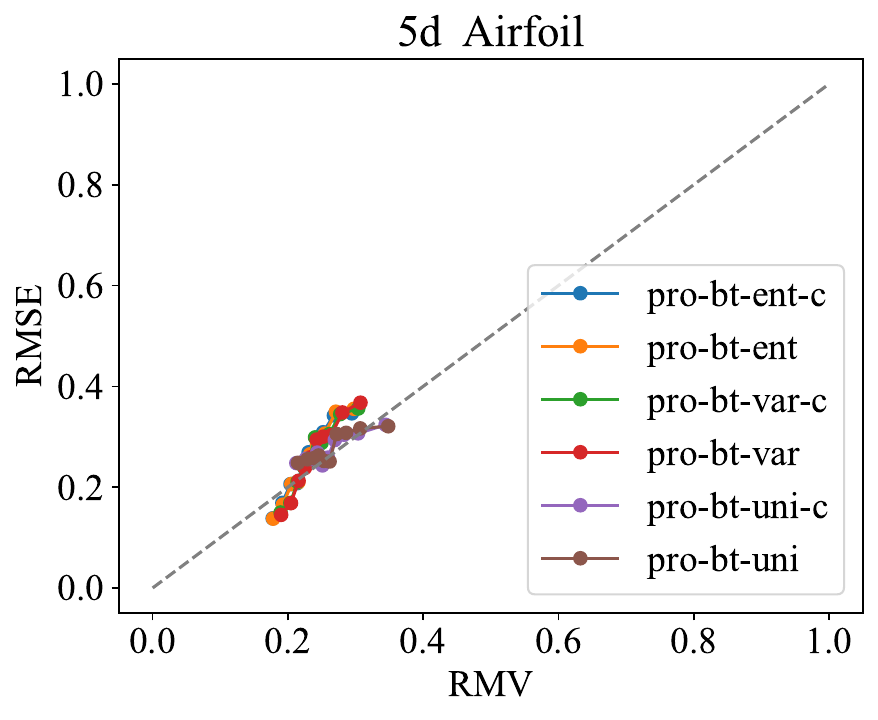}
\end{subfigure}\hfil 

\begin{subfigure}{0.3\textwidth}
  \includegraphics[width=\linewidth]{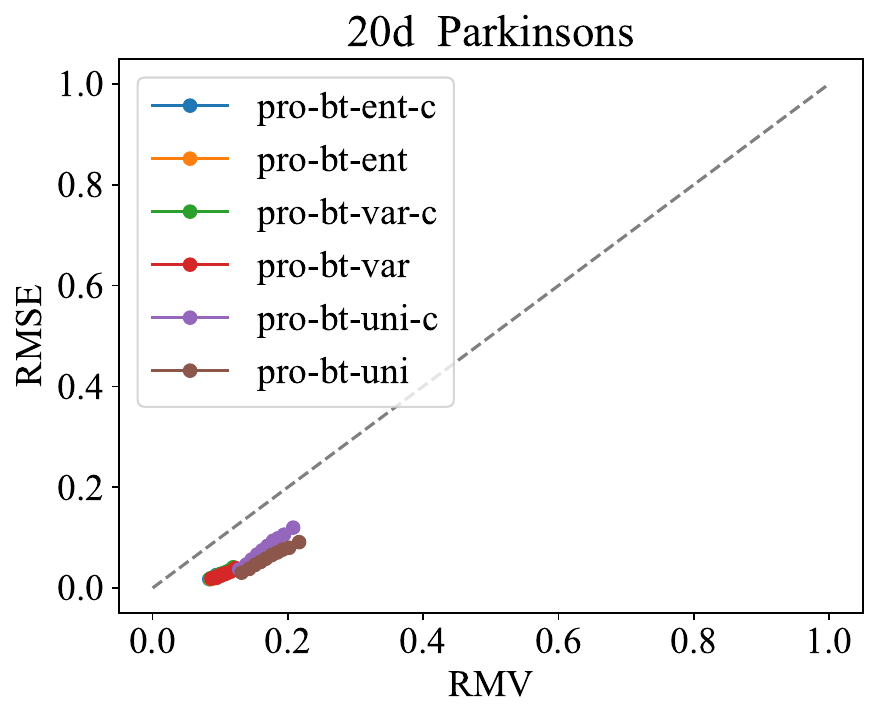}
\end{subfigure}\hfil 
\begin{subfigure}{0.3\textwidth}
  \includegraphics[width=\linewidth]{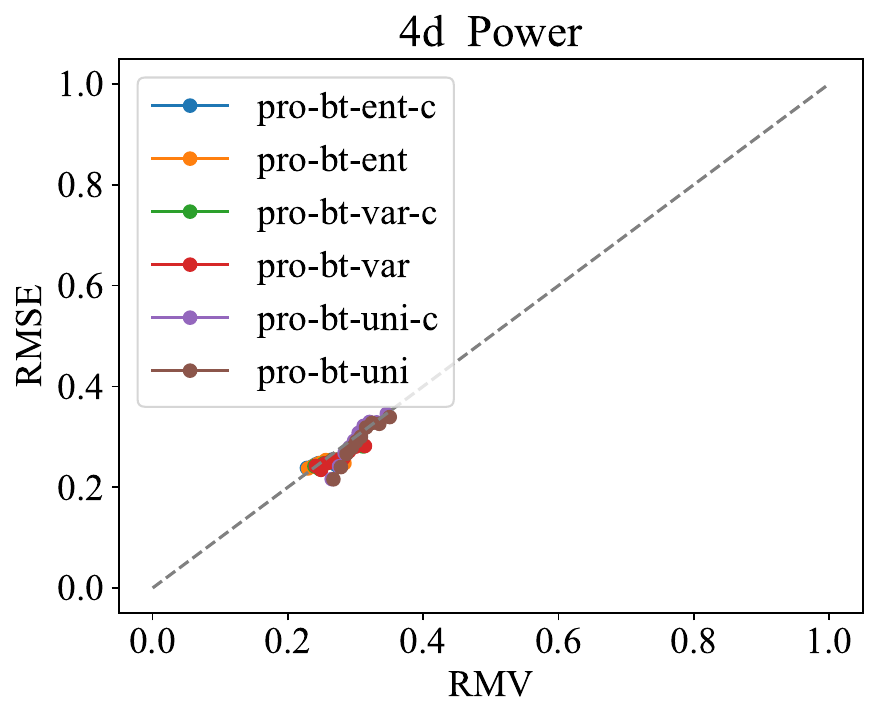}
\end{subfigure}\hfil 
\begin{subfigure}{0.3\textwidth}
  \includegraphics[width=\linewidth]{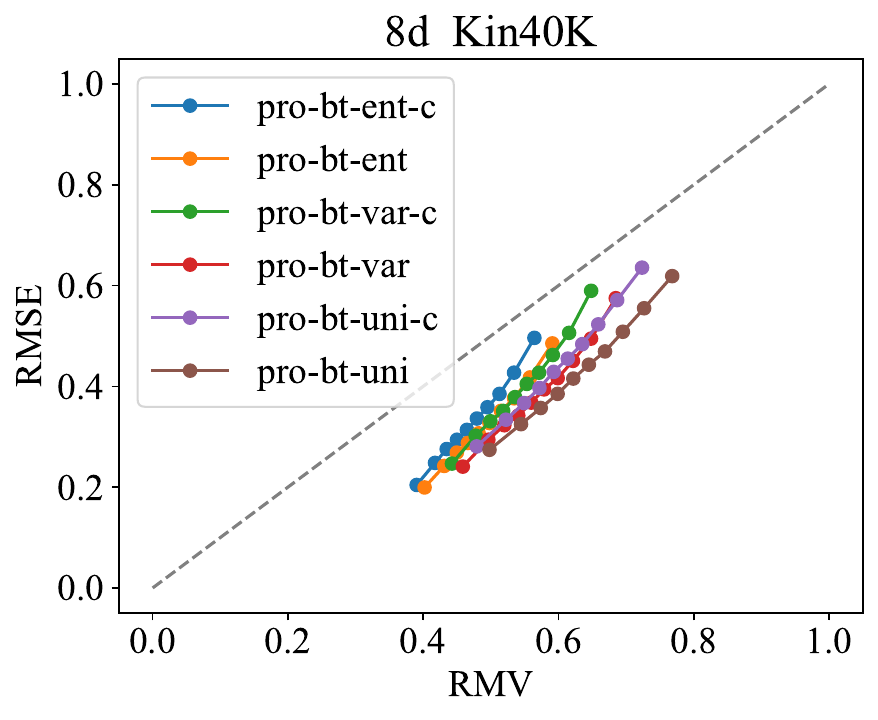}
\end{subfigure}\hfil 

\begin{subfigure}{0.3\textwidth}
  \includegraphics[width=\linewidth]{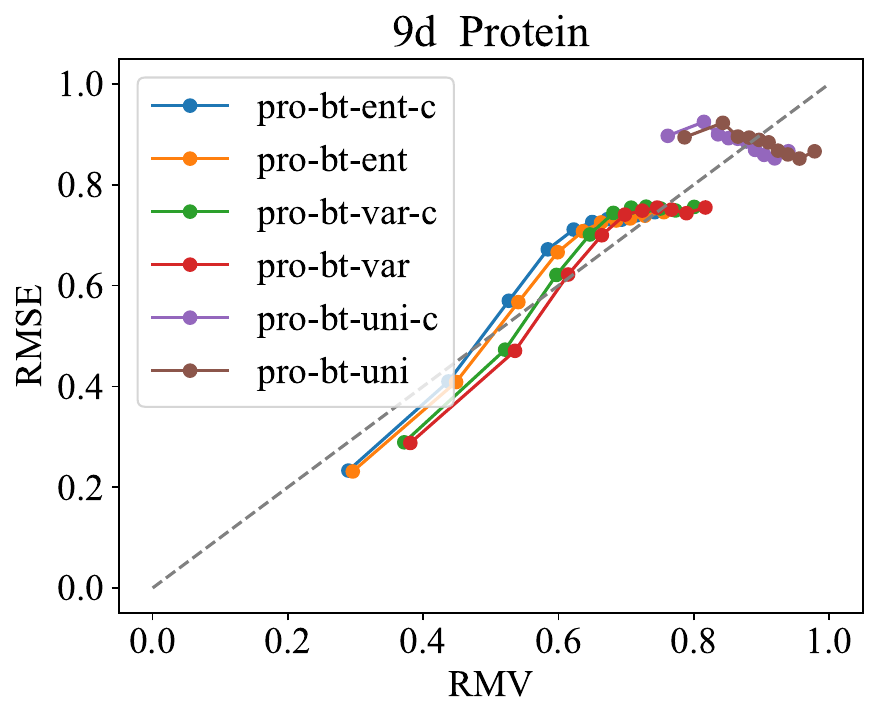}
\end{subfigure}\hfil 

\caption{Reliability diagrams for the GP-pro-c and GP-pro models using the balltree assignment method, comparing entropy-, variance-, and uniform-weighted schemes on four synthetic benchmark functions and six regression datasets. Under all weighting schemes, GP-pro-c models with uncertainty calibration produce curves closer to the ideal diagonal line than GP-pro models without uncertainty calibration.}

\Description{Ten line graphs show reliability diagrams for Gaussian process models on the 10-dimensional Ackley function, 10-dimensional Levy function, 10-dimensional Rastrigin function, 50-dimensional Rosenbrock function, 8-dimensional Concrete dataset, 5-dimensional Airfoil dataset, 20-dimensional Parkinsons dataset, 4-dimensional Power dataset, 8-dimensional Kin40K dataset, and 9-dimensional Protein dataset.

Each line graph has Root Mean Variance on the horizontal axis (ranging from 0 to 1 in increments of 0.2) and Root Mean Square Error on the vertical axis (ranging from 0 to 1 in increments of 0.2).

Each plot shows seven curves: a dashed diagonal reference line and six model variants corresponding to combinations of balltree assignment with entropy-, variance-, and uniform-weighted schemes, each evaluated with and without uncertainty calibration.

Across all datasets, models with uncertainty calibration consistently produce curves closer to the ideal diagonal line than their uncalibrated counterparts.}

\label{fig:fig_reliability_bt}
\end{figure}

Figure \ref{fig:fig_reliability_bt} shows reliability diagrams for the GP-pro-c and GP-pro models using the balltree assignment method, comparing entropy-, variance-, and uniform-weighted schemes. It is observed that, under all weighting schemes, variants of the GP-pro-c model produce curves closer to the ideal diagonal line than the corresponding GP-pro models.

\end{document}